\documentclass[lettersize,journal]{IEEEtran}
\usepackage{amsmath,amsfonts}
\usepackage{algorithmic}
\usepackage{algorithm}
\usepackage{array}
\usepackage{textcomp}
\usepackage{stfloats}
\usepackage{url}
\usepackage{verbatim}
\usepackage{graphicx}
\usepackage{cite}
\usepackage{color}
\usepackage{multirow}
\usepackage{mathtools}
\usepackage{array}%调整表格线与上下内容的间隔
\usepackage{booktabs} %调整表格线与上下内容的间隔
\usepackage{subfigure}
\graphicspath{{figure/}}

\newtheorem{theorem}{Theorem}

\newtheorem{lemma}{Lemma}

\newtheorem{remark}{Remark}

\newtheorem{definition}{Definition}

\newcounter{nextauthor}
\def\mathrm{\mbox}

\begin{document}

\title{Tensor Factorization via Transformed Tensor-Tensor Product  for  Image Alignment}

\author{Sijia Xia, Duo Qiu, and Xiongjun Zhang
        % <-this % stops a space
\thanks{ The research of D. Qiu was supported in part by
the National Natural Science Foundation of China under Grant No. 12201473
and the Science Foundation of
Wuhan Institute of Technology under Grant No. K202256.
The research of X. Zhang was supported in part by
the National Natural Science Foundation of China under Grant No. 12171189
and the Knowledge Innovation Project of Wuhan under Grant No. 2022010801020279.}% <-this % stops a space
\thanks{S. Xia is  with the School of Mathematics and Statistics,
Central China Normal University, Wuhan 430079, China (e-mail: sijiax@mails.ccnu.edu.cn).}
\thanks{D. Qiu is with the School of Mathematics and Physics, Wuhan Institute of Technology, Wuhan 430205, China (e-mail: qiuduoduo13@wit.edu.cn).}
\thanks{X. Zhang is with the School of Mathematics and Statistics, and Hubei Key Laboratory of Mathematical Sciences,
Central China Normal University, Wuhan 430079, China (e-mail: xjzhang@mail.ccnu.edu.cn).}
}



\maketitle

\begin{abstract}
In this paper, we study the problem of a batch of linearly correlated image alignment,
where the observed images are deformed by some
unknown domain transformations, and  corrupted
by additive Gaussian noise and sparse noise simultaneously.
By stacking these images as the frontal slices of a third-order tensor,
we propose to utilize the tensor factorization method via transformed tensor-tensor product to
explore the low-rankness of the underlying tensor,
which is factorized into the product of two smaller tensors via transformed tensor-tensor product
under any unitary transformation.
The main advantage of transformed tensor-tensor product is that
its  computational complexity is
lower compared
with the existing literature based on transformed tensor nuclear norm.
Moreover, the tensor $\ell_p$ $(0<p<1)$ norm is employed to characterize
the sparsity of sparse noise
and the tensor  Frobenius norm is  adopted to model additive Gaussian noise.
A generalized Gauss-Newton algorithm is designed to solve the resulting model
 by linearizing the domain transformations
 and a proximal
Gauss-Seidel algorithm is developed to solve the corresponding subproblem.
Furthermore, the convergence of
the proximal Gauss-Seidel algorithm is established, whose
convergence rate  is also analyzed based on the Kurdyka-$\L$ojasiewicz property.
Extensive numerical experiments on real-world image datasets
are carried out to demonstrate the superior performance of
the proposed method  as compared to  several state-of-the-art methods
in both accuracy and computational time.
\end{abstract}


\section{Introduction}
\IEEEPARstart{I}{mages} and videos sharing in social networks
have increased dramatically in the past decades.
However, these images may suffer from  various degradations during the acquirement and transmission process,
such as impulse noise, partial  occlusion, illumination variations, and misalignment \cite{szeliski2007image, Peng_2012}.
It is a vital task to recover these images from the observations with these degradations
for subsequent applications including  face recognition,
image classification, and object tracking \cite{yilmaz2006object, wright2008robust}.
In this paper, we study the problem of aligning
a batch of linearly correlated images, which are deformed by some unknown domain transformations and
 corrupted by sparse noise (such as impulse noise, partial  occlusion) and Gaussian noise.

A lot of work has been done for the task of batch image alignment in the literature,
where the images of an object or
objects of interest are aligned to a fixed canonical template \cite{Wu_2012}.
For example, Learned-Miller \cite{learned2005data} present a congealing approach for a batch of image alignment,
which aimed to minimize the sum of the pixel-stack entropies via transforming a set
of images of a class.
Here the main idea is to make a set of images appear as similar
as possible by removing nuisance variables within these images.
Cox et al. \cite{Cox2008,Cox2009} proposed least squares
congealing methods for large number of image alignment
by minimizing the sum of squared distances between pairs
of images to acquire a low-rank matrix.
Moreover, Vedaldi et al. \cite{Vedaldi_2008} proposed  to minimize a logarithmic
determinant loss function to align a large collection
of image data,
which can be regarded as a smooth  surrogate
of the rank function.
However, the above approaches may fail to
align a batch of linearly correlated images effectively when
there are large illumination
variations and gross pixel corruptions or partial occlusion simultaneously,
which often occurs in real-world images \cite{Peng_2012}.

By unfolding the well-aligned and linearly correlated images into vectors
and  stacking these vectors as columns of  a matrix,
the resulting matrix is low-rank and
some low-rank matrices based methods were proposed and studied for batch image alignment.
For instance, Peng et al. \cite{Peng_2012} first proposed
a robust image alignment method via seeking an optimal
set of image domain transformations such that the matrix of the transformed images
can be decomposed into the sum of a sparse error matrix and a low-rank matrix of the well-aligned images,
where  the misalignment of images was modeled as a domain transformation \cite{ma2004invitation}.
Then a sequence of convex programs that minimized the sum of $\ell_1$ norm and nuclear norm of
the two component matrices were proposed to approximate the previous problem by linearizing the domain transformations.
Furthermore, in order to improve the computational efficiency,
He et al. \cite{He_2014} proposed a low-rank matrix
factorization method by minimizing the $\ell_1$ norm of the error term
and decomposing the underlying matrix into the product of two small matrices,
where the first factor matrix belongs to an orthogonal subspace.
However, the $\ell_1$ norm is just suboptimal to
obtain a sparse solution in statistic learning \cite{fan2001variable, zou2006adaptive},
and some nonconvex penalized methods were proposed to
improve the sparsity for variable selection \cite{fan2001variable, zhang2010nearly}.
Moreover, Chen et al. \cite{Chen_2016} proposed a nonconvex model for noisy image alignment,
where the nonconvex functions were employed onto the singular values of the well-aligned images
and each entry of the sparse error component, respectively.
However, the matrix based methods will destroy the internal structure of these images by the unfolding approach.
Besides, only the correlation of different images is utilized in these methods.

On the other hand,
Wu et al. \cite{Wu_2012} proposed an online robust alignment method by
using a low-rank matrix decomposition technique and then updating the basis online,
where one needs to align a newly arrived image to previously well-aligned images.
Then Song et al. \cite{Song2016} proposed to combine the geometric transformation
with online principal component analysis to update the basis matrix online
via a stochastic gradient descent algorithm for image alignment.
Furthermore, Zheng et al. \cite{8630746} proposed an online subspace learning approach from
  image gradient orientations for robust image alignment, which provides a
 more reliable low-dimensional subspace than that from pixel intensities.
However, the online based methods need some well-aligned images in advance for seeking the basis matrix,
which may be challenged when  only noisy and misalignment images are available.

In order to explore the internal structure of these well-aligned images better,
some tensor based methods were proposed and studied for image alignment,
where each image is regarded as the frontal slice of a tensor.
By using the domain transformed images to  be decomposed
into the sum of a low Tucker rank tensor and a sparse error tensor,
Zhang et al. \cite{Zhang2013} proposed to employ
the sum of nuclear norms of unfolding matrices of a tensor \cite{Liu2013}
to approximate the sum of entries of the Tucker rank of the underlying images,
where the tensor $\ell_1$ norm is used to characterize the sparsity of the error tensor.
Li et al. \cite{Li2019}  present an online robust low-rank tensor method
by applying the matrix factorization technique for the unfolding matrices of a
tensor to approximate its Tucker rank and the tensor $\ell_1$ norm to approximate the sparse error.
However, the tensor $\ell_1$ norm may get a biased estimator
for a sparse solution, which was demonstrated in statistical learning \cite{fan2001variable, zou2006adaptive}.
Note that the sum of nuclear norms of unfolding matrices of a tensor
utilizes the matrix  nuclear norm to approximate  the rank of the correspondingly unfolding matrix.
While the nuclear norm of a matrix is just the $\ell_1$ norm of the singular value vector,
which also obtains a bias estimator \cite{Bunea2011}.
Furthermore, Zhang et al. \cite{Zhang_2021} extended the model in \cite{Zhang2013} to
the nonconvex model by using $\ell_p$ norm ($0<p<1$) for each entry of the sparse error term and the singular values of the unfolding matrices of the underlying tensor for the low Tucker rank component.
However, the sum of nuclear norms of unfolding matrices
of a tensor is not the convex envelope of the
sum of entries of the Tucker rank of a tensor \cite{RomeraParedes2013},
which is just suboptimal for tensor completion \cite{mu2014square}.

Recently, the tensor-tensor product and tensor singular value decomposition (SVD) based on tensor
algebra operations were proposed for  third-order tensors \cite{Kilmer2011, Kilmer2013},
where the tensor-tensor product was the multiplication of  a block-circulant matrix and a block column vector.
Then Semerci  et al. \cite{Semerci2014} proposed a tensor nuclear norm (TNN) method
to approximate the multi-rank of a tensor for multienergy computed tomography.
Moreover, Zhang et al. \cite{Zhang2014} applied the TNN  for low-rank tensor completion in visual data
based on tensor SVD, and showed that the TNN is the tightest convex relaxation of the sum
of multi-rank of a tensor.
The TNN and tensor-tensor product are widely used in low-rank tensor
recovery problems due to its superiority of characterising the low-rankness of a tensor,
see \cite{lu2019tensor,zhang2019corrected, zheng2019mixed, hao2013facial, Zhou_2018, martin2013order, hou2022robust, 9372832} and references therein.
However, the TNN is just the sum of nuclear norms
of all frontal slice matrices of a tensor in the Fourier domain.
And  a limitation by using discrete Fourier transform is
that the periodicity is assumed. Moreover, the discrete Fourier transform may not result in a
low-rank tensor, while a low-rank requirement is preferred for low-rank tensor recovery \cite{qiu2021nonlocal}.
In order to overcome the limitation of the Fourier transform in TNN,
Song et al. \cite{Song_2020} proposed a transformed
tensor-tensor product
and transformed tensor SVD under any unitary transformation for robust tensor completion,
which can obtain a lower transformed multi-rank tensor by using suitable unitary transformations.
The transformed tensor SVD based methods have been
applied in various low-rank tensor optimization problems,
see  \cite{zhang2021low, qiu2021nonlocal, song2020tensor, ng2020patched} and references therein.
While the above methods cannot be applied to the images with misalignment.
Very recently,  Qiu et al. \cite{Qiu_2021} proposed a transformed tensor nuclear norm
plus tensor $\ell_1$ norm method for batch image alignment,
where the  images after domain transformations can be decomposed into the sum
of a low transformed multi-rank tensor and a sparse error tensor.
Moreover, they extended it to a family of nonconvex models based on transformed tensor nuclear norm,
where the nonconvex functions are employed onto the singular values
of all frontal slices of the underlying tensor in the transformed domain and
each entry of the error tensor.
However, the above method is very expensive in the computational complexity since one needs
to compute  the SVDs of all frontal slices of a tensor in the transformed domain.

%\subsection{Contributions}
In this paper, we propose a tensor factorization method
for a batch of linearly correlated image alignment,
where the observed images are deformed by some unknown domain
transformations and corrupted by sparse noise (such as impulse noise,  partial occlusion)
and Gaussian noise simultaneously.
By stacking these images into a third-order tensor along the third-dimension,
the underlying tensor is factorized into the transformed  tensor-tensor product
of two smaller tensors under any unitary transformation,
which is capable of  reducing the computational complexity compared with the transformed tensor nuclear norm.
Moreover, the tensor $\ell_p$ norm  ($0<p<1$)  is employed to enhance the sparsity of the sparse noise,
where a sparser solution can be obtained  than that of the tensor $\ell_1$ norm.
And the tensor Frobenius norm of the difference
between the observation images with domain transformations and the underlying tensor,
the sparse tensor, is utilized to characterize the Gaussian noise.
Besides, a generalized Gauss-Newton algorithm \cite{Jittorntrum1980}
is developed to solve the proposed model by linearizing the domain transformations,
where a proximal Gauss-Seidel algorithm is applied to solve the corresponding subproblem.
Furthermore, the convergence of
 the proximal Gauss-Seidel algorithm is established under very mild conditions, whose
convergence rate is also analyzed based on the Kurdyka-$\L$ojasiewicz property \cite{Bolte2013}.
Numerical examples on many real-world image datasets
are reported to demonstrate that the proposed approach outperforms many state-of-the-art ones
in terms of visual quality and computational time for the task of image recovery and alignment.
%\subsection{Organization}

The remaining parts of this paper are organized as follows.
Section \ref{chap2} is devoted to some notation and preliminaries
about transformed tensor-tensor product under any unitary transformation for third-order tensors.
In Section \ref{chap3},
we propose a tensor factorization method based on transformed
tensor-tensor product for image alignment.
In Section \ref{OptAlgGGN},
a generalized Gauss-Newton algorithm is designed to solve
the resulting model and a proximal Gauss-Seidel algorithm is developed to solve the
corresponding subproblem,
where the convergence
and convergent rate of this algorithm are established. Numerical examples
are reported to demonstrate the effectiveness and efficiency  of the proposed method in Section \ref{NNMEX}.
We conclude this paper in Section \ref{Concludsion}.
The proofs of theorems are left in Appendix.
%supplementary material.
%Appendix.

\section{PRELIMINARY}\label{chap2}
In this section, we give some notation of tensors
and briefly introduce the definition of transformed tensor-tensor product under any unitary transformation.

\subsection{Notation and Notions}

Scalars, vectors, matrices, and tensors are denoted by  lowercase letters,
lowercase boldface letters, uppercase boldface letters, and  capital Euler script
letters, respectively, e.g., $x,\mathbf{x},\mathbf{X}, \mathcal{X}$.
The order of a tensor is the number of ways or modes \cite{Kolda2009}.
For an arbitrary third-order tensor $\mathcal A\in \mathbb {C}^{n_1\times n_2\times n_3}$,
its $(i,j,k)$th element is represented as $\mathcal{A}_{ijk}$.
The $i$th frontal slice of $\mathcal{A}$ is denoted as
$\mathbf{A}^{(i)}=\mathcal A(:,:,i)$, $i=1,2,\ldots,n_3$.
The $(i,j)$th tube  of $\mathcal{A}$ is denoted by $\mathcal{A}(i,j,:), i=1,\ldots, n_1,j=1,\ldots, n_2.$
$\mathcal{A}_{(3)}$ denotes the mode-$3$ unfolding of $\mathcal{A}$, which
 arranges the mode-$3$ fibers to be the columns of the resulting matrix.
 Here a fiber is defined by fixing every index but one \cite{Kolda2009}.
 Conversely, the inverse operator of mode-$3$ unfolding is denoted by $\textup{Fold}_3(\cdot)$, i.e.,
 $\textup{Fold}_3(\mathcal{A}_{(3)})=\mathcal{A}$.

The inner product of two tensors
$\mathcal{A},\mathcal{B}\in\mathbb{C}^{n_1\times n_2\times n_3}$
is defined as $\langle \mathcal{A},B\mathcal{} \rangle=\sum_{i=1}^{n_3}\text{Tr}(\mathbf{A}^{(i)}(\mathbf{B}^{(i)})^T)$,
where $\text{Tr}(\cdot)$ denotes the trace of a matrix and $\cdot^T$ denotes the conjugate transpose of a matrix.
The tensor Frobenius norm of $\mathcal A$ is defined as
$\|\mathcal A\|_{F}=\sqrt{\langle\mathcal A,\mathcal A\rangle}$.
%The $\ell_0$-norm of $\mathcal A$ is the number of non-zero entries in $\mathcal A$ .
%The $\ell_1$-norm of $\mathcal A$ is  defined as $\|\mathcal A\|_{1}=\sum_{i=1}^{n_1}\sum_{j=1}^{n_2}\sum_{k=1}^{n_3}|a_{ijk}|$.
The tensor $\ell_p$ $(0<p<1)$ norm of $\mathcal{A}$ is defined as
$\|\mathcal A\|_{p}=(\sum_{i,j,k}|\mathcal{A}_{ijk}|^{p})^\frac{1}{p}.$

For an extended real valued function $f:\mathbb{R}^n\rightarrow [-\infty,+\infty]$,
its effective domain is defined as
$\text{dom} \it(f)=\{x\in \mathbb R^{n}:f(x)< +\infty\}.$
For a nonempty closed set $K\subseteq\mathbb{R}^n$, we use
$\text{dist}(\mathbf{x},K)$ to denote the distance from $\mathbf{x}$ to $K$, where
$
\text{dist}(\mathbf{x},K)=\inf_{ \mathbf{z}\in K}\|\mathbf{x}-\mathbf{z}\|.
$
\subsection{Transformed Tensor-Tensor Product}
In this subsection, we introduce the definition of the transformed tensor-tensor
product under any unitary transformation for third-order tensors,
which is adopted from  \cite{Kernfeld2015, Song_2020} directly.
We denote a unitary matrix by $\mathbf U \in \mathbb {C}^ {n_3\times n_3}$
satisfying $\mathbf U\mathbf U^{T}=\mathbf U^{T}\mathbf U=\mathbf I_{n_3}$,
where  $\mathbf I_{n_3}$ is an $n_3\times n_3$ identity matrix. % and $\mathbf{U}^T$ denoted the conjugate transpose of $\mathbf{U}$.
For an arbitrary third-order tensor $\mathcal {A}\in \mathbb {C}^{n_1\times n_2\times n_3}$,
we denote $\hat {\mathcal A}_\mathbf U$ as the result of multiplying
by $\mathbf U$ on all tubes along the third-dimension of $\mathcal {A}$, i.e.,
$\hat{\mathcal A}_\mathbf U(i,j,:)=\mathbf U(\mathcal {A}(i,j,:)), i=1,\ldots, n_1, j=1,\ldots, n_2$.
For simplicity,
it is also expressed as $\hat{\mathcal A}_\mathbf U=\mathbf U[\mathcal {A}]$.
Next, we define the block diagonal matrix $\bar{\mathcal A}_\mathbf U$ as
$$
{\bar{\mathcal A}}_{\mathbf U}=\text{bdiag}({\hat{\mathcal A}}_{\mathbf U})
:=\begin{bmatrix}
    {\hat{\mathcal A}}_\mathbf U^{(1)}   & &     \\
    % & \hat{\mathcal A}_\mathbf U^{(2)} & &      \\
       &\ddots &  \\
      & & \hat{\mathcal A}_\mathbf U^{(n_3)}
\end{bmatrix},
$$
where
%$\text{bdiag}$$(\cdot)$ represents the diagonal matrix operator,
$\hat{\mathcal A}_\mathbf U^{(k)}$ is the $k$th frontal slice of the tensor $\hat{\mathcal A}_\mathbf U$, $k=1,\ldots, n_3$.
The inverse operator of $\text{bdiag}(\cdot)$, denoted by $\text{fold}_{3}(\cdot)$,
transforms the block diagonal matrix into a third-order tensor,
where the $i$th frontal slice of the resulting tensor is the $i$th diagonal block,
i.e.,
$
\text{fold}_{3}(\text{bdiag}({\hat{\mathcal A}}_{\mathbf U})):={\hat{\mathcal A}}_{\mathbf U}.
$

Now we define the transformed tensor-tensor product via an arbitrary unitary matrix (called $\mathbf U$-product).

\begin{definition}\cite[Definition 1]{Song_2020}
The $\mathbf U$-product of arbitrary
two tensors $\mathcal {A}\in \mathbb {C}^{n_1\times n_2\times n_3}$ , $\mathcal {B}\in \mathbb{C}^{n_2\times l\times n_3}$ is given by
$
\mathcal C=\mathcal A\diamond_{\mathbf U}\mathcal B=\mathbf U^{T}[\text{\rm{{fold}}}_{3}(\text{\rm{bdiag}}(\hat {\mathcal A}_\mathbf U)\cdot \text{\rm{bdiag}}(\hat
{\mathcal B}_\mathbf U) ].
$
\end{definition}

Note that
$\mathcal C=\mathcal A\diamond_{\mathbf U}\mathcal B\in\mathbb{C}^{n_1\times l\times n_3}$ is equivalent to
$\bar {\mathcal C}_{\mathbf U}=\bar {\mathcal A}_{\mathbf U}\bar {\mathcal B}_{\mathbf U}$,
which can be shown as follows:
First, by the definition of $\mathbf U$-product, we have
\begin{equation*}%加*表示不对公式编号
\begin{aligned}
\mathcal C=\mathcal A \diamond_{\mathbf U}\mathcal B
&=\mathbf U^{T}[\text{fold}_{3}(\text{bdiag}(\hat{\mathcal A}_\mathbf U)\cdot \text{bdiag}(\hat{\mathcal B}_\mathbf U))]\\
&=\mathbf U^{T}[\text{fold}_{3}(\bar{\mathcal A}_{\mathbf U}\cdot \bar{\mathcal B}_{\mathbf U})].
\end{aligned}
\end{equation*}
Moreover,
\begin{equation*}%加*表示不对公式编号
\begin{aligned}
{\bar{\mathcal C}}_{\mathbf U}
&=\text{bdiag}({\hat{\mathcal C}}_{\mathbf U})
=\text{bdiag}(\mathbf U[\mathcal {C}])\\
&=\text{bdiag}(\mathbf U[\mathbf U^{T}[\text{fold}_{3}(\bar{\mathcal A}_{\mathbf U}\cdot \bar{\mathcal B}_{\mathbf U})]])\\
&=\text{bdiag}([\text{fold}_{3}(\bar{\mathcal A}_{\mathbf U}\cdot \bar{\mathcal B}_{\mathbf U})])
=\bar{\mathcal A}_{\mathbf U}\bar{\mathcal B}_{\mathbf U}.
\end{aligned}
\end{equation*}

\begin{definition}\cite[Definition 2]{Song_2020}
The conjugate transpose of a tensor
$\mathcal A\in \mathbb {C}^{n_1\times n_2\times n_3}$
with respect to any unitary matrix $\mathbf U \in \mathbb {C}^ {n_3\times n_3}$ is a tensor
$\mathcal A^{T}\in \mathbb {C}^{n_2\times n_1\times n_3}$ given by
	$\mathcal A^{T}=\mathbf U^{T}[\text{\rm{{fold}}}_{3}(\text{\rm{bdiag}}(\hat {\mathcal A}_\mathbf U)^T)].$
\end{definition}

\begin{definition}\cite[Proposition 4.1]{Kernfeld2015}
Let each frontal slice of  $\mathcal T\in \mathbb{R}^{n\times n \times n_3}$ be the
$n\times n$ identity matrix.
The identity tensor $\mathcal I\in \mathbb {C}^{n\times n\times n_3}$
with respect to $\mathbf U$ is defined as $\mathcal I=\mathbf U^T[\mathcal T]$.
\end{definition}

\begin{definition}\cite[Definition 4.3]{Kernfeld2015}
The inverse of a tensor $\mathcal A\in \mathbb {C}^{n\times n\times n_3}$ with respect to $\mathbf U$,
denoted by $\mathcal A^{-1}$, is defined as
	$\mathcal A^{-1}\diamond_{\mathbf U}\mathcal A =\mathcal A\diamond_{\mathbf U}\mathcal A^{-1}=\mathcal I.$
\end{definition}

\begin{definition}\cite[Definition 5.1]{Kernfeld2015} If a tensor $\mathcal P\in \mathbb {C}^{n\times n\times n_3}$ with respect to $\mathbf U$-product satisfies $\mathcal
P^{T}\diamond_{\mathbf U}\mathcal P=\mathcal P\diamond_{\mathbf U}\mathcal P^{T}=\mathcal I$, then $\mathcal P$ is a unitary tensor.
\end{definition}

A diagonal tensor is defined that each frontal slice of the tensor is a diagonal matrix \cite{Kilmer2011}.
Now we give the definition of the transformed tensor SVD of  a third-order tensor.

\begin{definition}\cite[Theorem 5.1]{Kernfeld2015}
The transformed tensor SVD of $\mathcal A \in \mathbb {C}^{n_1\times n_2\times n_3}$ is defined as
	$\mathcal A= \mathcal U \diamond_{\mathbf U}\Sigma\diamond_{\mathbf U}\mathcal V^T ,$
where $\Sigma \in \mathbb {C}^{n_1\times n_2\times n_3}$ is a diagonal tensor,
$\mathcal U \in \mathbb {C}^{n_1\times n_1\times n_3},\mathcal V \in \mathbb {C}^{n_2\times n_2\times n_3}$ are unitary tensors with respect to $\mathbf U$-product.
\end{definition}
\begin{definition}\cite[Definition 6]{Song_2020}
For an arbitrary tensor $\mathcal A\in\mathbb{C}^{n_1\times n_2\times n_3}$,
its transformed multi-rank
is a vector  $\textbf{r}=(r_1,r_2,\ldots,r_{n_3}) \in \mathbb {R}^{n_3}$
with $r_i=\textup{rank}({\hat{\mathcal A}}_\mathbf U^{(i)}),i=1,2,\ldots,n_3$.
Let $\mathcal A= \mathcal U \diamond_{\mathbf U}\Sigma\diamond_{\mathbf U}\mathcal V^T $.
The transformed tubal rank $\hat{r}$ of  ${\mathcal A}$ is defined as the number of nonzero tubes of $\Sigma$,
i.e.,
$\hat{r}=\#\{i:\Sigma(i,i,:)\neq \mathbf{0}\}=\mathop{\text{\rm{max}}}\{r_1,r_2,\ldots,r_{n_3}\},$
where $\#\{\cdot\}$ denotes the cardinality of a set.
\end{definition}

\section{Transformed Tensor-Tensor Product for Image Alignment}\label{chap3}

In this section, we first formulate the problem of batch image alignment based on the tensor structure, and
then propose a tensor factorization model via transformed tensor-tensor product under any unitary transformation
for a batch of linearly correlated image alignment,
where the observed images are deformed by some unknown domain transformations and corrupted
by additive Gaussian noise and sparse noise simultaneously.

\subsection{Problem Formulation}

Given $n_3$ well-aligned images $\mathbf{L}_{i}\in\mathbb {R}^{n_1\times n_2}$
of the same object, $i=1,2,\ldots,n_3$.
These images $\mathbf{L}_{i}$ are stacked
 into a third-order tensor $\mathcal L\in \mathbb {R}^{n_1\times n_2\times n_3}$,
 whose $i$th frontal slice  is the image $\mathbf{L}_{i}$.
As a result, the tensor data $\mathcal L$ generated by the linearly correlated image has a low rank structure.
However, these images with same objects or scenes
may not be well-aligned,
which can be modeled
 as domain deformations, such as affine transformation,
 similarity transformation, translation transformation \cite{ma2004invitation}.
Assume that the domain transformations $\tau_i\in\mathbb{R}^m$ are from a finite-dimensional group $\mathbb {G}$.
Then we apply the transformation $\tau_i$ to each image given by
$$
(\mathcal D\circ\tau)(:,:,i)=\mathcal D(:,:,i)\circ\tau_{i},
$$
where $\tau=\{\tau_{1},\tau_{2},\ldots,\tau_{n_3}\}$ is the domain transformation,
$\mathcal{D}$ denotes the misaligned images,
and $\mathcal D\circ\tau \in \mathbb {R}^{n_1\times n_2\times n_3}$
represents the well-aligned images via the domain transformation $\tau$ \cite{Peng_2012, Zhang_2021}.
However, in real-world applications, the observed images not only suffer from misalignment,
but also are corrupted by additive Gaussian noise and sparse noise simultaneously.
Mathematically, the observed model can be described as
\begin{equation}\label{ObserM}
\mathcal D\circ\tau=\mathcal{L}+\mathcal S+\mathcal N,
\end{equation}
where $\tau$ represents the domain transformation, $\mathcal{L}$ denotes the clear and well-aligned images,
$\mathcal S$ and $\mathcal N$ denote the sparse noise and Gaussian noise, respectively.
Then we aim  to recover $\mathcal{L}$ from the noisy and misaligned observations $\mathcal{D}$.

\subsection{Tensor Factorization Model via Transformed Tensor-Tensor Product}

In this subsection,
we propose a tensor factorization method based on
$\mathbf{U}$-product for aligning a batch of linearly correlated images,
which can handle large-scale image data effectively.

For the underlying tensor $\mathcal{L}$ in (\ref{ObserM}), we propose to utilize the
transformed tensor-tensor product to explore its global low-rankness,
where the underlying tensor can be decomposed  into the $\mathbf{U}$-product of two small tensors under any unitary transformation.
In this situation, the transformed tubal rank of the underlying tensor is low.
More precisely,
assume that the transformed tubal rank of $\mathcal L\in \mathbb {R}^{n_1\times n_2\times n_3}$
is $r$.
Then $\mathcal L$ is factorized into the
$\mathbf{U}$-product of two small tensors $\mathcal X\in \mathbb {R}^{n_1\times r\times n_3}$
and $\mathcal Y\in \mathbb {R}^{r\times n_2\times n_3}$ such that
$
\mathcal L=\mathcal X\diamond_{\mathbf U}\mathcal Y.
$
In this case, if $r$ is small, the transformed tubal rank of $\mathcal{L}$ is low.
Then the
computational complexity of $\mathbf{U}$-product is much lower than that of
the transformed tensor nuclear norm method in \cite{Qiu_2021}.
Moreover, the tensor $\ell_p$ norm is utilized
to characterise the sparseness of the sparse noise, where $0<p<1$.
Compared with the tensor $\ell_1$ norm, the tensor $\ell_p$ norm can obtain a sparser solution \cite{Marjanovic_2012}.
For Gaussian noise, we use the tensor Frobenius norm
of the difference among $\mathcal D\circ\tau$, the underlying tensor with $\mathbf{U}$-product, and sparse noise.
Therefore, we propose the following model for a batch of image alignment:
\begin{equation}\label{test10} 	
	\mathop{\text{min}}\limits_{\tau,\mathcal X ,\mathcal Y,\mathcal S} \frac{1}{2} {\| \mathcal D\circ\tau-\mathcal X \diamond_{\mathbf U} \mathcal Y-\mathcal
S \|}_{F}^{2}+\lambda{\|\mathcal S \|}^{p}_{p},
\end{equation}
where $\lambda>0$ is a  given constant.

\begin{remark}
In model (\ref{test10}), we propose to utilize the transformed tensor-tensor
product via any unitary transformation to explore the low-rankness of these images,
which can reduce the computational complexity compared with the
transformed tensor nuclear norm method in \cite{Qiu_2021} if the transformed tubal rank is small.
And the superiority of the transformed tensor-tensor product is that the lower transformed multi-rank will be obtained under
suitable unitary transformation compared with the tensor-tensor product based on Fourier transform in \cite{Kilmer2011}.
Moreover, the tensor $\ell_p$ norm ($0<p<1$) is used to enhance the sparsity of the error tensor,
which can generate a sparser solution as compared to the tensor $\ell_1$ norm.
This is similar to the $\ell_p$ optimization problem in compressed sensing and matrix completion \cite{Marjanovic_2012}.
\end{remark}

\begin{remark}
For the matrix based methods in \cite{Peng_2012, Chen_2016} for image alignment,
one needs to unfold the images into  vectors and stack these vectors as columns of a matrix,
which is low-rank since the linear correlation of the images.
However, the unfolding method will destroy the internal structure of these images
and only utilizes the correlation among different images.
The tensor based method can  explore the  low-rankness better
since both the correlation among these images and the correlation of each image are used.
\end{remark}

\begin{remark}
Compared with the model in  \cite{Zhang_2021}, which
used $\ell_p$ norm based the sum of nuclear norms of unfolding matrices of a tensor to approximate the sum of entries of its Tucker rank,
we used the transformed tensor-tensor product to approximate the transformed tubal rank of a tensor.
Moreover, the sum of nuclear norms of unfolding matrices of a tensor is not the tightest convex
relaxation of the sum of entries of the Tucker rank of a tensor \cite{RomeraParedes2013}, which implies that
the sum of nuclear norms of the unfolding matrices of a tensor is just suboptimal.
Besides, the transformed tensor SVD based methods perform better than the Tucker rank based methods in low-rank tensor recovery,
see  \cite{Qiu_2021,qiu2021nonlocal, zhang2021low, ng2020patched} and references therein.
\end{remark}

\begin{remark}
Compared with the model in \cite{Qiu_2021}, which used the transformed
tensor nuclear norm based method to characterise the low-rankness of the underlying tensor,
 we employ the $\mathbf{U}$-product to enhance the low-rankness of the underlying tensor.
 The main superiority of $\mathbf{U}$-product is that lower computational complexity is required,
which avoids the SVD of the frontal slices of the underlying tensor in the transformed domain.
\end{remark}

\section{Optimization Algorithm}\label{OptAlgGGN}

In this section, a generalized  Gauss-Newton algorithm \cite{Jittorntrum1980}
is developed to solve the proposed model (\ref{test10}),
where a proximal Gauss-Seidel algorithm {\cite{Attouch_2011} is designed to solve the
resulting subproblem. Then, the convergence and convergence rate of the proximal Gauss-Seidel algorithm are established
under very mild conditions.

\subsection{Generalized Gauss-Newton Algorithm}\label{chap4.1}

By the nonlinearity of the formula $\mathcal D\circ\tau$, one cannot solve problem (\ref{test10}) directly.
We develop a generalized Gauss-Newton algorithm to solve (\ref{test10}),
which is based on a linear approximation to the nonlinear term in the objective.
When the change in $\tau$ is small,
we linearize the formula $\mathcal D\circ\tau$ by the first-order Taylor expansion.
A similar technique has been widely used in the existing literature \cite{Peng_2012,Zhang_2021,Qiu_2021}.
Suppose that the initial transformation $\tau^t\in\mathbb G$ is
known and $\mathbb G$ is some $m\mbox{-}$parameter group. For $\Delta\tau_i\in \mathbb {R}^{m}\text{ and }\Delta\tau=[\Delta\tau_1,\Delta\tau_2,\ldots,\Delta\tau_{n_3}]\in \mathbb {R}^{m\times n_3}$, we have
\begin{equation}\label{test200}%加*表示不对公式编号
\begin{aligned}
\mathcal D\circ\tau
&\approx\mathcal D\circ(\tau^t+\Delta\tau)\\
&\approx \mathcal D\circ\tau^t+
\text{Fold}_3\Big(\Big(\sum\limits_{i=1}^{n_3}\mathbf J_{i}\Delta\tau\varepsilon_{i}\varepsilon_{i}^{T}\Big)^{T}\Big),
\end{aligned}
\end{equation}
where $\mathbf J_{i}$ represents the Jacobian of the $i$th image
with respect to the domain transformation parameter $\tau_{i}^t$, $\varepsilon_{i}$
denotes the standard basis in $\mathbb R^{n_3}, i=1,2,\ldots, n_3$, and $\textup{Fold}_3(\cdot)$
is the inverse operator of mode-$3$ unfolding of a third-order tensor.
In this case, problem (\ref{test10}) can be relaxed to the following nonconvex optimization problem:
\begin{equation}\label{test11}%加*表示不对公式编号
\begin{aligned}
\mathop{\text{min}}\limits_{\mathcal X ,\mathcal Y,\mathcal S,\Delta\tau} \frac{1}{2} {\big \| \mathcal D\circ\tau^t +\Delta\Gamma-\mathcal X \diamond_{\mathbf U} \mathcal Y-\mathcal S\big\|}_{F}^{2}+\lambda{\|\mathcal S \|}_{p}^{p}\\
+\frac{\rho_{1}}{2}{\|\mathcal X \|}_{F}^{2}+\frac{\rho_{2}}{2}{\|\mathcal Y \|}_{F}^{2}+\frac{\rho_{3}}{2}{\|\Delta\tau\|}_{F}^{2},
\end{aligned}
\end{equation}
where $\rho_1,\rho_2,\rho_3>0$ are given constants and
$
 \Delta\Gamma=\text{Fold}_3((\sum_{i=1}^{n_3}\mathbf J_{i}\Delta\tau\varepsilon_{i}\varepsilon_{i}^{T})^{T}).
$
Here we add a quadratic term $\frac{\rho_{3}}{2}{\|\Delta\tau\|}_{F}^{2}$
to prevent $\Delta\tau$ from being too large since the Taylor expansion of $\tau$ is only locally valid.
Moreover, the quadratic terms with respect to $\mathcal{X}$ and $\mathcal{Y}$
are used to guarantee the uniqueness of the corresponding solutions.

When one get the solution of $\Delta\tau$ by solving model (\ref{test11}),
$\tau^t$  is updated by $\tau^t+\Delta\tau$ in the next iteration,
which is the main idea of the generalized Gauss-Newton algorithm \cite{Jittorntrum1980}
and has been widely applied in image alignment \cite{Peng_2012,Qiu_2021, Wu_2012,Zhang_2021}.
The iterative process of the generalized Gauss-Newton algorithm
is stated in Algorithm \ref{alg:algorithm1}.
Moreover, in order to eliminate trivial solutions, such as zooming in a single dark pixel
or dark area in the image,
we normalize the image data by
$\frac{\mathcal D(:,:,i)\circ\tau_{i}^t}{{\|\mathcal D(:,:,i)\circ\tau_{i}^t\|}_{F}}$
at the $t$th iteration.
%In Algorithm \ref{alg:algorithm1}, the algorithm iterates until the relative change of the value of the cost function between two consecutive iterations is smaller than the predetermined threshold, or the number of iterations reaches a certain value.
\begin{algorithm}[htbp]
\caption{A Generalized Gauss-Newton Algorithm for Solving (\ref{test10})}\label{alg:algorithm1}\label{alg:algorithm1}
\begin{algorithmic}
\STATE
\STATE {\textsc{\bf{Initialization:}}} Image data $\mathcal D\in \mathbb {R}^{n_1\times n_2\times n_3} $.
        Given $\lambda>0 $ and an initial transformation $\tau^0\in\mathbb{G}$. For $ t=0,1,2,\ldots$
\STATE {\textsc{\bf{repeat}}}
\STATE \hspace{0.5cm}$\textbf{Step 1.}$ Calculate the Jacobian matrices:
         {\small $$
		\mathbf J_{i}=\frac{\partial}{\partial\tau_{i}^{t}}\left(\frac{\mathcal D(:,:,i)\circ\tau_{i}^{t}}{{\|\mathcal D(:,:,i)\circ\tau_{i}^{t}\|}_{F}}\right)\bigg |_{\tau_{i}^{t}},i=1,2,\ldots,n_3.% \State 后写一般语句
         $$}\vspace{-3mm}\label{Step1}
\STATE \hspace{0.5cm} \textbf{Step 2.} Warp and normalize $\mathcal D\circ\tau^{t}$ :
         {\small $$
         (\mathcal D\circ\tau^{t})(:,:,i)=\frac{\mathcal D(:,:,i)\circ\tau_{i}^{t}}{{\|\mathcal D(:,:,i)\circ\tau_{i}^{t}\|}_{F}}.
         $$} \vspace{-3mm}
\STATE \hspace{0.5cm} \textbf{Step 3.} Compute $(\mathcal X, \mathcal Y, \mathcal S, \Delta\tau)$ by solving (\ref{test11}).\label{Step3}
\STATE \hspace{0.5cm} \textbf{Step 4.} Update $\tau^{t+1}=\tau^{t}+\Delta\tau$.
\STATE {\textsc{\bf{until}}} A stopping condition is satisfied.
\end{algorithmic}
\end{algorithm}
\subsection{Proximal Gauss-Seidel Algorithm}\label{chap4.2}
The key step of  Algorithm \ref{alg:algorithm1} is to solve problem (\ref{test11}).
Now we design a proximal Gauss-Seidel algorithm {\cite{Attouch_2011} to compute it.
Let
\begin{equation}\label{test23}
\begin{aligned}
F(\mathcal X ,\mathcal Y,\mathcal S,\Delta\tau):= Q(\mathcal X ,\mathcal Y,\mathcal S,\Delta\tau)+\lambda{\|\mathcal S \|}^{p}_{p}\\+\frac{\rho_{1}}{2}{\|\mathcal X \|}_{F}^{2}
+\frac{\rho_{2}}{2}{\|\mathcal Y\|}_{F}^{2}+\frac{\rho_{3}}{2}{\|\Delta\tau\|}_{F}^{2},
 \end{aligned}
\end{equation}
where
\begin{equation}\label{eq11}
 Q(\mathcal X ,\mathcal Y,\mathcal S,\Delta\tau):=\frac{1}{2} {\left \| \mathcal D\circ\tau^t +\Delta\Gamma-\mathcal X \diamond_{\mathbf U} \mathcal Y-\mathcal S\right \|}_{F}^{2}.
\end{equation}
Then the iterations of the proximal Gauss-Seidel algorithm
for solving (\ref{test11}) can be given explicitly as follows:
{\small \begin{align}
&\mathcal X^{k+1}=\mathop{\text{argmin}}\limits_{\mathcal X} F(\mathcal X ,\mathcal Y^{k},\mathcal S^{k},\Delta\tau^{k})+\frac{\alpha^{k}}{2}{\|\mathcal X-\mathcal {X}^{k} \|}_{F}^{2}, \label{eq12}
             \\[-3mm]%调整多行公式间距
&\mathcal Y^{k+1}=\mathop{\text{argmin}}\limits_{\mathcal Y} F(\mathcal X^{k+1} ,\mathcal Y,\mathcal S^{k},\Delta\tau^{k})+\frac{\beta^{k}}{2}{\|\mathcal Y-\mathcal {Y}^{k} \|}_{F}^{2},\label{eq13}
             \\[-3mm]
&\mathcal S^{k+1}=\mathop{\text{argmin}}\limits_{\mathcal S} F(\mathcal X^{k+1} ,\mathcal Y^{k+1},\mathcal S,\Delta\tau^{k})+\frac{\gamma^{k}}{2}{\|\mathcal S-\mathcal {S}^{k} \|}_{F}^{2}, \label{eq14}
             \\[-3mm]
&\Delta\tau^{k+1}=\mathop{\text{argmin}}\limits_{\Delta\tau} F(\mathcal X^{k+1} ,\mathcal Y^{k+1},\mathcal S^{k+1},\Delta\tau)+\frac{\xi^{k}}{2}{\| \Delta\tau-{\Delta\tau}^{k} \|}_{F}^{2}, \label{eq15}
\end{align}}%
where $\alpha^{k}, \beta^{k}, \gamma^{k}, \xi^{k} >0$ are the given parameters.
Now  the explicit solutions of each subproblem in (\ref{eq12})-(\ref{eq15}) are given below.

\begin{itemize}
\item Problem (\ref{eq12}) can be reformulated equivalently  as
\begin{eqnarray}\label{test17}	
{\mathcal X}^{k+1}=\mathop{\text{argmin}}\limits_{\mathcal X} \frac{1}{2} {\left \|{\mathcal
B}^{k}-{\mathcal X}\diamond_{\mathbf U}{\mathcal Y}^{k}\right \|}_{F}^{2}
\nonumber\\
+\frac{\rho_{1}+\alpha^{k}}{2}\Big \|{\mathcal
X}-\frac{\alpha^{k}}{\rho_{1}+\alpha^{k}}{\mathcal X}^{k} \Big \|_{F}^{2},
\end{eqnarray}
where
\begin{equation}\label{eq25}
\mathcal B^{k}=\mathcal D\circ\tau^t +\Delta\Gamma^k-{\mathcal S}^{k}.
\end{equation}
with $\Delta\Gamma^k=\text{Fold}_3((\sum_{i=1}^{n_3}\mathbf J_{i}\Delta\tau^k\varepsilon_{i}\varepsilon_{i}^{T})^{T})$.
Then the optimal solution $\mathcal X^{k+1}$ of (\ref{test17}) is given by
\begin{eqnarray}\label{eq17}
{\mathcal X}^{k+1}=\big(\mathcal B^{k}\diamond_{\mathbf U}({\mathcal Y}^{k})^{T}+\alpha^{k}{\mathcal X}^{k}\big)\diamond_{\mathbf U}\big({\mathcal Y}^{k}\diamond_{\mathbf U}({\mathcal Y}^{k})^{T}\nonumber
\\+(\rho_{1}+\alpha^{k})\mathcal I\big)^{-1},
\end{eqnarray}
where $\mathcal I \in \mathbb {C}^{r\times r\times n_3}$ denotes the identity tensor.

\item Problem (\ref{eq13}) can be rewritten as
\begin{equation}\label{test19}
\begin{aligned}	
{\mathcal Y}^{k+1}=\mathop{\text{argmin}}_{{\mathcal Y}} \frac{1}{2} \big \|{\mathcal
B}^{k}-{\mathcal X}^{k+1}\diamond_{\mathbf U}{\mathcal Y}\big \|_{F}^{2}
\\+\frac{\rho_{2}
+\beta^{k}}{2}\Big \|{\mathcal
Y}-\frac{\beta^{k}}{\rho_{2}+\beta^{k}}{\mathcal Y}^{k} \Big \|_{F}^{2},
 \end{aligned}
\end{equation}
where $\mathcal B^{k}$  is defined as (\ref{eq25}). The optimal solution  $\mathcal Y^{k+1}$ of (\ref{test19}) takes the form of
\begin{eqnarray} \label{eq19}	
{\mathcal Y}^{k+1}=\big(({\mathcal X}^{k+1})^{T}\diamond_{\mathbf U}{\mathcal X}^{k+1}+(\rho_{2}+\beta^{k})\mathcal I\big)^{-1}\nonumber \\
\diamond_{\mathbf U}\big(({\mathcal X}^{k+1})^{T}\diamond_{\mathbf U}\mathcal B^{k}+\beta^{k}{\mathcal Y}^{k}\big).
\end{eqnarray}
\item Problem (\ref{eq14}) is equivalent to
\begin{equation}\label{test20}
\mathcal S^{k+1}= \mathop{\text{argmin}}\limits_{\mathcal S}\lambda{\left \|\mathcal S \right \|}^{p}_{p}+\frac{1+\gamma^{k}}{2}\left\|\mathcal S- \mathcal {H} ^k \right \|_{F}^{2},
\end{equation}
where
 $$
 \mathcal {H} ^k=\frac{1}{1+\gamma_{k}}\left(\mathcal D\circ\tau^t+\Delta\Gamma^k- \mathcal X^{k+1} \diamond_{\mathbf U}\mathcal Y^{k+1}+\gamma^{k}\mathcal S^k\right).
 $$
%\begin{eqnarray}\label{test20}	
%\mathcal S=\mathop{\text{argmin}}\limits_{\mathcal S}\lambda{\left \|\mathcal S \right \|}^{p}_{p,p}+\frac{1+\gamma_{k}}{2} {\left \|\mathcal S -\mathcal {H} ^k \right \|}_{F}^{2},
%\end{eqnarray}
Note that problem (\ref{test20}) is just the thresholding of the $\ell_p$ optimization problem ($0<p<1$) for the tensor case.
By \cite[Theorem 1]{Marjanovic_2012},
one can deduce
{\small \begin{eqnarray} \label{test21}
(\mathcal S^{k+1})_{ijt}=\begin{cases}
0,&\text{if }  |(\mathcal {H} ^k)_{ijt}|<\mu^{*}, \\
\{0,\text{sgn}((\mathcal {H} ^k)_{ijt})\bar{b}\},&\text{if }  |(\mathcal {H} ^k)_{ijt}|=\mu^{*},\\
\text{sgn}((\mathcal {H} ^k)_{ijt})\bar{b}^*,&\text{if }  |(\mathcal {H} ^k)_{ijt}|>\mu^{*},
\end{cases}
\end{eqnarray}}%
where $\text{sgn}(\cdot)$ denotes the signum function,
$
\bar{b}=(\frac{2\lambda(1-p)}{1+\gamma^k})^{\frac{1}{2-p}},  \mu^{*}=\bar{b}+\frac{\lambda p{\bar{b}}^{p-1}}{1+\gamma^k},
$
and $\bar{b}^*$ is the largest solution
to the equation $b+\frac{\lambda}{1+\gamma^k} p{b}^{p-1}=|(\mathcal {H} ^k)_{ijt}|$ with $b>0$,
which can be obtained from the iteration
\begin{equation}\label{equbs}
b^{s+1}=|(\mathcal {H} ^k)_{ijt}|-\frac{\lambda}{1+\gamma^k} p({b^s})^{p-1}
\end{equation}
with	
the initial value $b^0\in (\bar{b},|(\mathcal {H} ^k)_{ijt}|), s=0,1,2,\ldots$
\item Let $\mathcal C^{k+1}=\mathcal D\circ\tau^t-\mathcal X^{k+1} \diamond_{\mathbf U}\mathcal Y^{k+1}-\mathcal S^{k+1}$. Then  problem (\ref{eq15}) can be reformulated equivalently as
\begin{eqnarray*} \label{test22}
\Delta\tau^{k+1}=\mathop{\text{argmin}}\limits_{\Delta\tau}
\frac{\rho_{3}+\xi^{k}}{2}\Big \|\Delta\tau-\frac{\xi^{k}}{\rho_{3}+\xi^{k}}{\Delta\tau}^{k}\Big \|_{F}^{2}\\
+
\frac{1}{2} \Big \| \text{Fold}_3\big(\big(\sum\limits_{i=1}^{n_3}\mathbf J_{i}\Delta\tau\varepsilon_{i}\varepsilon_{i}^{T}\big)^{T}\big)
+\mathcal C^{k+1}\Big \|_{F}^{2}.
\end{eqnarray*}
It follows from \cite[Theorem 3]{Qiu_2021} that
\begin{eqnarray}\label{eq22}
\Delta\tau^{k+1}=\sum\limits_{i=1}^{n_3}\left(\mathbf J_{i}^{T}\mathbf J_{i}+(\rho_{3}+\xi^{k})\mathbf I\right)^{-1}\nonumber\left(\xi^{k}{\Delta\tau}^{k}\right.\\
\left. -\mathbf J_{i}^{T} (\mathcal{C}_{(3)}^{k+1})^{T}\right)\varepsilon_{i}\varepsilon_{i}^{T},
\end{eqnarray}
where $\mathbf I$ denote the $m\times m$ identity matrix.
\end{itemize}

Now the proximal Gauss-Seidel algorithm for solving problem (\ref{test11}) is summarized in Algorithm \ref{alg:algorithm2}.
\begin{remark}
The parameters $\alpha^k,\beta^k,\gamma^k,\xi^k$ in Algorithm \ref{alg:algorithm2}
are just needed to have lower and upper bounds in the implementation,
which is used to guarantee the convergence of Algorithm \ref{alg:algorithm2}.
\end{remark}
\begin{remark}
The computational complexity of Algorithm \ref{alg:algorithm2} is
given as follows.
For any two tensors $\mathcal A\in \mathbb {R}^{n_1\times r\times n_3}, \mathcal B\in \mathbb {R}^{r\times n_2\times n_3}$, and unitary matrix $\mathbf U\in \mathbb {C}^{n_3\times n_3}$,
the complexity of performing a unitary matrix ($n_3\times n_3$) to an $n_3$-tube is $O(n_3^2)$.
The complexity of performing the matrix-matrix product of the frontal slices is $O(rn_1n_2)$ after unitary transformations.
Therefore, the computational complexity of $\mathcal{A}\diamond_{\mathbf U}\mathcal{B}$ is $O(rn_1n_2n_3+\max\{n_1,n_2\}rn_3^2)$.
The computational complexity of the inverse of an $n\times n\times n_3$ tensor is $O(n^2n_3^2+n^3n_3)$.
Assume that $r\leq \min\{n_1,n_2\}$ and $m\leq n_3$.
Then the computational complexity of computing ${\mathcal X}^{k+1}$ is $O(rn_1n_2n_3+n_1n_2n_3^2)$.
Similarly, the computational complexity of computing ${\mathcal Y}^{k+1}$ is $O(rn_1n_2n_3+n_1n_2n_3^2)$.
Moreover, the computational cost of ${\mathcal S}^{k+1}$ is $O(n_1n_2n_3n_s)$,
where $n_s$ is the maximum number of iterations for solving (\ref{equbs}).
And the computational cost of $\Delta\tau^{k+1}$ is $O(mn_1n_2n_3+mn_3^2)$.
As a result, the computational complexity of the proximal Gauss-Seidel algorithm is $O(rn_1n_2n_3+n_1n_2n_3^2+ n_1n_2n_3n_s+mn_3^2)$.
\end{remark}
\begin{algorithm}[htbp]
\caption{A Proximal Gauss-Seidel Algorithm for Solving Problem (\ref{test11}).}
    \label{alg:algorithm2}
    \begin{algorithmic}%[1]有没有[1]区别在于前面有没有数字标号
\STATE {\textsc{\bf{Initialization:}}} Given ${\mathcal X}^0,{\mathcal Y}^0,\mathcal S^0,\Delta\tau^0$, $\lambda,\rho_{1},\rho_{2},\rho_{3}>0,\underline d \leq \alpha^k,\beta^k,\gamma^k,\xi^k  \leq \overline d,$ $0 < \underline d < \overline d < +\infty.$
\STATE {\textsc{\bf{repeat}}}
\STATE \hspace{0.5cm} \textbf{Step 1.} Update ${\mathcal X}^{k+1}$ by (\ref{eq17}).\label{step21}
\STATE \hspace{0.5cm} \textbf{Step 2.} Update ${\mathcal Y}^{k+1}$ by (\ref{eq19}).
\STATE \hspace{0.5cm} \textbf{Step 3.} Update ${\mathcal S}^{k+1}$ by (\ref{test21}).
\STATE \hspace{0.5cm} \textbf{Step 4.} Update $\Delta\tau^{k+1}$   by (\ref{eq22}).
\STATE {\textsc{\bf{until}}} A stopping condition is satisfied.
\end{algorithmic}
\end{algorithm}
\begin{remark}
The main difference between \cite{Qiu_2021} and our proposed algorithm in Algorithm \ref{alg:algorithm2} is
the computations of $\mathcal{X}^{k+1}, \mathcal{Y}^{k+1}$ and the transformed tensor nuclear norm.
Assume that $n_2\leq n_1$, then the computational complexity of the transformed tensor nuclear norm
in \cite{Qiu_2021} is $O(n_1n_2^2n_3+n_1n_2n_3^2)$, while the computational cost
of $\mathcal{X}^{k+1}, \mathcal{Y}^{k+1}$ is $O(rn_1n_2n_3+n_1n_2n_3^2)$.
Therefore, if $r$ is much smaller than $n_2$, the computational
complexity of Algorithm \ref{alg:algorithm2} is much lower than that in \cite{Qiu_2021}.
Note that $r$ is much smaller than $\min\{n_1, n_2\}$
in general since a low-rank requirement is preferred for a batch of linearly correlated image alignment.
\end{remark}
\subsection{Convergence Analysis}\label{chap4.3}
In this subsection, the convergence of the proximal Gauss-Seidel algorithm is established, which is stated in the following theorem.

\begin{theorem}\label{thm1}
Supposed that
$0<\underline d<\overline d< +\infty $ are given constants.
Assume that $\underline d \leq \alpha^k,\beta^k,\gamma^k,\xi^k  \leq \overline d $ and $p$ is rational.
Let $\{(\mathcal X^{k} ,\mathcal Y^{k},\mathcal S^{k},\Delta\tau^{k})\}$
be the sequence generated by Algorithm \ref{alg:algorithm2}.
Then the sequence $\{(\mathcal X^{k} ,\mathcal Y^{k},\mathcal S^{k},\Delta\tau^{k})\}$
converges to a critical point of $F(\mathcal X ,\mathcal Y,\mathcal S,\Delta\tau)$ in (\ref{test23}).
\end{theorem}

\begin{remark}
The assumptions in Theorem \ref{thm1} are very mild.
One only needs $p$ to be rational and $\alpha^k,\beta^k,\gamma^k,\xi^k$ to
be bounded from below and above for any $k$.
\end{remark}

\subsection{Rate of Convergence}
In this subsection, we establish the rate of convergence of the proximal
Gauss-Seidel algorithm for solving problem (\ref{test11}).
Before proceeding with the analysis, we introduce some properties of the sequence
$\{(\mathcal X^{k} ,\mathcal Y^{k},\mathcal S^{k},\Delta\tau^{k})\}$ generated by Algorithm \ref{alg:algorithm2}.
To simplify the notation for the following analysis,
we denote $\mathcal Z^{k}=(\mathcal X^{k} ,\mathcal Y^{k},\mathcal S^{k},\Delta\tau^{k})$.

The KL exponent, which is defined in Appendix A,}
%Appendix,
is closely related to the convergence rate
of many descent algorithms \cite{attouch2009convergence, Li2017}. Note that  the proper closed semi-algebraic
functions are KL functions with exponent $\theta\in[0,1)$
(e.g., \cite{Attouch_2010,Li2017,Yu2021}), which implies that
$F(\mathcal X ,\mathcal Y,\mathcal S,\Delta\tau)$ in (\ref{test23}) is a KL function
with exponent $\theta\in[0,1)$.
In the following, we state the main result with respect to the convergence
rate of the proximal Guass-Seidel algorithm in Algorithm \ref{alg:algorithm2}.
%\textcolor{red}{The proof is given in the Supplementary Material.}
%The detailed proof of Theorem \ref{thm4} are provided in the supplementary material.
\begin{theorem}\label{thm4}
Let $\{\mathcal Z^{k}\}$ be a sequence generated by Algorithm  \ref{alg:algorithm2},
whose limit point is denoted by $\mathcal Z^{*}$,
and $\theta\in[0,1)$ be  the KL exponent of $F$.
Suppose that $p$ is rational.
Then the following assertions hold:

(i)  If $\theta=0$, then the sequence $\{\mathcal Z^{k}\}$  converges in a finite number of steps.\par
(ii)  If $\theta\in(0,\frac{1}{2}]$, then the sequence $\{\mathcal Z^{k}\}$  converges R-linearly, i.e.,
there exist $c >0$ and $\vartheta\in (0,1)$ such that
$
\|\mathcal Z^{k}-\mathcal Z^{*}\|_{F} \leq c \vartheta^k.
$
\par
(iii)  If $\theta\in(\frac{1}{2},1)$, then the sequence $\{\mathcal Z^{k}\}$  converges R-sublinearly,
i.e., there exists $c >0$ such that
$
\|\mathcal Z^{k}-\mathcal Z^{*}\|_{F} \leq c k^{-\frac{1-\theta}{2\theta-1}}.
$
 \end{theorem}

\begin{remark}
When  the KL exponent of $F$ in (\ref{test23}) can be computed,
the detailed convergence rate of the proximal Gauss-Seidel algorithm can be obtained.
\end{remark}

\section{Numerical Experiments}\label{NNMEX}
In this section, we conduct numerical experiments on several image datasets to
verify the effectiveness and efficiency  of the tensor factorization method
via transformed tensor-tensor product (TFM-TTP) for image alignment,
and compare with some state-of-the-art methods,
including robust alignment by
sparse and low-rank decomposition\footnote{\footnotesize \url{http://people.eecs.berkeley.edu/~yima/matrix-rank/rasl.html}} (RASL) \cite{Peng_2012},
transformed Grassmannian
robust adaptive subspace tracking algorithm\footnote{\footnotesize \url{http://sites.google.com/site/hejunzz/t-grasta}} (t-GRASTA) \cite{He_2014},
nonconvex plus quadratic penalized low-rank and
sparse decomposition (NQLSD) \cite{Chen_2016},
alternating direction method of multipliers with $\ell_p$ norm ($\ell_p$-ADMM) \cite{Zhang_2021},
and nonconvex image alignment by the
low transformed multi-rank tensor plus sparsity (NCALTS) method \cite{Qiu_2021}.
All experiments are performed on MATLAB (R2018b) with an Intel Core i7-10750H 2.6GHz and 16GB RAM.

\subsection{Stopping Criteria}\label{chap4.4}

For the inner iterations of Algorithm \ref{alg:algorithm2}, we adopt the distance
$\text{dist}(0,\partial F(\mathcal X^{k+1} ,\mathcal Y^{k+1},\mathcal S^{k+1},\Delta\tau^{k+1}))$
to measure the accuracy of the optimal solution.
However, it is hard to compute it directly.
In light of Lemma 1(i) in Appendix C,
we can find an upper bound of
$\text{dist}(0,\partial F(\mathcal X^{k+1} ,\mathcal Y^{k+1},\mathcal S^{k+1},\Delta\tau^{k+1}))$
to stop  Algorithm \ref{alg:algorithm2} in practice. Define
$
\nu:=\|\mathcal V_\mathcal X^{k+1}+\nabla_{\mathcal X}Q^{k+1}\|_{F}+\|\mathcal V_\mathcal Y^{k+1}+\nabla_{\mathcal Y}Q^{k+1}\|_{F}
+\|\mathcal V_\mathcal S^{k+1}+\nabla_{\mathcal S}Q^{k+1}\|_{F}
+\|\xi^{k}(\Delta\tau^{k}-{\Delta\tau}^{k+1})\|_{F}
$.
Let $\textup{tol}$ be a given tolerance.
Algorithm \ref{alg:algorithm2} will be terminated if
{\small$$
\frac{\nu}{1+{\left\|\mathcal
D\circ\tau^t\right\|}_{F}+\|\mathcal X^{k+1}\|_{F}+\|\mathcal Y^{k+1}\|_{F}+\|\mathcal S^{k+1}\|_{F}}\leq\text{tol},
$$}%
or the maximum number of iterations reaches $200$. %(\ref{test106})
In the experiments, we set $\textup{tol}$ to be $1\times10^{-2}$ in all experiments except for the handwritten number ``8'' with mixed noise, which is set to be $5\times10^{-2}$.
%By (\ref{test110}), we use an upper bound
%of $\text{dist}(0,\partial F(\mathcal X^{k+1} ,\mathcal Y^{k+1},\mathcal S^{k+1},\Delta\tau^{k+1}))$
%to stop  Algorithm \ref{alg:algorithm2} in practice.
%Let $\textup{tol}$ be a given tolerance and $F'=\|\mathcal V_\mathcal X^{k+1}+\nabla_{\mathcal X}Q^{k+1}\|_{F}+\|\mathcal V_\mathcal Y^{k+1}+\nabla_{\mathcal Y}Q^{k+1}\|_{F}+\|\mathcal V_\mathcal S^{k+1}+\nabla_{\mathcal S}Q^{k+1}\|_{F}$
%Based on (\ref{test109}), Algorithm \ref{alg:algorithm2} will be terminated if
%\begin{center}
%${\bar w}^{k+1}=
%\frac{\|\mathcal V_\mathcal X^{k+1}+\nabla_{\mathcal X}Q^{k+1}\|_{F}+\|\mathcal V_\mathcal Y^{k+1}+\nabla_{\mathcal Y}Q^{k+1}\|_{F}+\|\mathcal V_\mathcal S^{k+1}+\nabla_{\mathcal S}Q^{k+1}\|_{F}+\|\xi^{k}(\Delta\tau^{k}-{\Delta\tau}^{k+1})\|_{F}}{1+{\left\|\mathcal
%D\circ\tau^t\right\|}_{F}+\|\mathcal X^{k+1}\|_{F}+\|\mathcal Y^{k+1}\|_{F}+\|\mathcal S^{k+1}\|_{F}}\leq\text{tol}$,
%\end{center}

For the outer iterations, we stop
 Algorithm \ref{alg:algorithm1} until the relative change of
 the value of the cost function between two consecutive iterations is smaller than $1\times 10^{-2}$,
 or the maximum number of iterations reaches $60$.
\subsection{Implementation Details}
\subsubsection{Parameters setting}
As suggested by \cite[Theorem 2]{Song_2020} for robust tensor completion,
we set $\lambda=a/{\sqrt{\max\{n_1,n_2\}n_3}}$,
where $a$ is chosen from the range $[0.0005,0.2]$ to get the best possible results.
%We adjust the parameter $a$.
For the parameters $\rho_i$, we take them the same for different $i=1,2,3$,
which are chosen from the interval $[0.01, 0.4]$ to
get the best recovery images in all experiments.
%We set the regularization parameters $\rho_i(i=1,2,3)$ lying in an interval the penalty parameter $\left[0,1\right]$.$\lambda=\frac{a}{\sqrt{\max(n_1,n_2)n_3}}$
The parameters $\alpha^k,\beta^k,\gamma^k,\xi^k$
of the proximal terms in Algorithm \ref{alg:algorithm2}
are not sensitive to the results and
we set $\alpha^k=\beta^k=\gamma^k=\xi^k=1\times10^{-5}$ in all
experiments.
For the parameter $p$ in the tensor $\ell_p$ norm of sparse noise, the recovery
performance may be influenced for different image datasets,
and we choose it from the set $\{0.1, 0.25, 0.3, 0.4, 0.5, 0.6, 0.65, 0.7, 0.75\}$
to get the best recovery results in the experiments.

\subsubsection{Unitary transform matrix in $\mathbf U$-product}
%\subsubsection{Unitary transform matrix}
The selection of the unitary transform matrix $\mathbf{U}$
in the transformed tensor-tensor product plays a vital role  for TFM-TTP.
Similar to \cite{Qiu_2021,qiu2021nonlocal, Song_2020, zhang2021low, song2020tensor, ng2020patched},
we adopt the following approach to obtain the data-dependent unitary  matrix.
First, we get an initial estimator  $\hat {\mathcal L}$ by model (\ref{test10})
based on the discrete cosine transform
in the $\mathbf{U}$-product. Then $\hat {\mathcal L}$ is
unfolded along the third dimension, i.e., $\hat {\mathcal L}_{(3)}={\mathbf L}$.
Assume that the SVD of $\mathbf L$ is given
by $\mathbf L = \bar{\mathbf U} \bar{\mathbf \Sigma} \bar{\mathbf V}^T$.
Then we choose $\bar{\mathbf U}^T$ as the unitary matrix in the
$\mathbf U$-product. In this case, one can get lower transformed multi-rank of a tensor
by using the data-dependent transformation.
More details about the construction of the
unitary matrix in the $\mathbf{U}$-product can be found in \cite{ng2020patched, Song_2020}.

\begin{table*}[!t]\scriptsize
\centering
    \setlength{\abovecaptionskip}{0cm} % 调整间距
    \setlength{\belowcaptionskip}{-0.25cm}
	\caption{\label{tab:table1} Reconstruction errors and CPU time (in seconds) of different methods
for the AR Face database corrupted by salt-and-pepper impulse noise with noise level 30\%.}
\label{tab:chap:table_1}
	%\begin{tabular}
       %\resizebox{\linewidth}{!}{
		\begin{tabular}{cccccccccccccc}
        \bottomrule
        \specialrule{0em}{0.5pt}{0.5pt}
        \multirow{2}{*}{s}&\multirow{2}{*}{$\alpha$}&\multicolumn{2}{c}{RASL}&\multicolumn{2}{c}{t-GRASTA}&\multicolumn{2}{c}{NQLSD}
            &\multicolumn{2}{c}{$\ell_p$-ADMM}&\multicolumn{2}{c}{NCALTS}&\multicolumn{2}{c}{TFM-TTP}\\
            \specialrule{0em}{0.5pt}{0.5pt}
            \cline{3-14}
			%Ion&1st alternative&2nd alternative&lst alternative
			%&2nd alternative\\ \hline
            && error  & time & error & time &  error & time & error  & time & error & time &  error & time \\ \hline
            \specialrule{0em}{1.2pt}{1.2pt}
    3&3&0.1491&\underline{6.11} &0.1465	&54.35&0.1394&15.53&0.1377&145.64&\underline{0.1336}&26.82 &\textbf{0.1311}&{\bf 5.93}\\
     &6&0.1524&13.40&0.1513 &44.71&0.1403&\underline{11.94}&0.1401&196.23&\underline{0.1339}&38.51 &\textbf{0.1326}&{\bf 6.15}\\
     &9&0.1568&26.57&0.1531 &63.74&0.1449&\underline{17.13}&0.1459&205.85&\underline{0.1366}&51.33 &\textbf{0.1344}&{\bf 6.05}\\
            \hline
            \specialrule{0em}{1pt}{1pt}
    5&3&0.1820&\underline{6.47}&0.1811 &69.13&0.1747    &{\bf 5.31}&0.1775&48.10&\underline{0.1662}&21.62&\textbf{0.1654}&6.51\\
     &6&0.1847&\underline{17.21}&0.1820 &59.53&0.1770&19.92 &0.1770&60.09&\underline{0.1684}&31.97&\textbf{0.1681}&{\bf 17.00} \\
     &9&0.1908&\underline{26.29}&0.1907 &51.19&0.1848&29.55&0.1814&53.43 &\underline{0.1749}&40.06&\textbf{0.1741}&{\bf 15.48} \\
          \hline
          \specialrule{0em}{1pt}{1pt}
    7&3&0.1981&11.81&0.1971&19.29&0.1906&{\bf 5.67}&0.1930&48.28&\underline{0.1807}&17.24 &\textbf{0.1791}&\underline{10.89}\\
     &6&0.2035&24.02&0.2020&41.32&0.1961&{\underline {18.45}}&0.1970&93.77& \underline{0.1872}&33.52&\textbf{0.1866}&\bf{17.00}\\
     &9&0.2078&40.24&0.2048&80.87&\underline{0.2003}&\underline{9.42}&0.2029&47.38& \textbf{0.1895}&26.26&\textbf{0.1895}&{\bf 9.39}\\
     \specialrule{0em}{0.1pt}{0.1pt}
     \bottomrule
	\end{tabular}%}
\end{table*}

%figure6
\begin{figure*}[!t]
\subfigcapskip=3pt%子标题与图片的距离
\setlength{\abovecaptionskip}{-0.5pt}%图片与注释之间的距离
	\centering
	\subfigure[Original]{
		\begin{minipage}[t]{0.12\textwidth}
			\centering
			\vspace{0.5mm}
			\includegraphics[width=2.2cm,height=3cm]{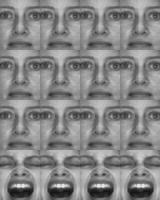}
		\end{minipage}
	}\hspace{-3mm}
	\subfigure[Observation]{
		\begin{minipage}[t]{0.12\textwidth}
			\centering
			\vspace{0.5mm}
			\includegraphics[width=2.2cm,height=3cm]{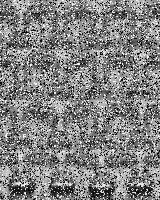}
		\end{minipage}
	}\hspace{-3mm}
	\subfigure[RASL]{
		\begin{minipage}[t]{0.12\textwidth}
			\centering
			{\scriptsize CPU: 6.11}\vspace{0.5mm}
			\includegraphics[width=2.2cm,height=3cm]{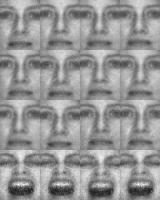}
		\end{minipage}
	}\hspace{-3mm}
	\subfigure[t-GRASTA]{	
		\begin{minipage}[t]{0.12\textwidth}
			\centering
			{\scriptsize{CPU: 54.35}}\vspace{0.5mm}
			\includegraphics[width=2.2cm,height=3cm]{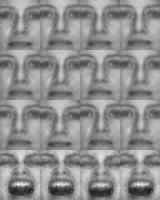}
		\end{minipage}
	}\hspace{-3mm}
     \subfigure[NQLSD]{
		\begin{minipage}[t]{0.12\textwidth}
			\centering
			{\scriptsize{CPU: 15.53}}\vspace{0.5mm}
			\includegraphics[width=2.2cm,height=3cm]{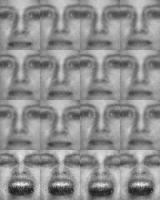}
		\end{minipage}
	}\hspace{-3mm}
	\subfigure[$\ell_p$-ADMM]{
		\begin{minipage}[t]{0.12\textwidth}
			\centering
			{\scriptsize{CPU:145.64}}\vspace{0.5mm}
			\includegraphics[width=2.2cm,height=3cm]{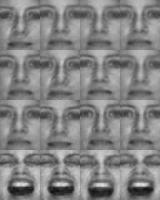}
		\end{minipage}
	}\hspace{-3mm}
	\subfigure[NCALTS]{
		\begin{minipage}[t]{0.12\textwidth}
			\centering
			{\scriptsize{CPU: 26.82}}\vspace{0.5mm}
			\includegraphics[width=2.2cm,height=3cm]{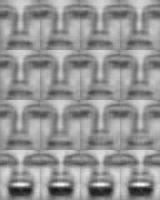}
		\end{minipage}
	}\hspace{-3mm}
	\subfigure[TFM-TTP]{	
		\begin{minipage}[t]{0.12\textwidth}
			\centering
			{\scriptsize{CPU: 5.93}}\vspace{0.5mm}
			\includegraphics[width=2.2cm,height=3cm]{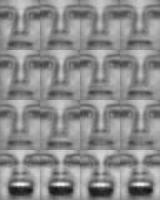}
		\end{minipage}
	}
\caption{Recovered images and CPU time (in seconds) of different methods
for the AR Face database corrupted
by salt-and-pepper impulse noise with noise level 30\%,
where $ s=3, \alpha=3$.}
	\label{Fig: AR salt}
\end{figure*}
\begin{table*}[!t]
\scriptsize
\centering
    \setlength{\abovecaptionskip}{0cm} % 调整间距
    \setlength{\belowcaptionskip}{-0.25cm}
	\caption{\label{tab:table1} Reconstruction errors and CPU time (in seconds) of different methods
 for the AR Face database corrupted by salt-and-pepper impulse noise with noise level 20\% and Gaussian noise with zero-mean and variance $\sigma^2=0.01$.}
	%\begin{tabular}
\label{tab:chap:table_2}
      % \resizebox{\linewidth}{!}{
		\begin{tabular}{cccccccccccccc}
        \bottomrule			
        \specialrule{0em}{0.5pt}{0.5pt}
        \multirow{2}{*}{s}&\multirow{2}{*}{$\alpha$}&\multicolumn{2}{c}{RASL}&\multicolumn{2}{c}{t-GRASTA}&\multicolumn{2}{c}{NQLSD}
            &\multicolumn{2}{c}{$\ell_p$-ADMM}&\multicolumn{2}{c}{NCALTS}&\multicolumn{2}{c}{TFM-TTP}\\
            \specialrule{0em}{0.5pt}{0.5pt}
            \cline{3-14}
			%Ion&1st alternative&2nd alternative&lst alternative
			%&2nd alternative\\ \hline
     && error  & time & error & time &  error & time & error  & time & error & time &  error & time \\
          \hline
          \specialrule{0em}{1.2pt}{1.2pt}
    3&3&0.1448&5.81 &0.1383	&83.97&0.1295&\underline{5.56} &0.1271&29.84&\underline{0.1248}&20.55 &\bf{0.1206}&\bf{5.02}\\
     &6&0.1468&12.81&0.1427 &81.05&0.1342&\underline{8.71} &0.1292&45.90&\underline{0.1265}&42.38 &\bf{0.1231}&\bf{8.30}\\
     &9&0.1523&24.84&0.1512 &80.88&0.1385&\underline{13.77}&0.1353&32.01&\underline{0.1295}&56.54 &\bf{0.1264}&\bf{11.56}\\
          \hline
          \specialrule{0em}{1pt}{1pt}
    5&3&0.1705&17.12&0.1693 &221.99&0.1633&\underline{10.15}&0.1639&38.22&\underline{0.1541}&26.69&\bf{0.1532}&\bf{7.03}\\
     &6&0.1754&           18.03 &0.1763 &615.73&0.1665&\underline{13.94}&0.1651&59.00&\underline{0.1570}&33.67&\bf{0.1567}&\bf{7.46}\\
     &9&0.1791&           29.49 &0.1798 &108.36&0.1732&\underline{27.49} &0.1714&36.99&\underline{0.1616}&73.96&\bf{0.1605}&\bf{24.72}\\
          \hline
          \specialrule{0em}{1pt}{1pt}
    7&3&0.1940&24.91&0.1945&40.80&0.1878&\underline{15.56}&0.1894&24.55&\bf{0.1777}&24.91 &\underline{0.1792}&\bf{15.39}\\
     &6&0.1991&\underline{17.43}&0.1982&36.39&0.1925&23.85&0.1922&22.74&\underline{0.1830}&37.38&\bf{0.1829}&\bf{15.17}\\
     &9&0.2036&55.76&0.2049&37.64&0.1981&\underline{31.94}&0.1972&52.91&\underline{0.1899}&73.27&\bf{0.1885}&\bf{29.07}\\
    \specialrule{0em}{0.1pt}{0.1pt}
    \bottomrule
\end{tabular}%}
\end{table*}

\begin{figure*}[!t]
    \subfigcapskip=3pt
    \setlength{\abovecaptionskip}{-0.5pt}%图片与注释之间的距离
	\centering
	\subfigure[Original]{
		%\scriptsize{Arad}
		\begin{minipage}[t]{0.12\textwidth}
			\centering
			\vspace{0.5mm}
			\includegraphics[width=2.2cm,height=3cm]{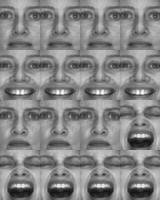}
		\end{minipage}
	}\hspace{-3mm}
	\subfigure[Observation]{
		\begin{minipage}[t]{0.12\textwidth}
			\centering
			\vspace{0.5mm}
			\includegraphics[width=2.2cm,height=3cm]{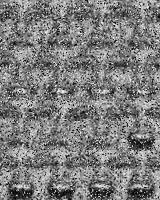}
		\end{minipage}
	}\hspace{-3mm}
	\subfigure[RASL]{
		\begin{minipage}[t]{0.12\textwidth}
			\centering
			{\scriptsize{CPU: 24.84}}\vspace{0.5mm}
			\includegraphics[width=2.2cm,height=3cm]{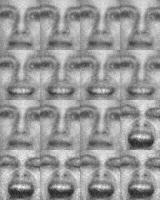}
		\end{minipage}
	}\hspace{-3mm}
	\subfigure[t-GRASTA]{	
		\begin{minipage}[t]{0.12\textwidth}
			\centering
			{\scriptsize{CPU: 80.88}}\vspace{0.5mm}
		\includegraphics[width=2.2cm,height=3cm]{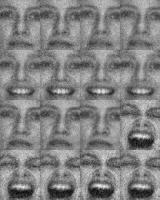}
		\end{minipage}
	}\hspace{-3mm}
        \subfigure[NQLSD]{
		\begin{minipage}[t]{0.12\textwidth}
			\centering
			{\scriptsize{CPU: 13.77}}\vspace{0.5mm}
			\includegraphics[width=2.2cm,height=3cm]{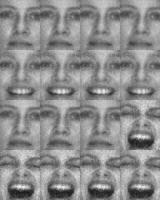}
		\end{minipage}
	}\hspace{-3mm}
	\subfigure[$\ell_p$-ADMM]{
		\begin{minipage}[t]{0.12\textwidth}
			\centering
			{\scriptsize{CPU: 32.01}}\vspace{0.5mm}
			\includegraphics[width=2.2cm,height=3cm]{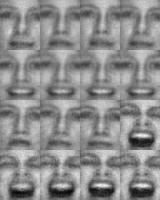}
		\end{minipage}
	}\hspace{-3mm}
	\subfigure[NCALTS]{
		\begin{minipage}[t]{0.12\textwidth}
			\centering
			{\scriptsize{CPU: 56.54}}\vspace{0.5mm}
			\includegraphics[width=2.2cm,height=3cm]{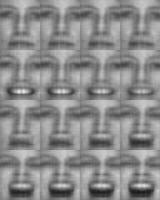}
		\end{minipage}
	}\hspace{-3mm}
	\subfigure[TFM-TTP]{	
		\begin{minipage}[t]{0.12\textwidth}
			\centering
			{\scriptsize{CPU: 11.56}}\vspace{0.5mm}
			\includegraphics[width=2.2cm,height=3cm]{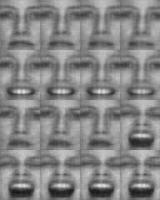}
		\end{minipage}
	}
\caption{ Recovered images and CPU time (in seconds) of different methods for the AR Face database corrupted by salt-and-pepper impulse noise with noise level 20\% and Gaussian noise with zero-mean and variance $\sigma^2=0.01$,
where $ s=3, \alpha=9$.}\label{Fig: AR gauss}
\vspace{-8pt}
\end{figure*}

\subsubsection{Adaptive Transformed Multi-Rank Update}
For the transformed multi-rank of a tensor, we adopt an adaptive update strategy in the experiments.
At the $k$th iteration of Algorithm \ref{alg:algorithm2},
assume that   the tensor
$\mathcal L^k\in \mathbb {R}^{n_1\times n_2\times n_3}$ is decomposed
into the $\mathbf{U}$-product of two smaller tensors
$\mathcal L^k=\mathcal X^k \diamond_{\mathbf U} \mathcal Y^k$,
where $\mathcal X^k\in \mathbb {R}^{n_1\times r\times n_3}$, $\mathcal Y^k\in \mathbb {R}^{r\times n_2\times n_3}$,
and $r$ is the transformed tubal rank of $\mathcal L^k$.
%In general, we do not know the true transformed tubal rank in advance.
Similar to \cite{Yang_2015,Zhou_2018, wen2012solving}, the
rank-decreasing scheme is adopted  to estimate the transformed multi-rank of $\mathcal L^{k+1}$,
whose details are given as follows.

First,
%the tensor $\hat{\mathcal L}_{\mathbf{U}}$ is
%obtained by multiplying the third dimension with the unitary
%transformation matrix $\mathbf{U}$.
suppose that the transformed
multi-rank of $\mathcal L^k$ is $\mathbf{r}^k=(r_1^k,r_2^k,\ldots,r_{n_3}^k)$.
%First, we give an initial overestimated transformed multi-rank $\textbf{r}^0=(r_1^0,r_2^0,\cdots,r_{n_3}^0)$.
%$\mathcal L\in \mathbb {R}^{n_1\times n_2\times n_3}$,
The matrix $(\hat{\mathcal L}_{\mathbf{U}}^{k})^{(i)}\in \mathbb {R}^{n_1\times n_2}$
is factorized into the product of two smaller matrices
$(\hat{\mathbf X}^{k})^{(i)}\in \mathbb {R}^{n_1\times r_i^k}$
and $(\hat{\mathbf Y}^{k})^{(i)}\in \mathbb {R}^{r_i^k \times n_2}$.
%Then $\hat{\mathcal L}_{\mathbf{U}}^{(i)}$ can be written into a matrix product form $\hat{\mathcal L}_{\mathbf{U}}^{(i)}=\hat{\mathbf X}^{(i)}\hat{\mathbf Y}^{(i)}$, where $\hat{\mathbf X}^{(i)}\in \mathbb {R}^{n_1\times r_i^k}$ and $\hat{\mathbf Y}^{(i)}\in \mathbb {R}^{r_i^k \times n_2}$.
Second, we calculate the eigenvalues of
$((\hat{\mathbf X}^{k})^{(i)})^{T}(\hat{\mathbf X}^{k})^{(i)}, i=1,2,\ldots,n_3$,
and sort all eigenvalues of the $n_3$ matrices,
which are denoted by $\lambda_1\geq\lambda_2\geq\cdots\geq\lambda_{n_a}$,
where $n_a=\sum_{i=1}^{n_3}r_i^k$. Finally, we compute the quotients
 $\hat{\lambda}_i=\frac{\lambda_i}{\lambda_{i+1}}, i=1, 2,\ldots,n_a-1$.

Let the index of maximum $\hat{\lambda}_i$ be $m$, i.e.,
$m=\mathop{\text{argmax}}_{1\leq i\leq n_a-1}\hat{\lambda}_i.$
Define
$$
\text{gap}:=\frac{(m-1)\hat{\lambda}_{m}}{\sum_{i\neq m}\hat{\lambda}_{i}}.
$$
If $\text{gap}\geq 10$, it means that there exists a large drop in the
magnitude of eigenvalues. Then one should reduce each entry of $\textbf{r}^k$.
Specifically, we find ${\lambda}_{s^k}$ such that $ \sum_{i=1}^{s^k}{\lambda}_{i}/\sum_{i=1}^{n_a}{\lambda}_{i}\geq 95\%$,
where $s^k$ is the corresponding index of $\lambda_{{s^k}}$.
Suppose that there are $d_i^k$ eigenvalues of $((\hat{\mathbf X}^{k})^{(i)})^{T}(\hat{\mathbf X}^{k})^{(i)}$
belonging to $\{{\lambda}_{s^k+1},\ldots,{\lambda}_{n_a}\}$,
then we set $r_i^k=r_i^k-d_i^k$.
Let the skinny
SVD of $((\hat{\mathbf X}^{k})^{(i)})^{T}(\hat{\mathbf X}^{k})^{(i)}$ be
${\mathbf U}^{(i)}{\mathbf \Sigma}^{(i)}({\mathbf V}^{(i)})^{T}$,
then we update $(\hat{\mathbf X}^{k})^{(i)}={{\mathbf  U}_{r_i^k}^{(i)}}{{\mathbf \Sigma}_{r_i^k}^{(i)}}$
and $(\hat{\mathbf Y}^{k})^{(i)}=({\mathbf V}^{(i)}_{r_i^k})^{T}$,
where ${\mathbf U}_{r_i^k}^{(i)}$ consists of the first $r_i^k$
%is composed of the first $r_i^k$
columns of ${\mathbf U}^{(i)}$, ${\mathbf V}_{r_i^k}^{(i)}$ consists of
%is composed of
the first $r_i^k$ rows of ${\mathbf V}^{(i)}$, and
${{\mathbf \Sigma}_{r_i^k}^{(i)}}$ is a diagonal matrix whose
elements are the largest $r_i^k$ diagonal elements of ${{\mathbf \Sigma}^{(i)}}$.

In the experiments, we set the initial rank estimate $\mathbf{r}^0$ to be
$\text{[}30, \ldots , 30\text{]} \in \mathbb {R}^{n_3}$
for all testing datasets in Algorithm \ref{alg:algorithm2}.

\subsection{Image Alignment with Controlled Images}

Table \ref{tab:chap:table_1} reports the reconstruction errors
and CPU time (in seconds) of different methods with different misaligned levels for the AR Face database
corrupted by salt-and-pepper impulse noise with noise level 30\%,
where the boldface number is the best and the underline number is the second best.
%No matter what the misaligned levels,
We can observe that the TFM-TTP performs better than RASL, t-GRASTA, NQLSD,
and $\ell_p$-ADMM in terms of reconstruction errors for all cases.
For $s=7$ and $\alpha=9$, the reconstruction error achieved
by TFM-TTP is the same as that obtained by NCALTS.
However, the CPU time (in seconds) required by NCALTS are much more than that required by TFM-TTP.
Moreover, the NQLSD and $\ell_p$-ADMM perform better than RASL,
which implies that the nonconvex matrix based methods are superior
to the nuclear norm minimization method in terms of reconstruction errors.
Although the CPU time required by NQLSD is small for few cases,
the reconstruction errors of NQLSD are much larger than those of TFM-TTP.

\begin{figure*}[!t]
\subfigcapskip=3pt%子标题与图片的距离
\setlength{\abovecaptionskip}{-0.5pt}%图片与注释之间的距离
\centering
	\subfigure[Observation]{
		\begin{minipage}[t]{0.13\textwidth}
			\centering
			\vspace{0.5mm}%调整上面字体与图片距离
			\includegraphics[width=2.4cm,height=2.4cm]{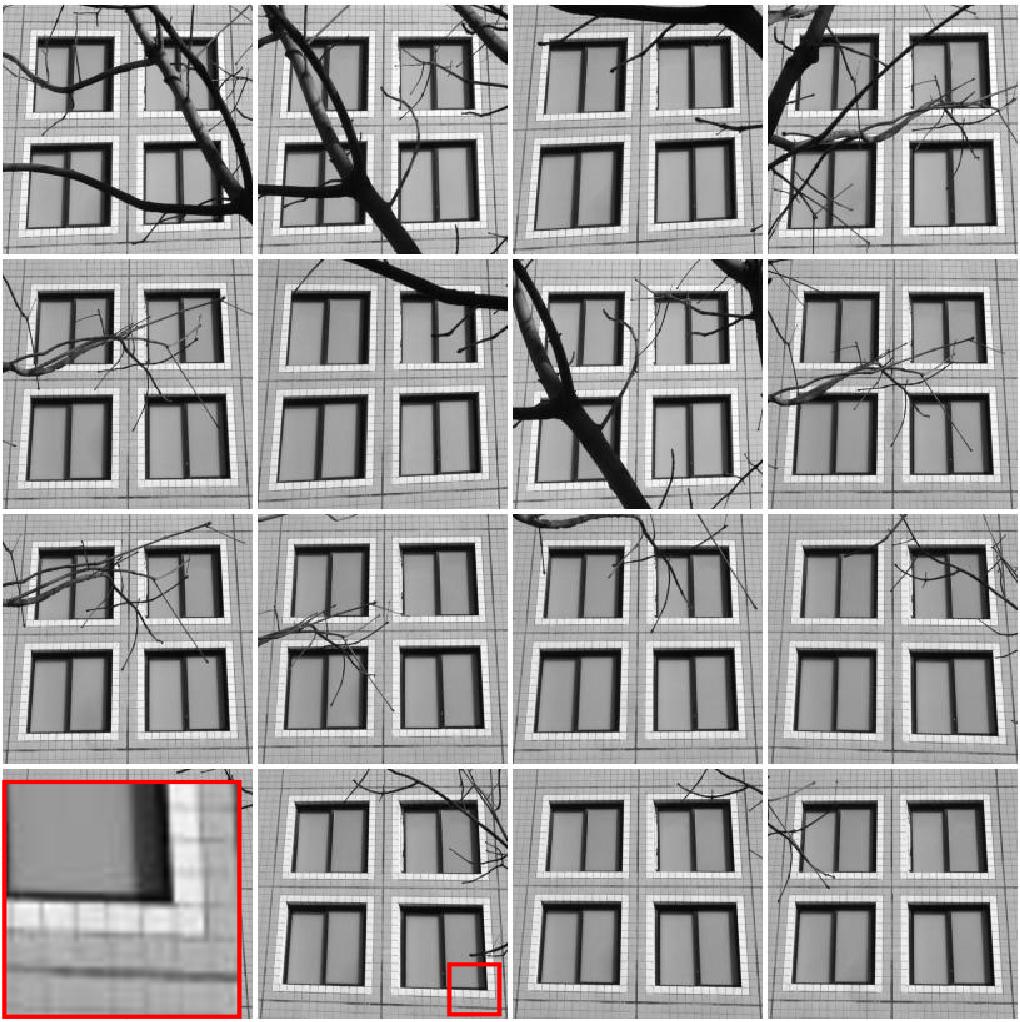}
		\end{minipage}}%\hspace{-4.5mm}
	\subfigure[RASL]{
		\begin{minipage}[t]{0.13\textwidth}
			\centering
			{\scriptsize{CPU: 122.44}}\vspace{0.5mm}
			\includegraphics[width=2.4cm,height=2.4cm]{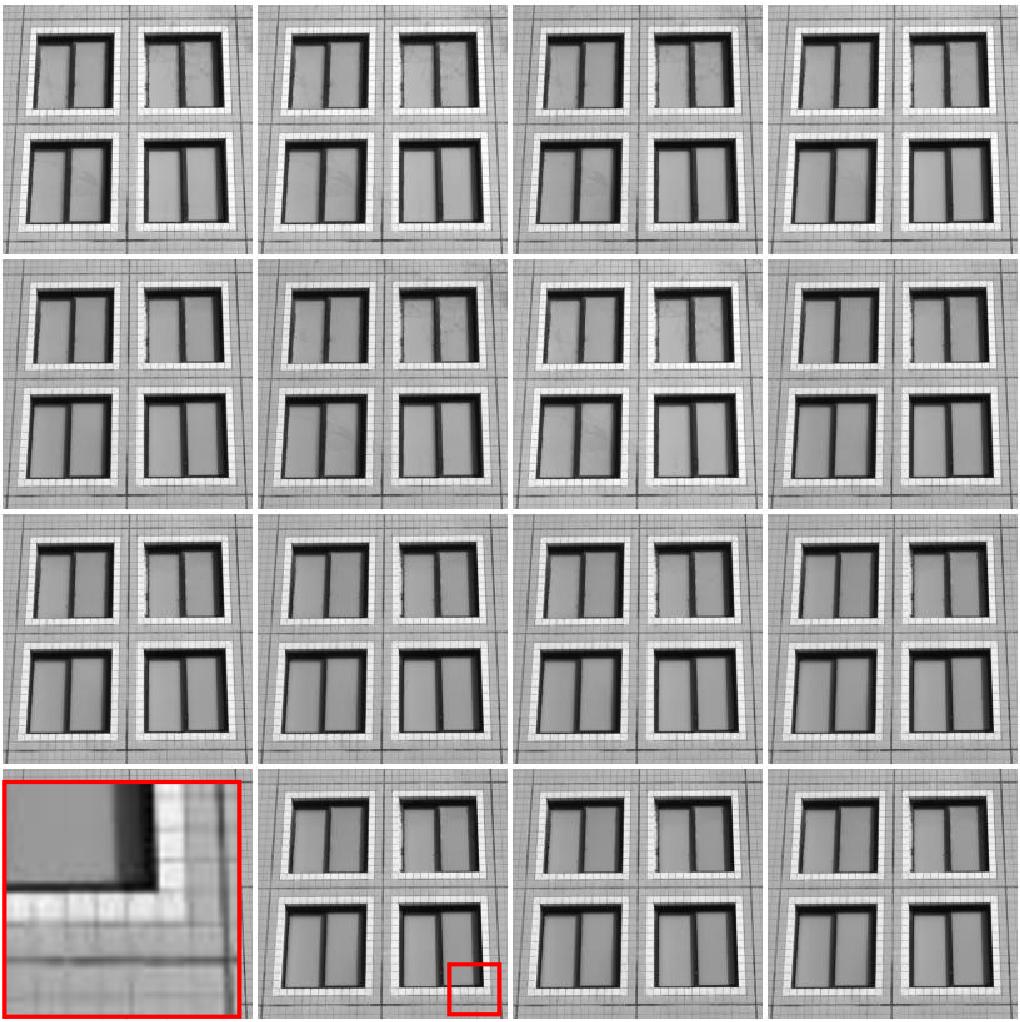}
		\end{minipage}}%\hspace{-4.5mm}
	\subfigure[t-GRASTA]{	
		\begin{minipage}[t]{0.13\textwidth}
			\centering
			{\scriptsize{CPU: 206.66}}\vspace{0.5mm}
			\includegraphics[width=2.4cm,height=2.4cm]{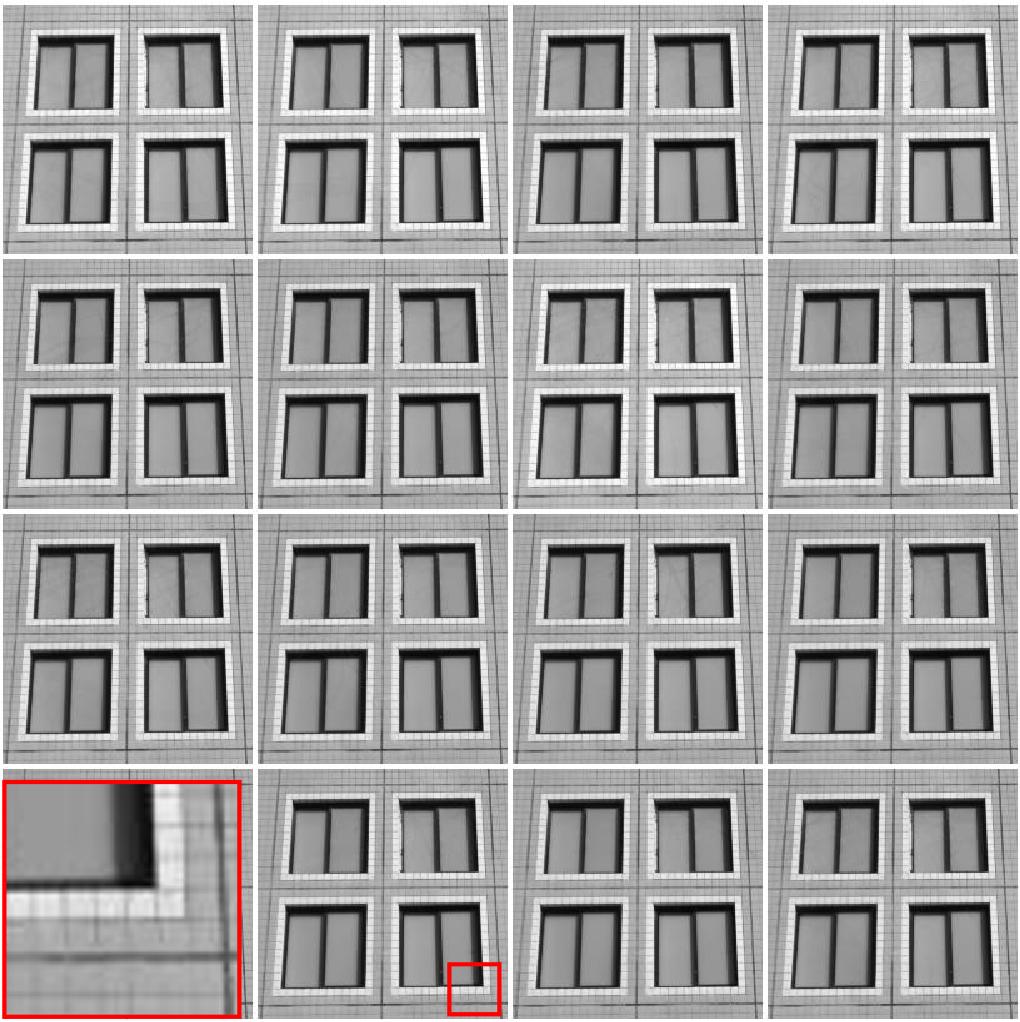}
		\end{minipage}}%\hspace{-4.5mm}
        \subfigure[NQLSD]{
		\begin{minipage}[t]{0.13\textwidth}
			\centering
			{\scriptsize{CPU: 259.48}}\vspace{0.5mm}
			\includegraphics[width=2.4cm,height=2.4cm]{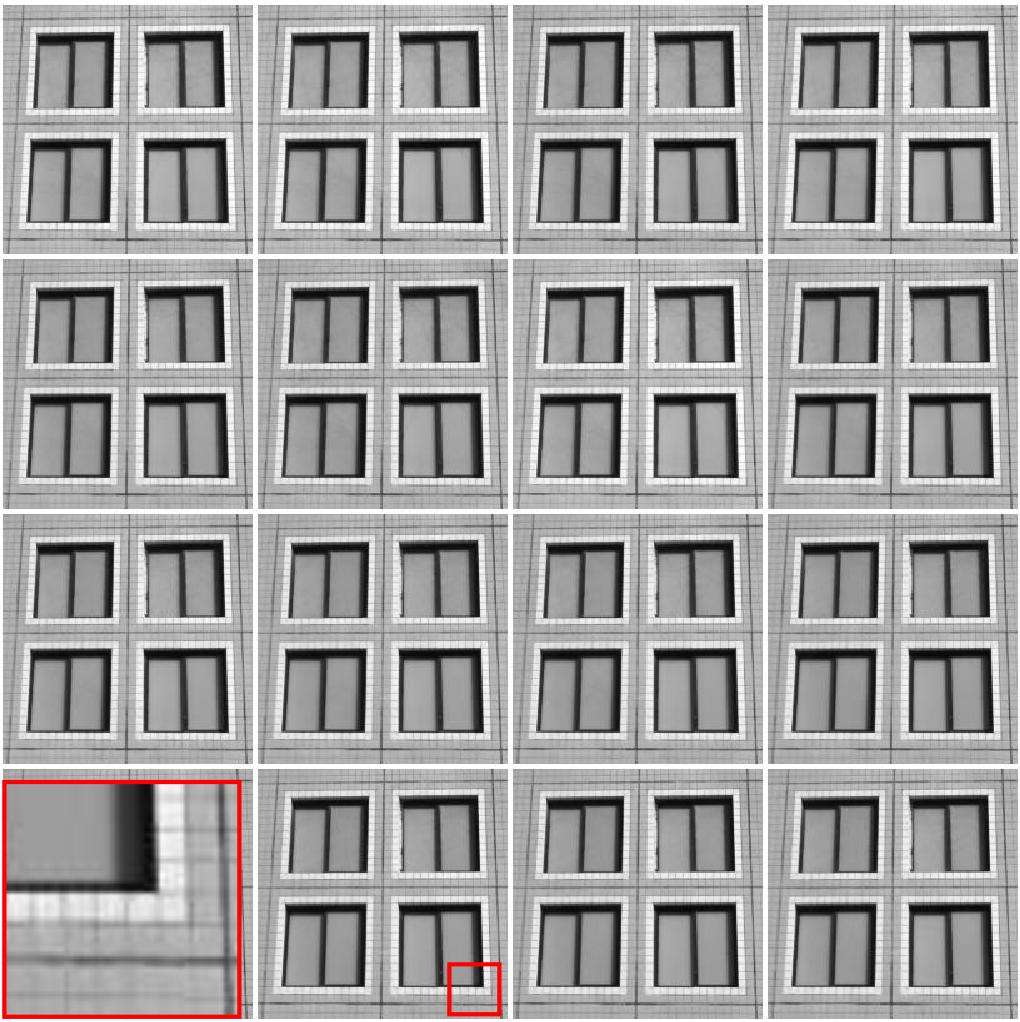}
		\end{minipage}}%\hspace{-4.5mm}
	\subfigure[$\ell_p$-ADMM]{
		\begin{minipage}[t]{0.13\textwidth}
			\centering
			{\scriptsize{CPU: 554.45}}\vspace{0.5mm}
			\includegraphics[width=2.4cm,height=2.4cm]{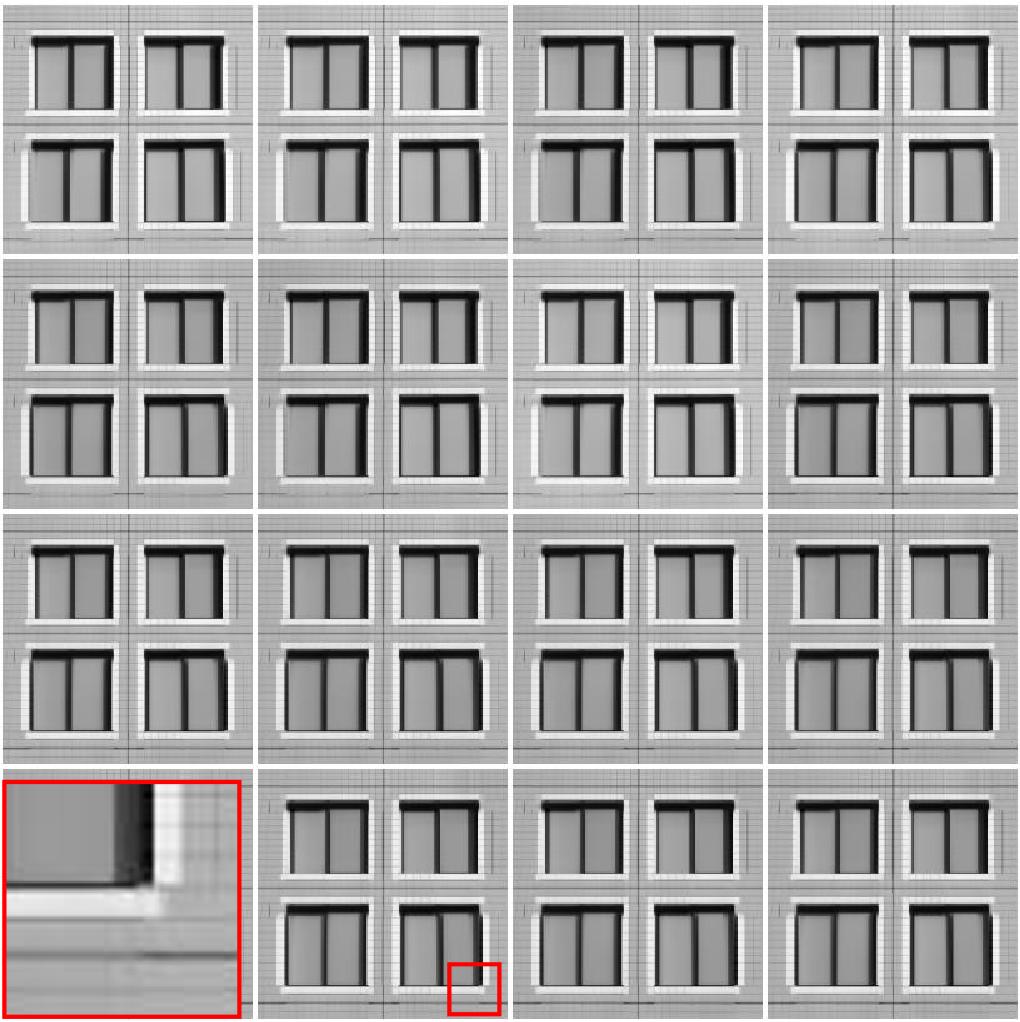}
		\end{minipage}}%\hspace{-4.5mm}
	\subfigure[NCALTS]{
		\begin{minipage}[t]{0.13\textwidth}
			\centering
			{\scriptsize{CPU: 927.35}}\vspace{0.5mm}
			\includegraphics[width=2.4cm,height=2.4cm]{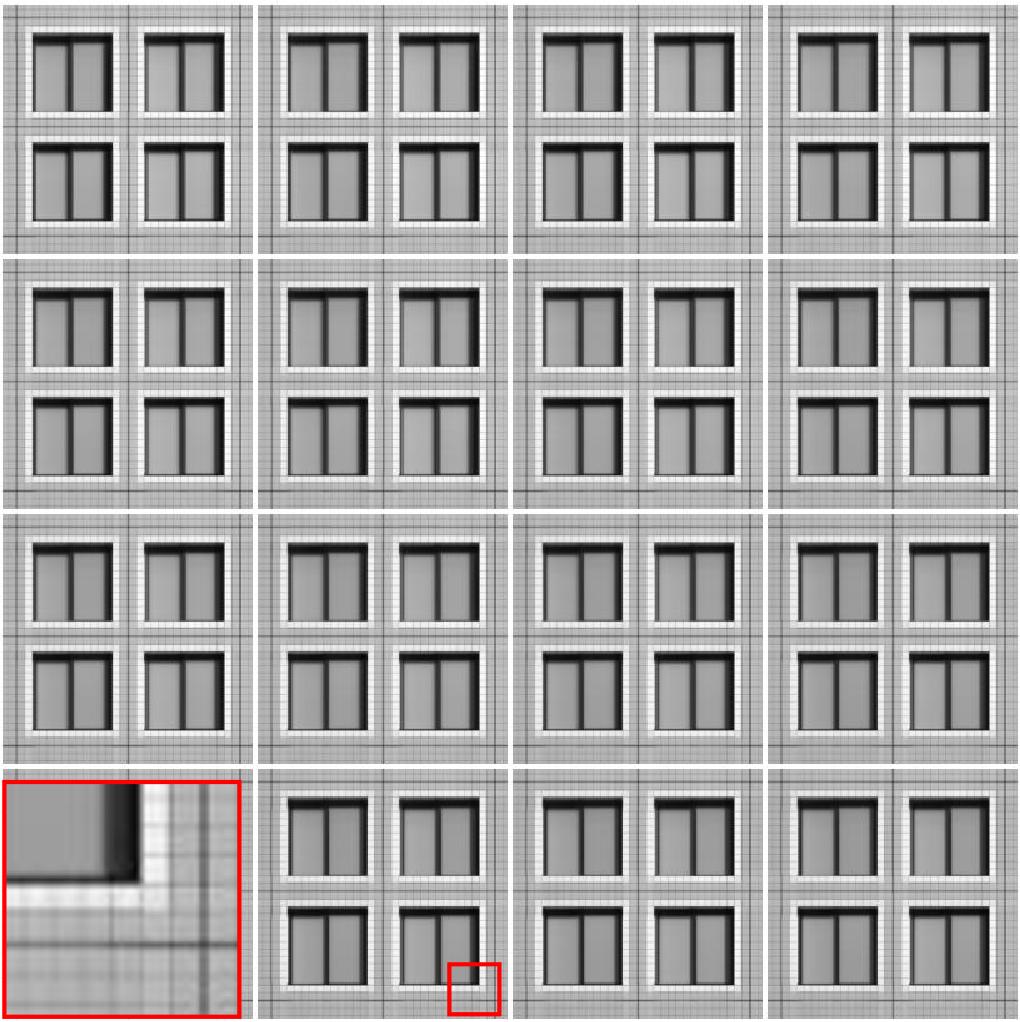}
		\end{minipage}}%\hspace{-4.5mm}
	\subfigure[TFM-TTP]{	
		\begin{minipage}[t]{0.13\textwidth}
			\centering
			{\scriptsize{CPU: 187.79}}\vspace{0.5mm}
			\includegraphics[width=2.4cm,height=2.4cm]{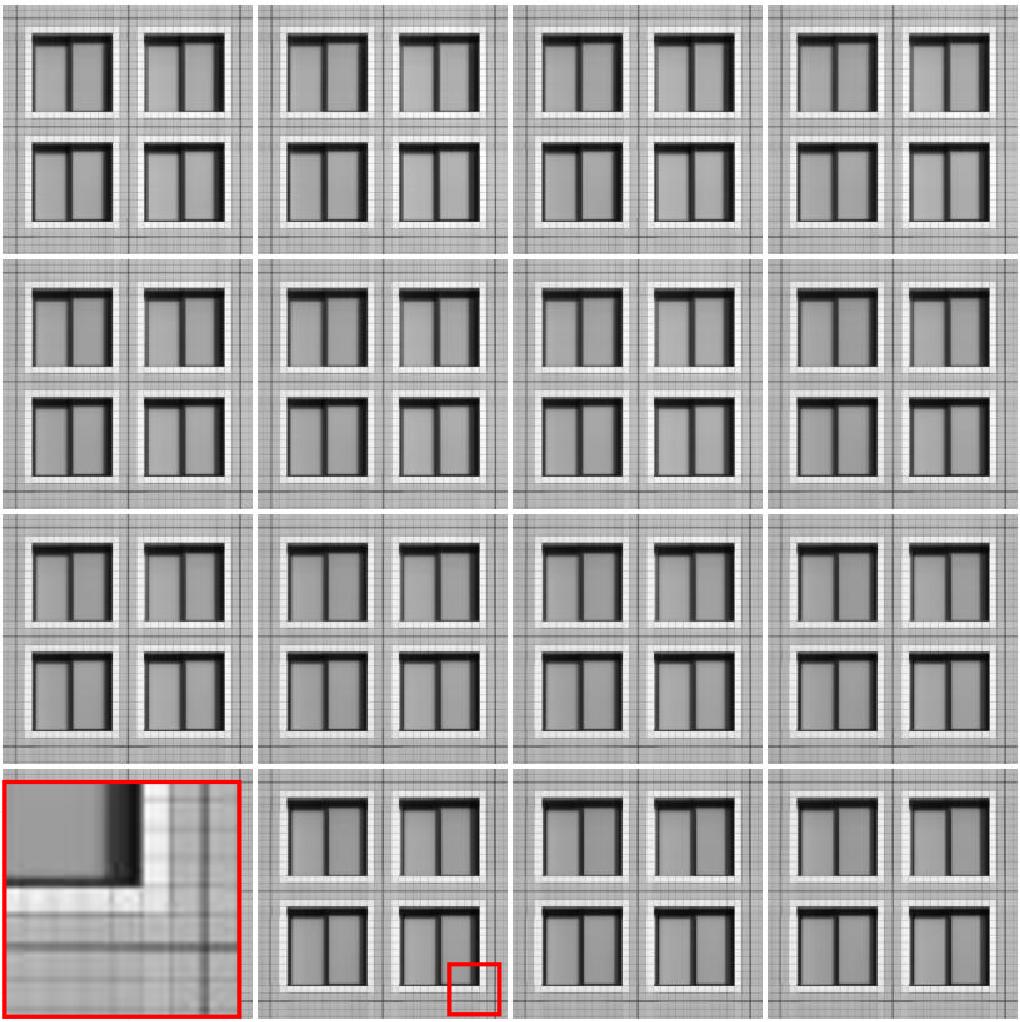}
		\end{minipage}}
	\caption{ Recovered images, zoomed regions,
and CPU time (in seconds) of different methods for the windows dataset.}\label{Fig: window}
\vspace{-8pt}
\end{figure*}
\begin{figure*}[!t]
\subfigcapskip=3pt%子标题与图片的距离
\setlength{\abovecaptionskip}{-0.5pt}%图片与注释之间的距离
\centering
	\subfigure[Observation]{
		\begin{minipage}[t]{0.13\textwidth}
			\centering
			\vspace{0.5mm}
			\includegraphics[width=2.4cm,height=2.4cm]{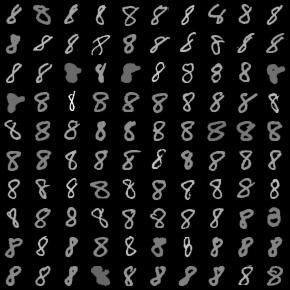}
		\end{minipage}}%\hspace{-4.5mm}
	\subfigure[RASL]{
		\begin{minipage}[t]{0.13\textwidth}
			\centering
			{\scriptsize{CPU: 29.51}}\vspace{0.5mm}
			\includegraphics[width=2.4cm,height=2.4cm]{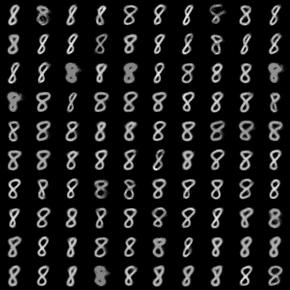}
		\end{minipage}}%\hspace{-4.5mm}
	\subfigure[t-GRASTA]{	
		\begin{minipage}[t]{0.13\textwidth}
			\centering
			{\scriptsize{CPU: 28.35}}\vspace{0.5mm}
			\includegraphics[width=2.4cm,height=2.4cm]{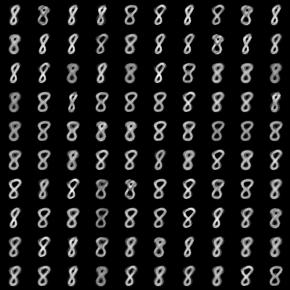}
		\end{minipage}}%\hspace{-4.5mm}
        \subfigure[NQLSD]{
		\begin{minipage}[t]{0.13\textwidth}
			\centering
			{\scriptsize{CPU: 31.97}}\vspace{0.5mm}
			\includegraphics[width=2.4cm,height=2.4cm]{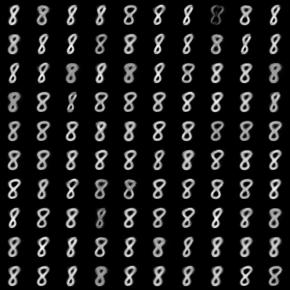}
		\end{minipage}}%\hspace{-4.5mm}
	\subfigure[$\ell_p$-ADMM]{
		\begin{minipage}[t]{0.13\textwidth}
			\centering
			{\scriptsize{CPU: 32.53}}\vspace{0.5mm}
			\includegraphics[width=2.4cm,height=2.4cm]{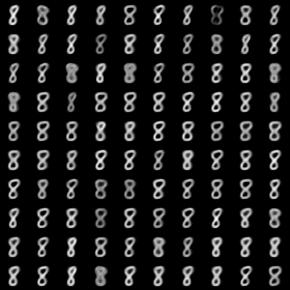}
		\end{minipage}}%\hspace{-4.5mm}
	\subfigure[NCALTS]{
		\begin{minipage}[t]{0.13\textwidth}
			\centering
			{\scriptsize{CPU: 37.06}}\vspace{0.5mm}
			\includegraphics[width=2.4cm,height=2.4cm]{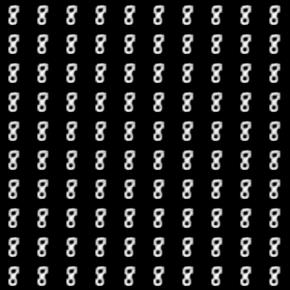}
		\end{minipage}}%\hspace{-4.5mm}
	\subfigure[TFM-TTP]{	
		\begin{minipage}[t]{0.13\textwidth}
			\centering
			{\scriptsize{CPU: 25.07}}\vspace{0.5mm}
			\includegraphics[width=2.4cm,height=2.4cm]{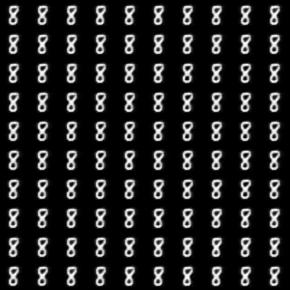}
		\end{minipage}}%\hspace{-4.5mm}
	\caption{ Recovered images and CPU time (in seconds) of different methods for the handwritten digit ``8''.}\label{Fig: digits}
\vspace{-8pt}	
\end{figure*}
\begin{figure*}[!t]
\subfigcapskip=3pt%子标题与图片的距离
\setlength{\abovecaptionskip}{-0.5pt}%图片与注释之间的距离
\centering
	\subfigure[Observation]{
		\begin{minipage}[t]{0.13\textwidth}
			\centering
			\vspace{0.5mm}
			\includegraphics[width=2.4cm,height=3.2cm]{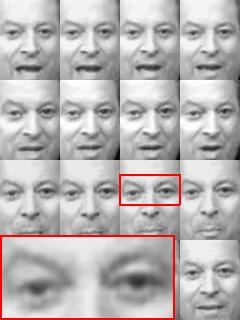}
		\end{minipage}}%\hspace{-2mm}
	\subfigure[RASL]{
		\begin{minipage}[t]{0.13\textwidth}
			\centering
			{\scriptsize{CPU: 24.88}}\vspace{0.5mm}
			\includegraphics[width=2.4cm,height=3.2cm]{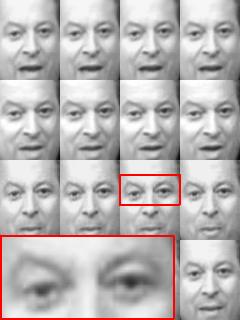}
		\end{minipage}}%\hspace{-2mm}
	\subfigure[t-GRASTA]{	
		\begin{minipage}[t]{0.13\textwidth}
			\centering
			{\scriptsize{CPU: 81.46}}\vspace{0.5mm}
			\includegraphics[width=2.4cm,height=3.2cm]{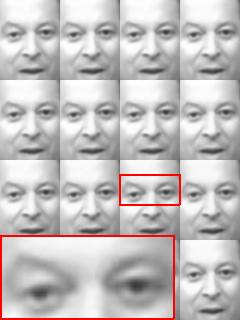}
		\end{minipage}}%\hspace{-2mm}
        \subfigure[NQLSD]{
		%\scriptsize{Arad}
		\begin{minipage}[t]{0.13\textwidth}
			\centering
			{\scriptsize{CPU:153.45}}\vspace{0.5mm}
			\includegraphics[width=2.4cm,height=3.2cm]{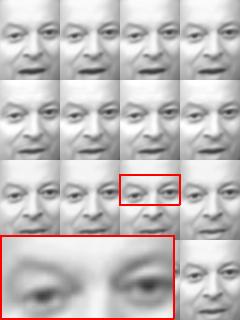}
		\end{minipage}}%\hspace{-2mm}
	\subfigure[$\ell_p$-ADMM]{
		\begin{minipage}[t]{0.13\textwidth}
			\centering
			{\scriptsize{CPU: 52.71}}\vspace{0.5mm}
			\includegraphics[width=2.4cm,height=3.2cm]{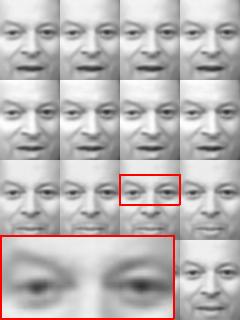}
		\end{minipage}}%\hspace{-2mm}
	\subfigure[NCALTS]{
		\begin{minipage}[t]{0.13\textwidth}
			\centering
			{\scriptsize{CPU: 56.80}}\vspace{0.5mm}
			\includegraphics[width=2.4cm,height=3.2cm]{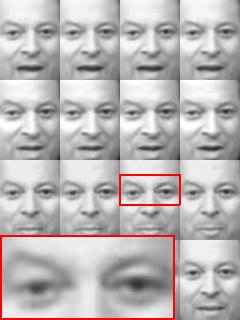}
		\end{minipage}}%\hspace{-2mm}
	\subfigure[TFM-TTP]{	
		\begin{minipage}[t]{0.13\textwidth}
			\centering
			{\scriptsize{CPU: 43.54}}\vspace{0.5mm}
			\includegraphics[width=2.4cm,height=3.2cm]{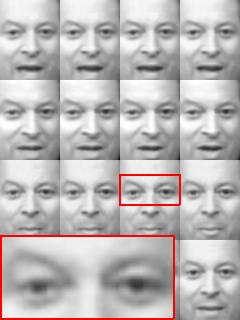}
		\end{minipage}}
	\caption{ Recovered images, zoomed regions,
and CPU time (in seconds) of different methods for the AI Gore talking dataset.}\label{Fig: Gore}	
\vspace{-8pt}
\end{figure*}
In this subsection, we  assess the effectiveness of TFM-TTP
 on the AR Face database\footnote{\url{http://www2.ece.ohio-state.edu/~aleix/ARdatabase.html}} for image alignment,
where the constructions of the images
with misalignment are similar to \cite{Peng_2012, Zhang_2021, Qiu_2021}.
%We evaluate our method and other state-of-the-art methods on color image inpainting and gray scale video inpainting tasks.
%To demonstrate the effectiveness of our proposed model to handle different levels of misalignment,
First, we select $8$ images from the AR Face
database
and resize them to $80\times 60$.
Second, for each image,
we generate 15 images using the Euclidean transformation
with uniformly distributed angles of rotation in the
range $[-\alpha/2,\alpha/2]$ degrees, and uniformly
distributed $x$- and $y$-translations in the range $[-s/2,s/2]$ pixels,
%We generate 15 images for each image using Euclidean transformations whose angles of rotation are uniformly distributed in the range $[-\alpha/2,\alpha/2]$ degrees and whose $x$- and $y$-translations are uniformly distributed in the range $[-s/2,s/2]$ pixels.
where $\alpha$ is selected from the set \{3, 6, 9\} and
$s$ is selected from the set \{3, 5, 7\} in the experiments.
For the 120 images with misalignment,
we add salt-and-pepper impulse noise with noise level $30\%$, or
the mixed noise (salt-and-pepper impulse noise with noise level $20\%$ and
Gaussian noise with zero-mean and variance $\sigma^2=0.01$).
Since the rotation or translation in the Euclidean transformation
may cause some pixels out of the
image boundary, we crop each transformed and noisy image with size $50\times40$ in
the center to construct the tensor $\mathcal D$.
The reconstruction error is used to measure the accuracy of the recovered tensor,
which is defined as ${\|\mathcal L-\mathcal L _{0}\|_{F}}/{\|\mathcal L _{0}\|_{F}}$.
Here $\mathcal L _{0}$  and $\mathcal L$ are the
ground-truth tensor and the recovered tensor, respectively.
For each pair $(\alpha,s)$,
we run 10 trials at random and take the average value of these results as the final result.

In Figure \ref{Fig: AR salt},  we show the recovered images
of different methods for the AR Face database corrupted by
salt-and-pepper impulse noise with noise level 30\%, where $ s=3, \alpha=3$.
We can see from this figure that the visual quality of the recovered images by TFM-TTP is better
than that by  RASL, t-GRASTA, NQLSD, $\ell_p$-ADMM, and NCALTS.
In particular, the last rows of Figure \ref{Fig: AR salt} (c)-(h)
show that the images recovered by TFM-TTP are more clear than those recovered by
 RASL, t-GRASTA, NQLSD, $\ell_p$-ADMM, and NCALTS,
%Compared with NCALTS, the images recovered by our method provide clearer visual quality,
such as the regions of mouths.
Moreover, the TFM-TTP can align the images better than RASL, t-GRASTA, and $\ell_p$-ADMM.
And the CPU time required by TFM-TTP is much less than other compared methods.

Table \ref{tab:chap:table_2} reports the reconstruction errors and CPU time (in seconds)
of different methods for the AR Face database corrupted by salt-and-pepper impulse noise with noise level $20\%$ and Gaussian noise with zero-mean and variance $\sigma^2=0.01$ simultaneously,
where the best results are highlighted in bold and the second best results are highlighted in underline.
We can see that the reconstruction errors of NQLSD
and $\ell_p$-ADMM are smaller than those of RASL and t-GRASTA for all cases.
Moreover, the TFM-TTP outperforms  RASL, t-GRASTA,
NQLSD, $\ell_p$-ADMM, and NCALTS for most testing cases in terms of construction errors.
And the computational time required by t-GRASTA is the most compared with other methods.
%As for the computational time, \textcolor{red}{the t-GRASTA takes most time.}
%one can observe that the t-GRASTA takes most time.
Besides, the TFM-TTP is significantly faster than other compared methods
at nearly the same (or lower) level of reconstruction errors.

Figure \ref{Fig: AR gauss} shows the visual quality  and CPU time (in seconds) comparisons
of the recovered images of different methods for the AR Face database
corrupted by  salt-and-pepper impulse noise with noise level 20\% and zero-mean Gaussian noise with variance $\sigma^2=0.01$,
where $ s=3, \alpha=9$.
It can be seen that the images recovered by TFM-TTP
are more clear than those recovered by RASL, t-GRASTA, NQLSD, $\ell_p$-ADMM, and NCALTS,
which can be seen from the eyes and mouths of these images, e.g., the image in the second row and the first column.
Moreover,
the alignment of these images obtained by TFM-TTP is better than that obtained by RASL, t-GRASTA and NQLSD.
Besides, the CPU time required by TFM-TTP is much less than that required by other compared methods for this case.

\begin{figure*}[!t]
\subfigcapskip=3pt%子标题与图片的距离
\setlength{\abovecaptionskip}{-0.5pt}%图片与注释之间的距离
\centering
	\subfigure[Observation]{
		\begin{minipage}[t]{0.13\textwidth}
			\centering
			\vspace{0.5mm}
			\includegraphics[width=2.4cm,height=2.4cm]{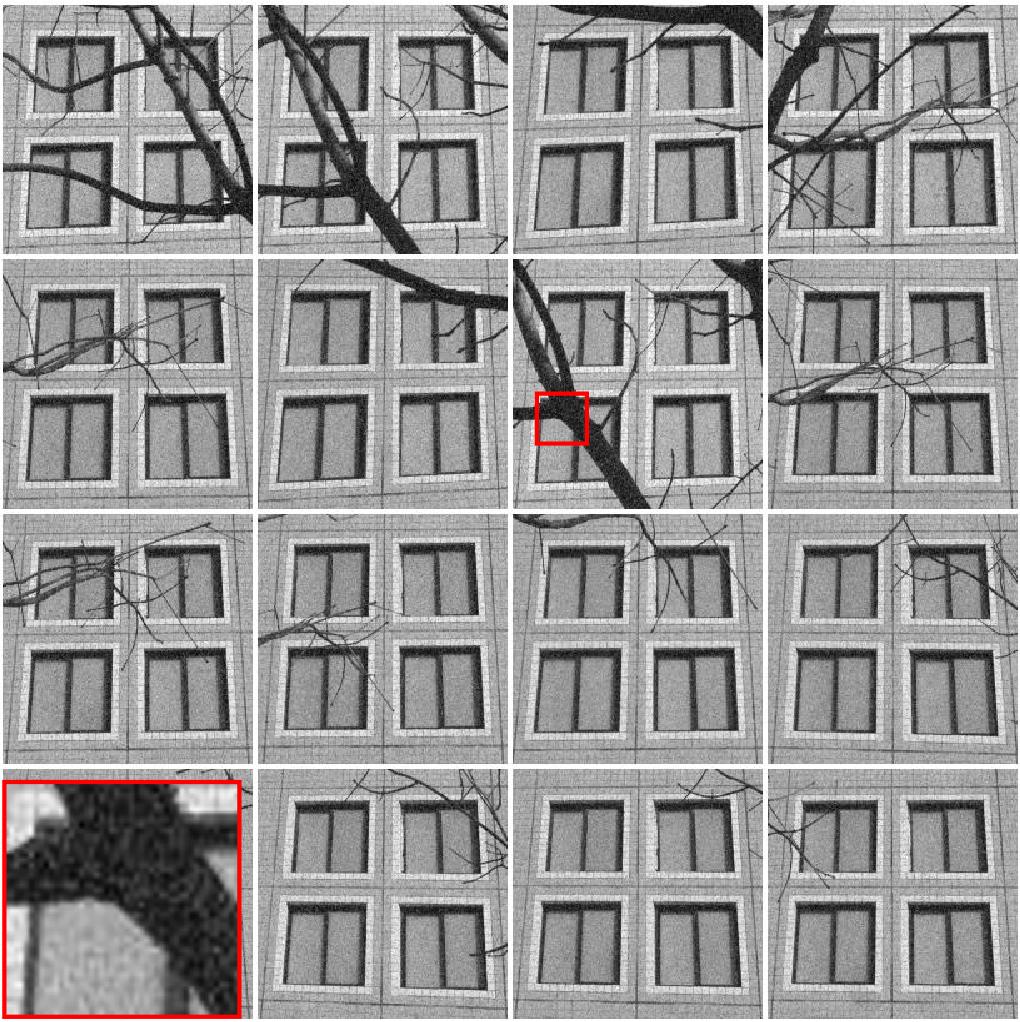}
		\end{minipage}}%\hspace{-2mm}
	\subfigure[RASL]{
		\begin{minipage}[t]{0.13\textwidth}
			\centering
			{\scriptsize{CPU: 283.85}}\vspace{0.5mm}
			\includegraphics[width=2.4cm,height=2.4cm]{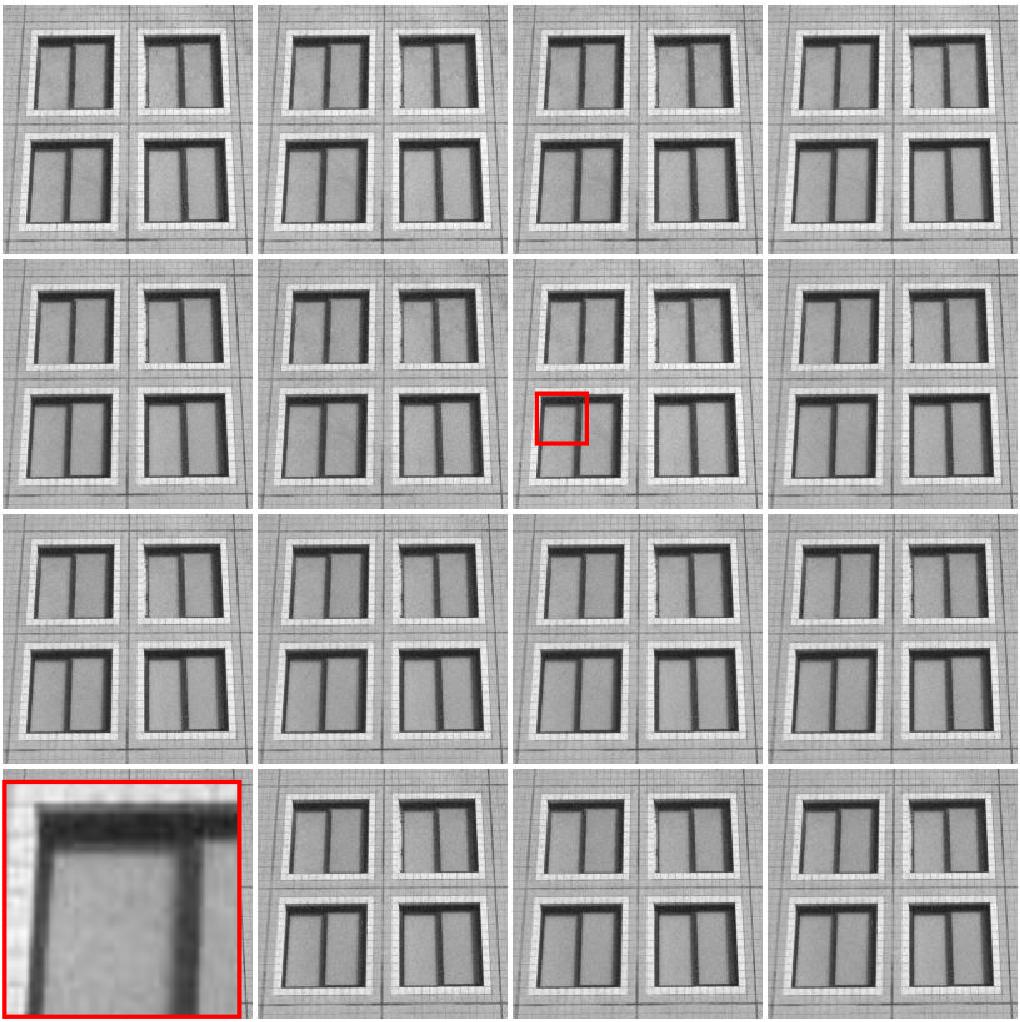}
		\end{minipage}}%\hspace{-2mm}
	\subfigure[t-GRASTA]{	
		\begin{minipage}[t]{0.13\textwidth}
			\centering
			{\scriptsize{CPU: 476.48}}\vspace{0.5mm}
			\includegraphics[width=2.4cm,height=2.4cm]{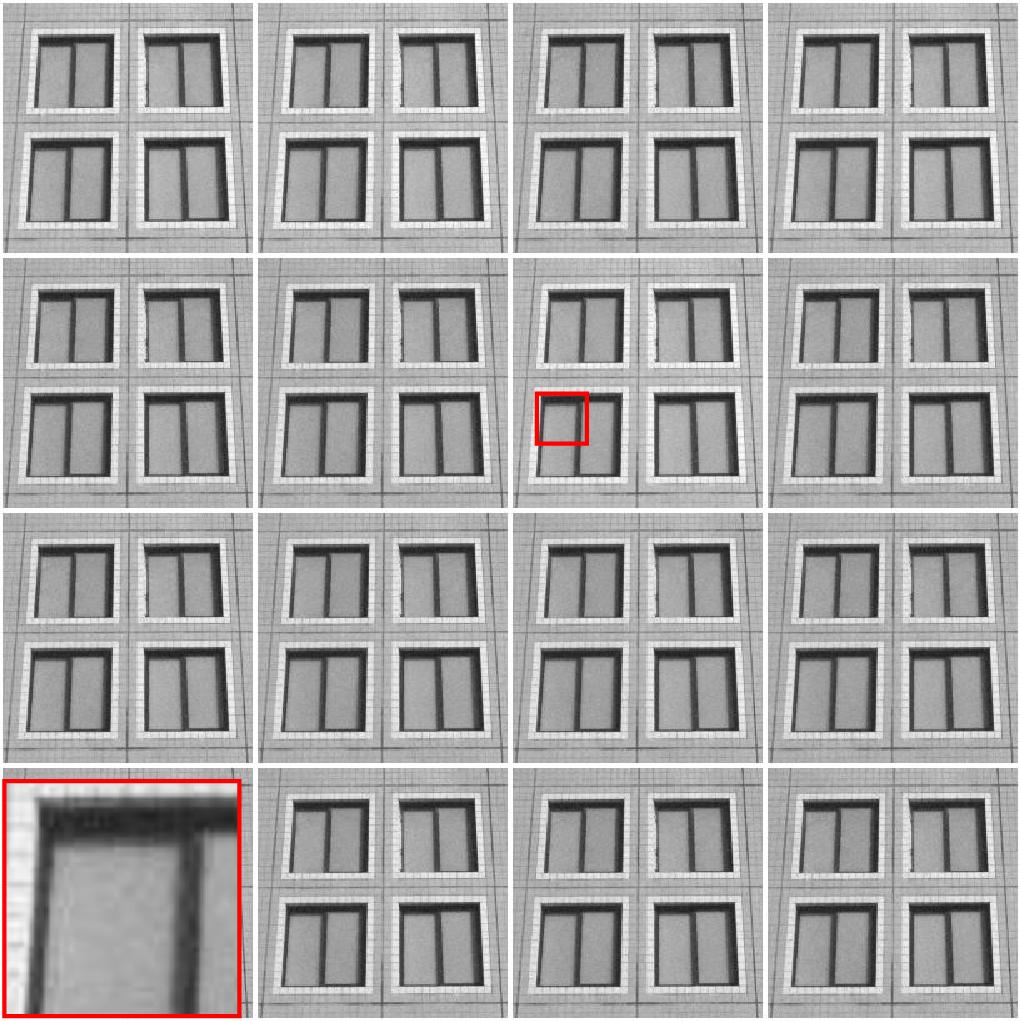}
		\end{minipage}}%\hspace{-2mm}
        \subfigure[NQLSD]{
		\begin{minipage}[t]{0.13\textwidth}
			\centering
			{\scriptsize{CPU:489.89}}\vspace{0.5mm}
			\includegraphics[width=2.4cm,height=2.4cm]{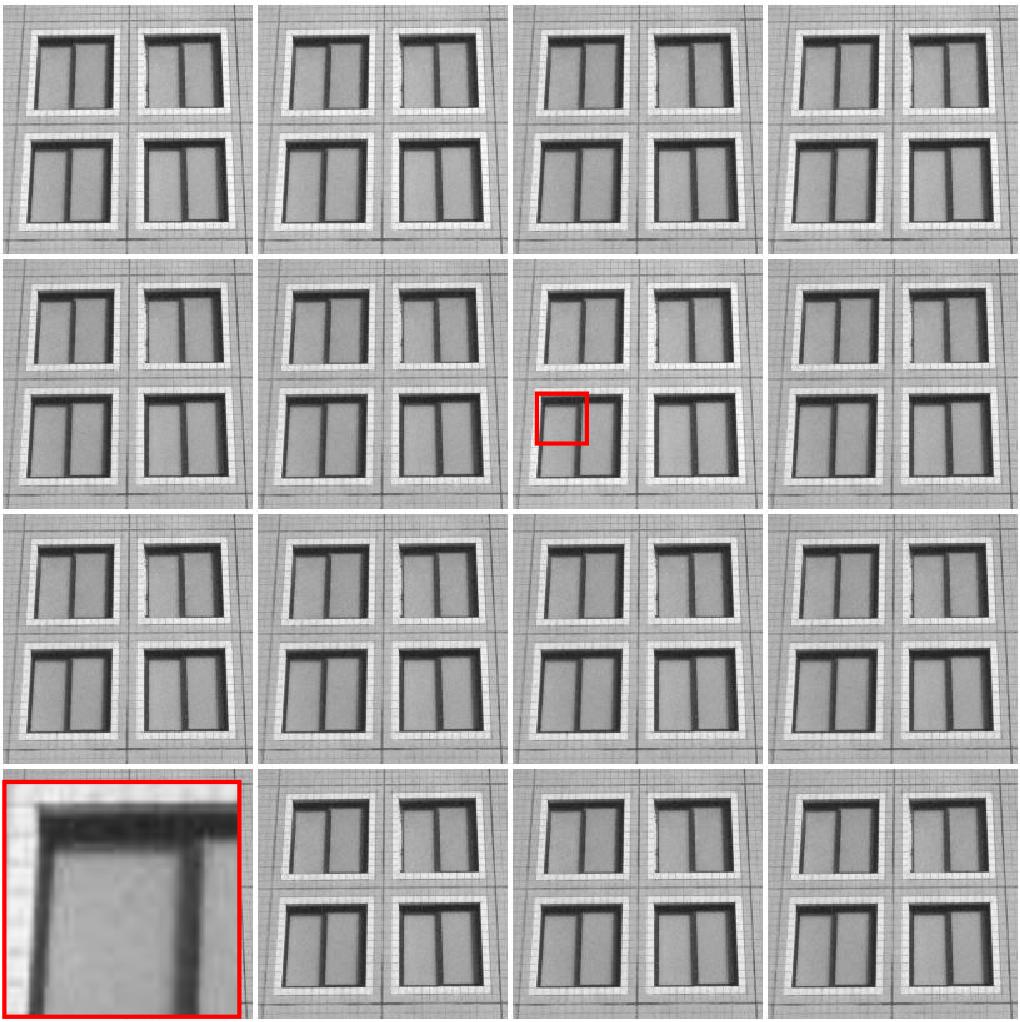}
		\end{minipage}}%\hspace{-2mm}
	\subfigure[$\ell_p$-ADMM]{
		\begin{minipage}[t]{0.13\textwidth}
			\centering
			{\scriptsize{CPU: 433.03}}\vspace{0.5mm}
			\includegraphics[width=2.4cm,height=2.4cm]{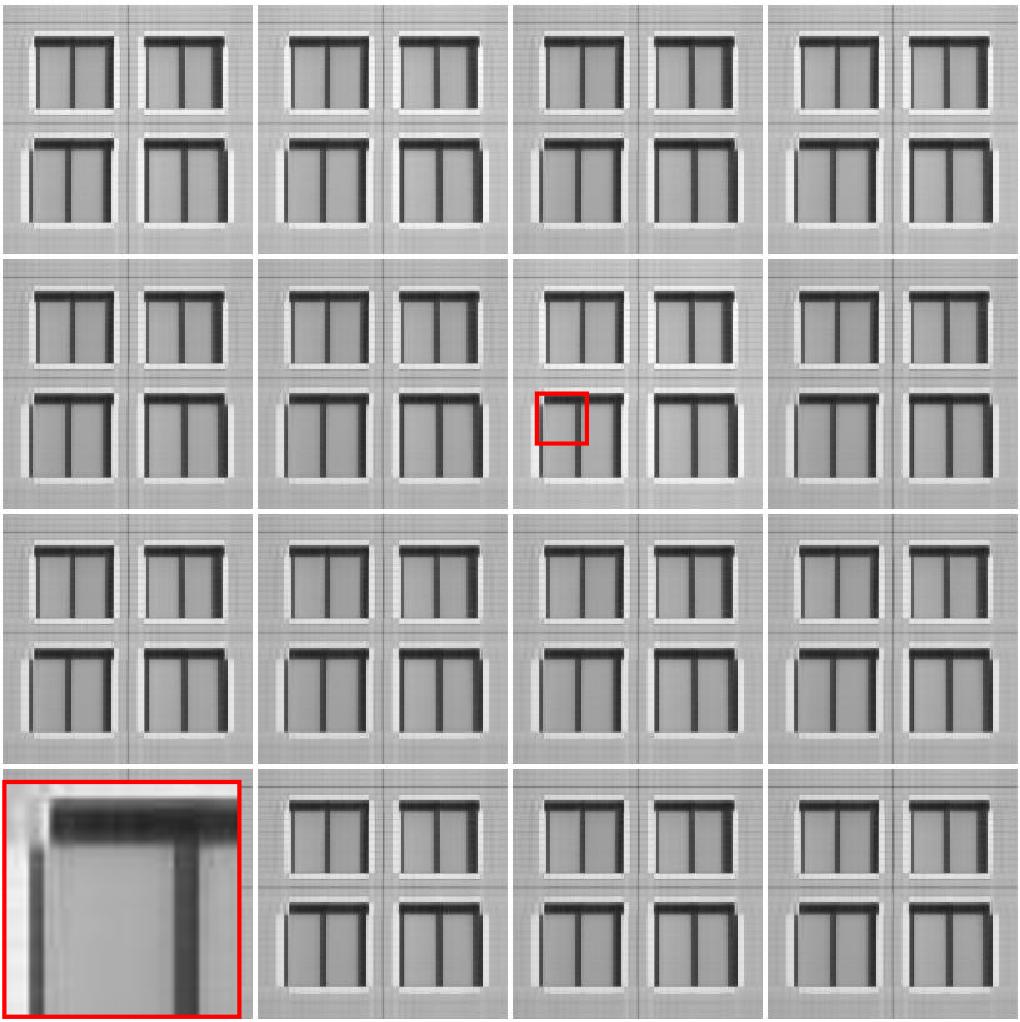}
		\end{minipage}}%\hspace{-2mm}
	\subfigure[NCALTS]{
		\begin{minipage}[t]{0.13\textwidth}
			\centering
			{\scriptsize{CPU: 471.71}}\vspace{0.5mm}
			\includegraphics[width=2.4cm,height=2.4cm]{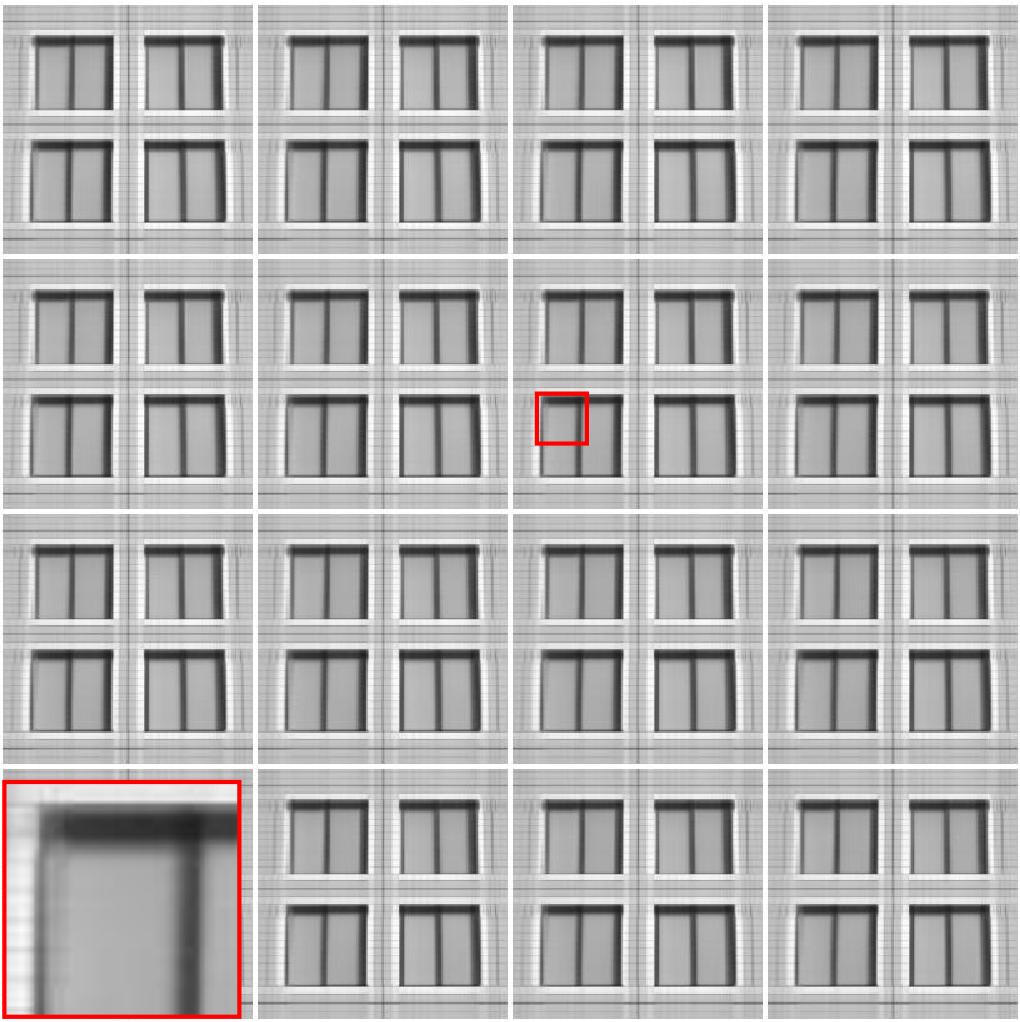}
		\end{minipage}}%\hspace{-2mm}
	\subfigure[TFM-TTP]{	
		\begin{minipage}[t]{0.13\textwidth}
			\centering
			{\scriptsize{CPU: 242.87}}\vspace{0.5mm}
			\includegraphics[width=2.4cm,height=2.4cm]{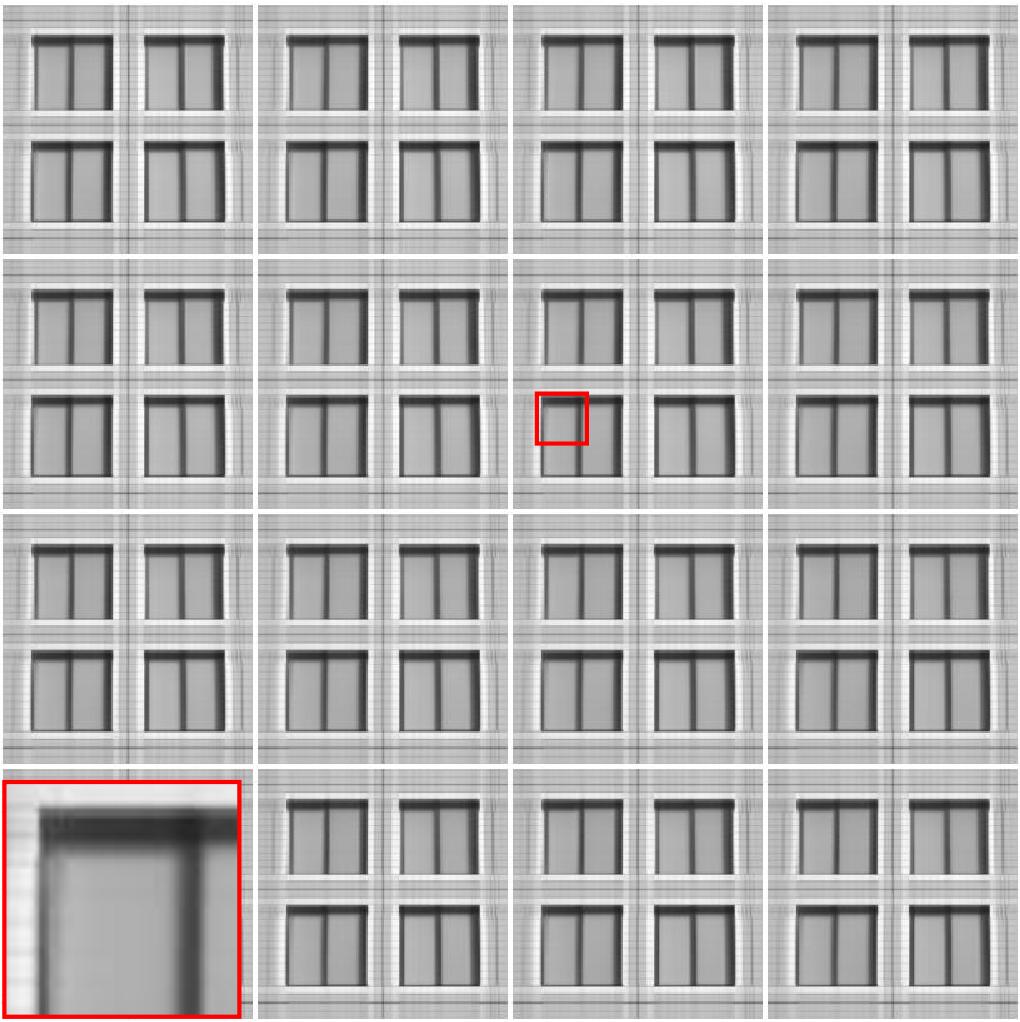}
		\end{minipage}}%\hspace{-2mm}
\caption{Recovered images, zoomed regions, and CPU time (in seconds) of different methods
for the window images corrupted by 20\% salt-and-pepper impulse noise.}\label{Fig: window salt}	
\vspace{-8pt}
\end{figure*}
	
\begin{figure*}[!t]
\subfigcapskip=3pt%子标题与图片的距离
\setlength{\abovecaptionskip}{-0.5pt}%图片与注释之间的距离
\centering
	\subfigure[Observation]{
		\begin{minipage}[t]{0.13\textwidth}
			\centering
			\vspace{0.5mm}
			\includegraphics[width=2.4cm,height=2.4cm]{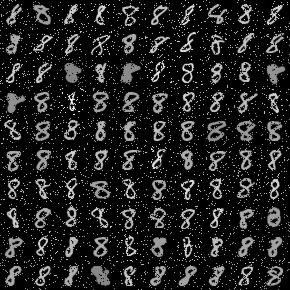}
		\end{minipage}}%\hspace{-2mm}
	\subfigure[RASL]{
		\begin{minipage}[t]{0.13\textwidth}
			\centering
			{\scriptsize{CPU: 4.09}}\vspace{0.5mm}
			\includegraphics[width=2.4cm,height=2.4cm]{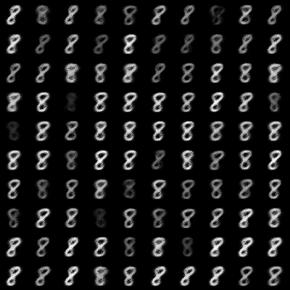}
		\end{minipage}}%\hspace{-2mm}
	\subfigure[t-GRASTA]{	
		\begin{minipage}[t]{0.13\textwidth}
			\centering
			{\scriptsize{CPU: 42.67}}\vspace{0.5mm}
			\includegraphics[width=2.4cm,height=2.4cm]{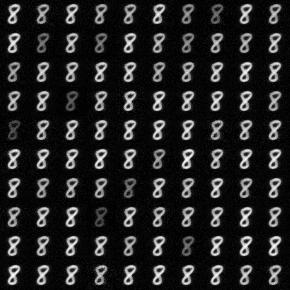}
		\end{minipage}}%\hspace{-2mm}
        \subfigure[NQLSD]{
		\begin{minipage}[t]{0.13\textwidth}
			\centering
			{\scriptsize{CPU: 19.16}}\vspace{0.5mm}
			\includegraphics[width=2.4cm,height=2.4cm]{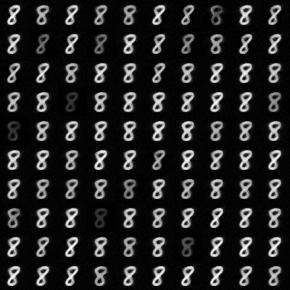}
		\end{minipage}}%\hspace{-2mm}
	\subfigure[$\ell_p$-ADMM]{
		\begin{minipage}[t]{0.13\textwidth}
			\centering
			{\scriptsize{CPU: 32.70}}\vspace{0.5mm}
			\includegraphics[width=2.4cm,height=2.4cm]{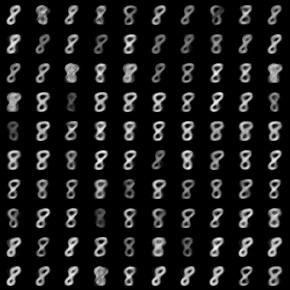}
		\end{minipage}}%\hspace{-2mm}
	\subfigure[NCALTS]{
		\begin{minipage}[t]{0.13\textwidth}
			\centering
			{\scriptsize{CPU: 19.35}}\vspace{0.5mm}
			\includegraphics[width=2.4cm,height=2.4cm]{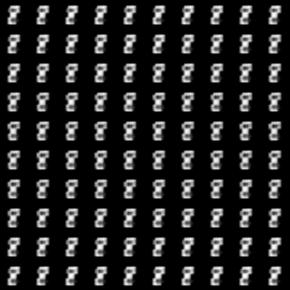}
		\end{minipage}}%\hspace{-2mm}
	\subfigure[TFM-TTP]{	
		\begin{minipage}[t]{0.13\textwidth}
			\centering
			{\scriptsize{CPU: 14.63}}\vspace{0.5mm}
			\includegraphics[width=2.4cm,height=2.4cm]{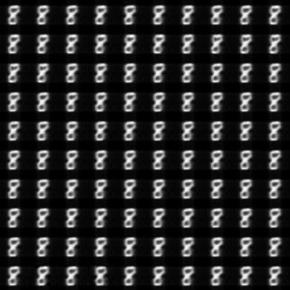}
		\end{minipage}}%\hspace{-2mm}
	\caption{ Recovered images and CPU time (in seconds) of different methods
for the handwritten digits ``8'' corrupted by  10\% salt-and-pepper impulse noise.}\label{Fig: digits salt}	
\vspace{-8pt}
\end{figure*}

\subsection{Image Alignment with Uncontrolled Images}

In this subsection, we compare TFM-TTP with RASL, t-GRASTA, NQLSD,
$\ell_p$-ADMM, NCALTS on three real-world image datasets,
where there are no ground-truth images for comparisons.
Here the original images are misaligned
and  corrupted by sparse noise.
The first image dataset (windows) includes 16 images of the side of
a building \cite{Peng_2012}, which are taken by a
personal camera from different viewpoints.
The 16 images have some occlusions due to the tree branches,
and the size of each image is $1200 \times 1600$.
 We crop these images to $200 \times 200$ in the center and
then the size of the resulting tensor is $200\times 200\times 16$.
The second dataset contains 100 images of
handwritten digit ``8'' taken from the MNIST
dataset\footnote{\url{http://yann.lecun.com/exdb/mnist/}},
which are misaligned due to the different characteristic of the handwritten  digit.
We use $100$ images with size $29 \times 29$ and then the resulting tensor is $29\times 29\times 100$.
The third dataset is the Al Gore
talking \cite{Peng_2012}, which contains 140 frames with size of each image being $80 \times 60$ and are misaligned
by a significant jitter of the detector from frame
to frame.
Then the size of resulting tensor  is  $80 \times 60\times 140$.
To demonstrate the effectiveness of our proposed  TFM-TTP method,
we conduct three experiments in the following subsections:
(1) Recover the natural images;
(2) Recover the natural images with salt-and-pepper impulse noise;
(3) Recover the natural images with salt-and-pepper impulse noise and Gaussian noise.

\begin{figure*}[!t]
\subfigcapskip=3pt%子标题与图片的距离
\setlength{\abovecaptionskip}{-0.5pt}%图片与注释之间的距离
\centering
	\subfigure[Observation]{
		\begin{minipage}[t]{0.13\textwidth}
			\centering
			\vspace{0.5mm}
            \includegraphics[width=2.4cm,height=3.3cm]{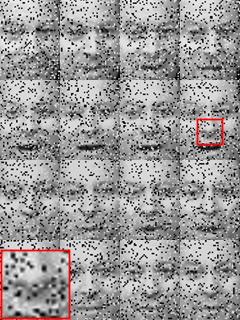}
		\end{minipage}}%\hspace{-2mm}
	\subfigure[RASL]{
		\begin{minipage}[t]{0.13\textwidth}
			\centering
			{\scriptsize{CPU: 121.42}}\vspace{0.5mm}
			\includegraphics[width=2.4cm,height=3.3cm]{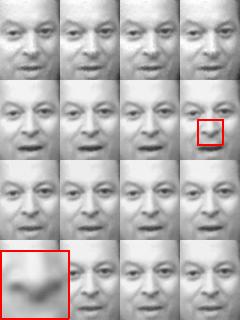}
		\end{minipage}}%\hspace{-2mm}
	\subfigure[t-GRASTA]{	
		\begin{minipage}[t]{0.13\textwidth}
			\centering
			{\scriptsize{CPU: 151.37}}\vspace{0.5mm}
			\includegraphics[width=2.4cm,height=3.3cm]{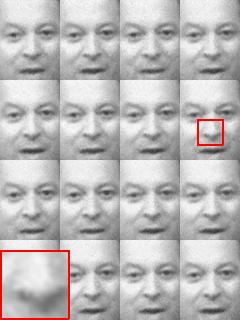}
		\end{minipage}}%\hspace{-2mm}
        \subfigure[NQLSD]{
		\begin{minipage}[t]{0.13\textwidth}
			\centering
			{\scriptsize{CPU: 256.91}}\vspace{0.5mm}
			\includegraphics[width=2.4cm,height=3.3cm]{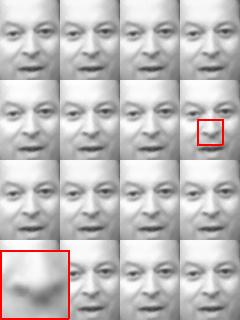}
		\end{minipage}}%\hspace{-2mm}
	\subfigure[$\ell_p$-ADMM]{
		\begin{minipage}[t]{0.13\textwidth}
			\centering
			{\scriptsize{CPU: 431.52}}\vspace{0.5mm}
			\includegraphics[width=2.4cm,height=3.3cm]{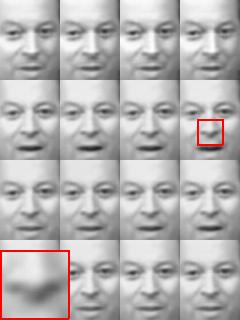}
		\end{minipage}}%\hspace{-2mm}
	\subfigure[NCALTS]{
		\begin{minipage}[t]{0.13\textwidth}
			\centering
			{\scriptsize{CPU: 95.22}}\vspace{0.5mm}
			\includegraphics[width=2.4cm,height=3.3cm]{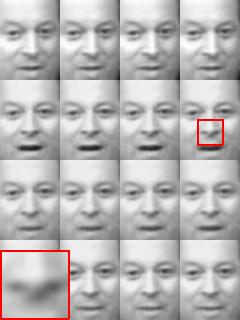}
		\end{minipage}}%\hspace{-2mm}
	\subfigure[TFM-TTP]{	
		\begin{minipage}[t]{0.13\textwidth}
			\centering
			{\scriptsize{CPU: 44.46}}\vspace{0.5mm}
			\includegraphics[width=2.4cm,height=3.3cm]{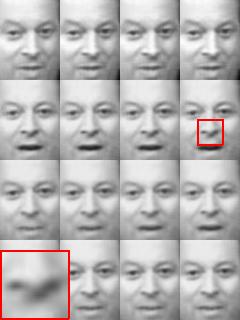}
		\end{minipage}}%\hspace{-2mm}
	\caption{ Recovered images, zoomed regions, and CPU time (in seconds) of different
methods for the AI Gore talking dataset corrupted by  20\% salt-and-pepper impulse noise.}\label{Fig: Gore-salt}
\vspace{-8pt}	
\end{figure*}

\begin{figure*}[!t]
\subfigcapskip=3pt%子标题与图片的距离
\setlength{\abovecaptionskip}{-0.5pt}%图片与注释之间的距离
	\centering
	\subfigure[Observation]{
		\begin{minipage}[t]{0.13\textwidth}
			\centering
			\vspace{0.5mm}
			\includegraphics[width=2.4cm,height=2.4cm]{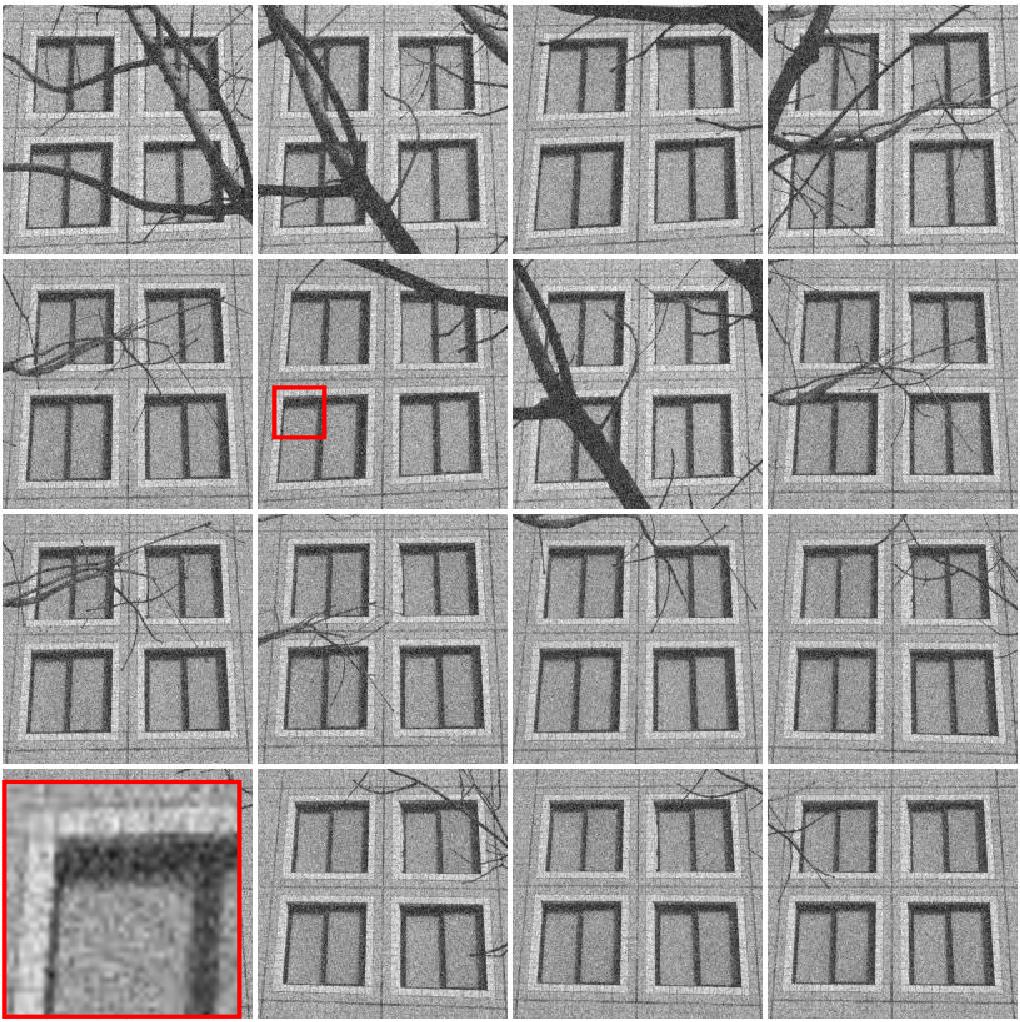}
		\end{minipage}}%\hspace{-2mm}
	\subfigure[RASL]{
		\begin{minipage}[t]{0.13\textwidth}
			\centering
			{\scriptsize{CPU: 363.49}}\vspace{0.5mm}
			\includegraphics[width=2.4cm,height=2.4cm]{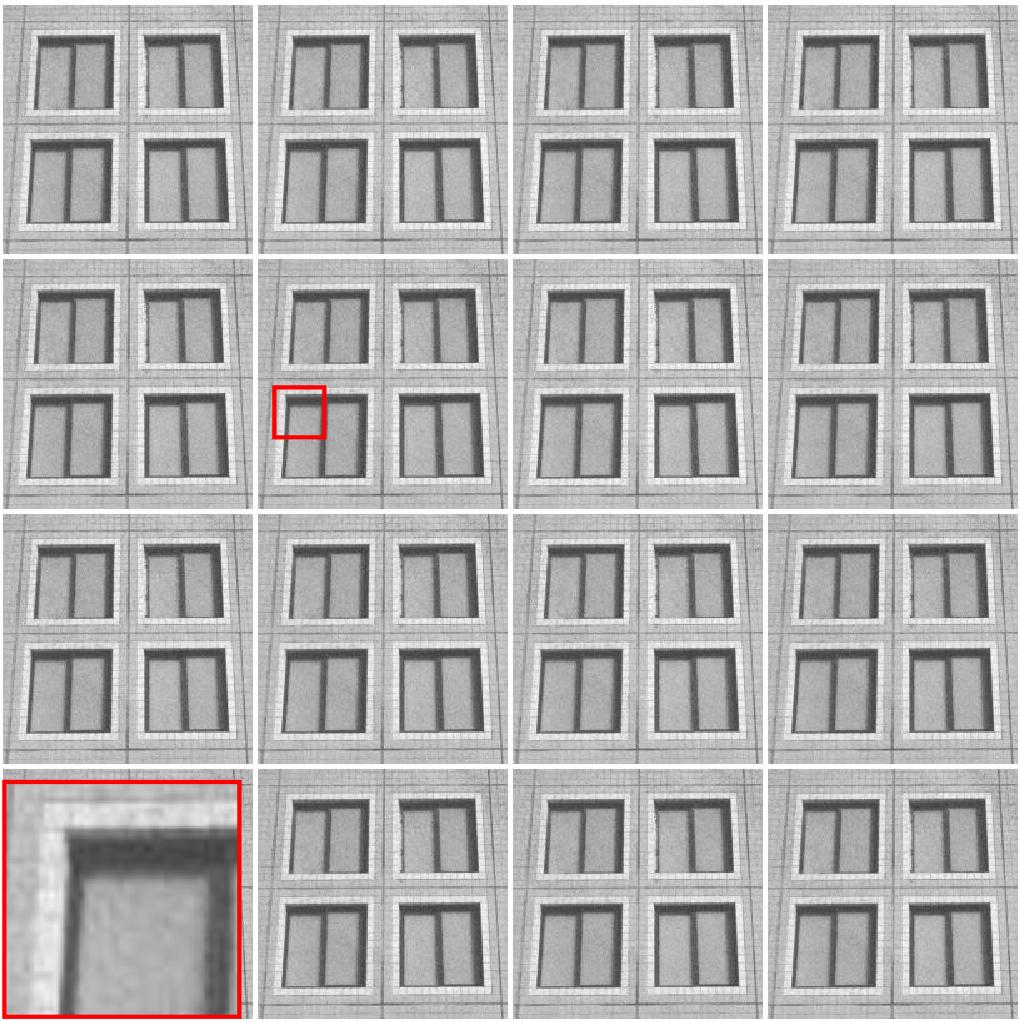}
		\end{minipage}}%\hspace{-2mm}
	\subfigure[t-GRASTA]{	
		\begin{minipage}[t]{0.13\textwidth}
			\centering
			{\scriptsize{CPU: 430.07}}\vspace{0.5mm}
			\includegraphics[width=2.4cm,height=2.4cm]{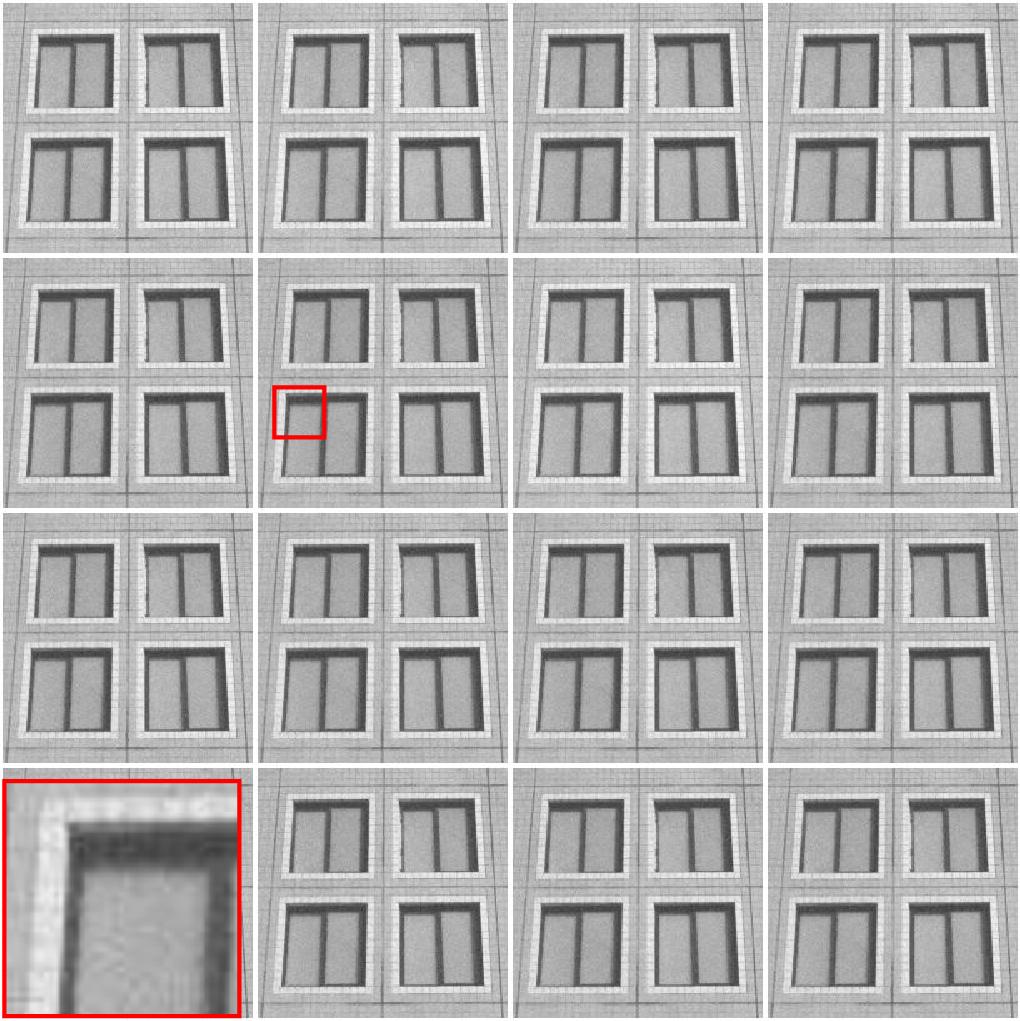}
		\end{minipage}}%\hspace{-2mm}
        \subfigure[NQLSD]{
		\begin{minipage}[t]{0.13\textwidth}
			\centering
			{\scriptsize{CPU: 201.47}}\vspace{0.5mm}
			\includegraphics[width=2.4cm,height=2.4cm]{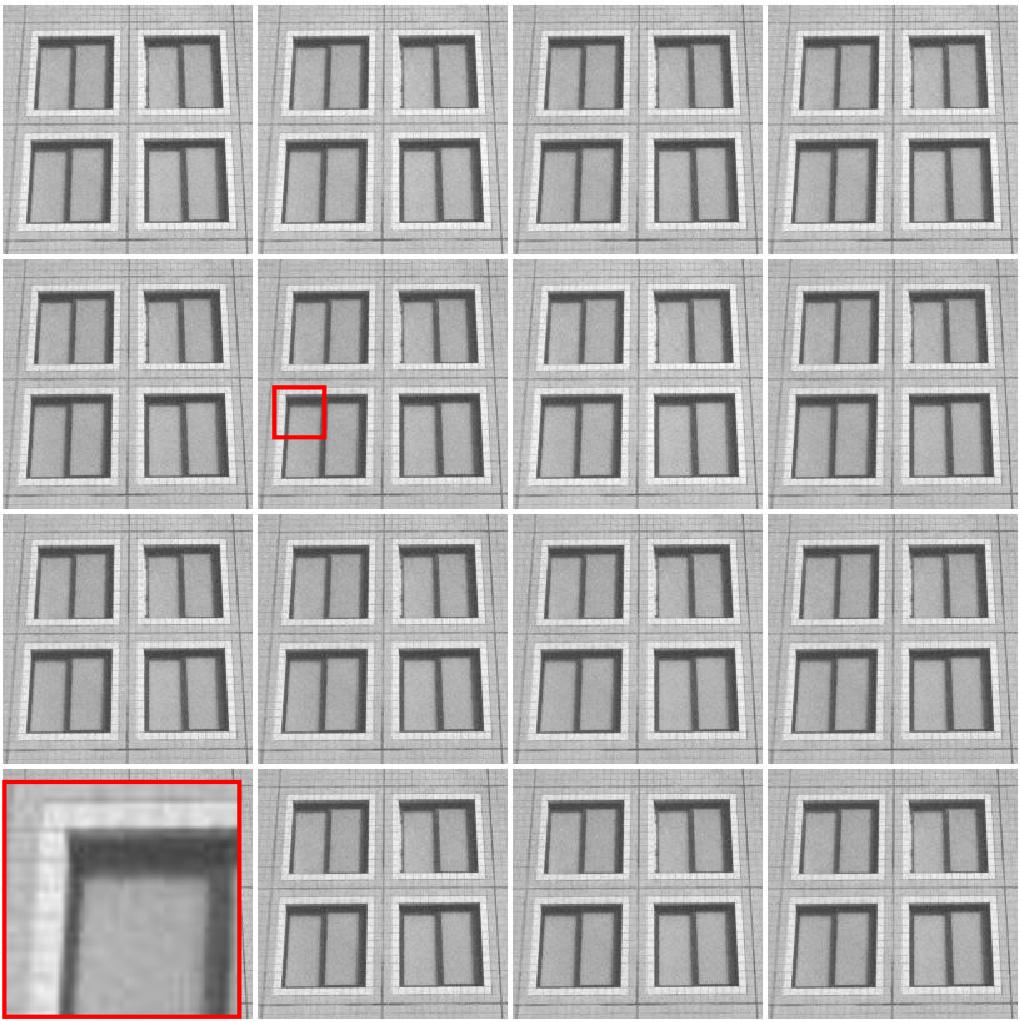}
		\end{minipage}}%\hspace{-2mm}
	\subfigure[$\ell_p$-ADMM]{
		\begin{minipage}[t]{0.13\textwidth}
			\centering
			{\scriptsize{CPU:1213.49}}\vspace{0.5mm}
			\includegraphics[width=2.4cm,height=2.4cm]{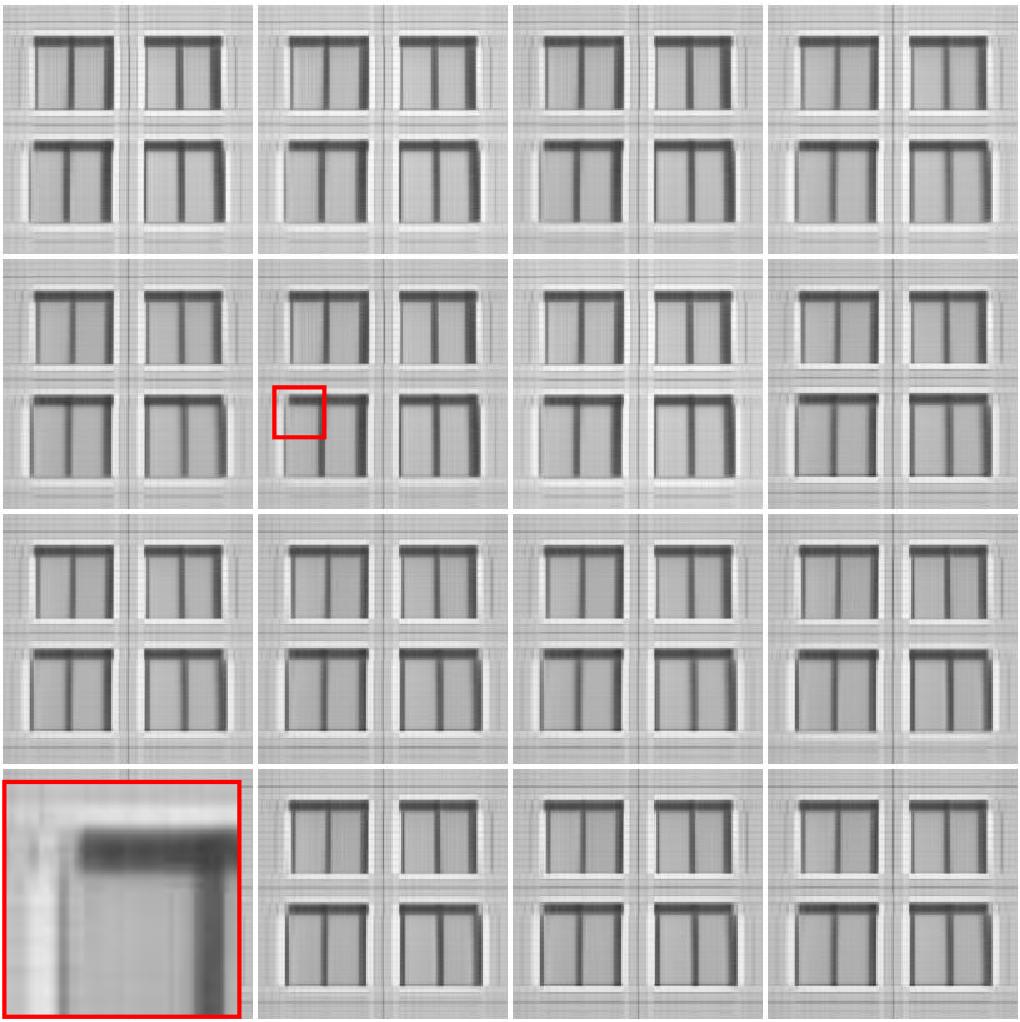}
		\end{minipage}}%\hspace{-2mm}
	\subfigure[NCALTS]{
		\begin{minipage}[t]{0.13\textwidth}
			\centering
			{\scriptsize{CPU: 469.78}}\vspace{0.5mm}
			\includegraphics[width=2.4cm,height=2.4cm]{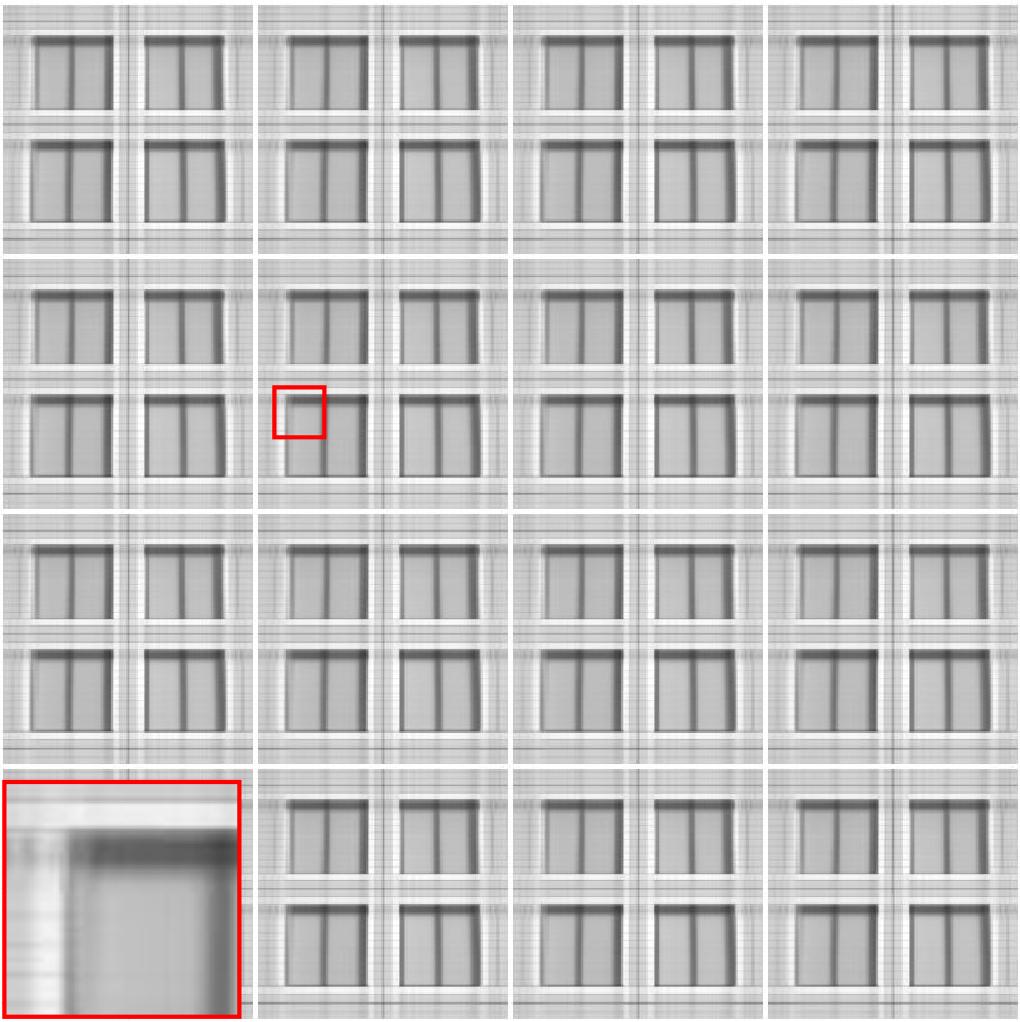}
		\end{minipage}}%\hspace{-2mm}
	\subfigure[TFM-TTP]{	
		\begin{minipage}[t]{0.13\textwidth}
			\centering
			{\scriptsize{CPU: 192.60}}\vspace{0.5mm}
			\includegraphics[width=2.4cm,height=2.4cm]{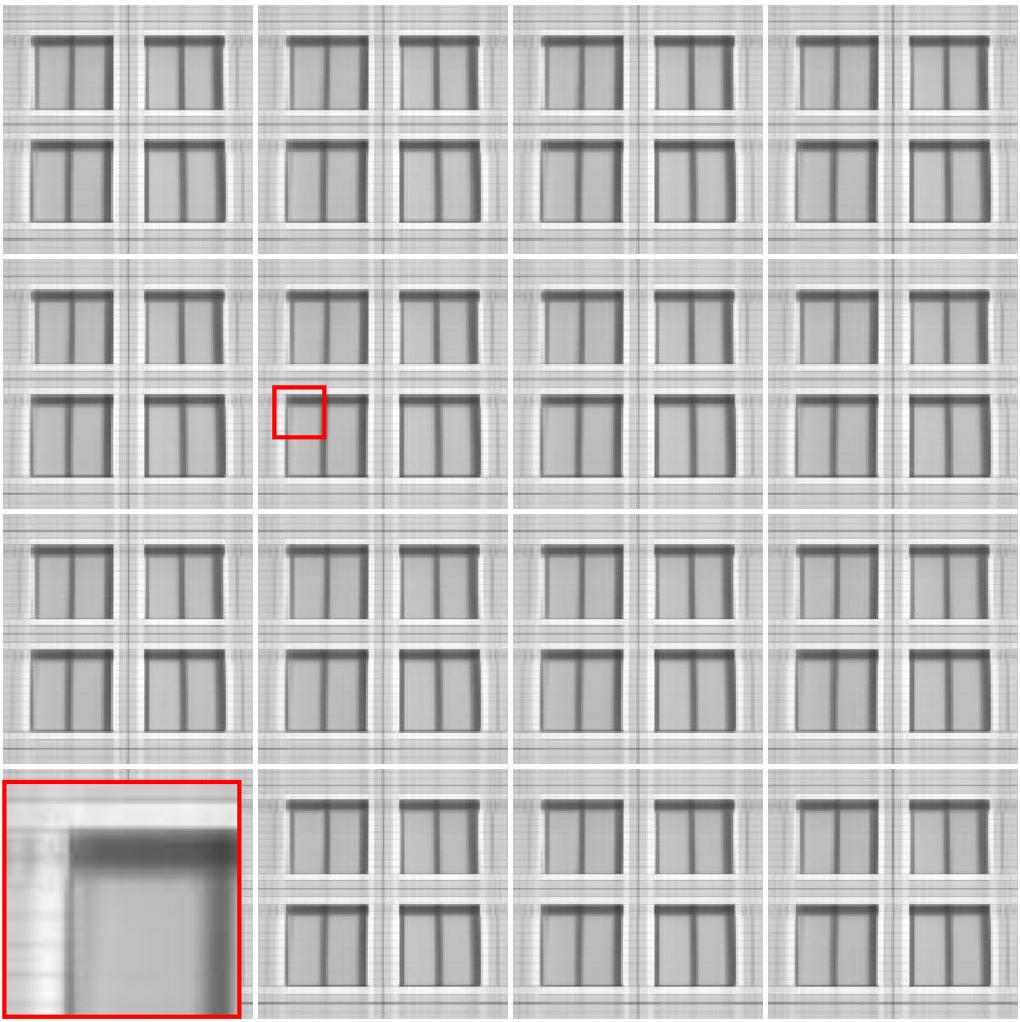}
		\end{minipage}}%\hspace{-2mm}
	\caption{ Recovered images, zoomed regions, and CPU time (in seconds) of different methods
for the window images corrupted by salt-and-pepper
impulse noise with noise level 30\% and zero-mean Gaussian noise with variance $\sigma^2=0.05$.}\label{Fig: window-mixnoise}
\vspace{-8pt}
\end{figure*}

\begin{figure*}[!t]
\subfigcapskip=3pt%子标题与图片的距离
\setlength{\abovecaptionskip}{-0.5pt}%图片与注释之间的距离
\centering
	\subfigure[Observation]{
		\begin{minipage}[t]{0.13\textwidth}
			\centering
			\vspace{0.5mm}
			\includegraphics[width=2.4cm,height=2.4cm]{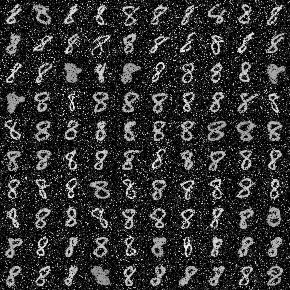}
		\end{minipage}}%\hspace{-2mm}
	\subfigure[RASL]{
		\begin{minipage}[t]{0.13\textwidth}
			\centering
			{\scriptsize{CPU: 28.39}}\vspace{0.5mm}
			\includegraphics[width=2.4cm,height=2.4cm]{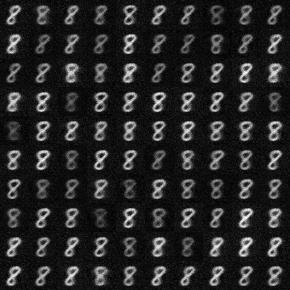}
		\end{minipage}}%\hspace{-2mm}
	\subfigure[t-GRASTA]{	
		\begin{minipage}[t]{0.13\textwidth}
			\centering
			{\scriptsize{CPU: 43.98}}\vspace{0.5mm}
			\includegraphics[width=2.4cm,height=2.4cm]{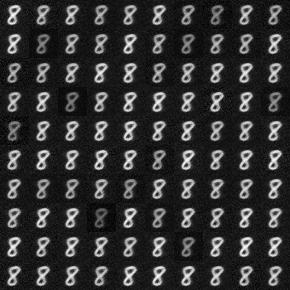}
		\end{minipage}}%\hspace{-2mm}
        \subfigure[NQLSD]{
		\begin{minipage}[t]{0.13\textwidth}
			\centering
			{\scriptsize{CPU: 26.43}}\vspace{0.5mm}
			\includegraphics[width=2.4cm,height=2.4cm]{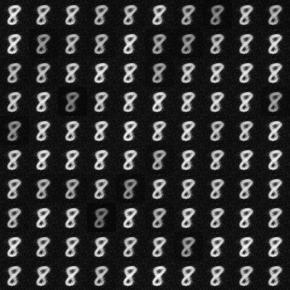}
		\end{minipage}}%\hspace{-2mm}
	\subfigure[$\ell_p$-ADMM]{
		\begin{minipage}[t]{0.13\textwidth}
			\centering
			{\scriptsize{CPU: 28.16}}\vspace{0.5mm}
			\includegraphics[width=2.4cm,height=2.4cm]{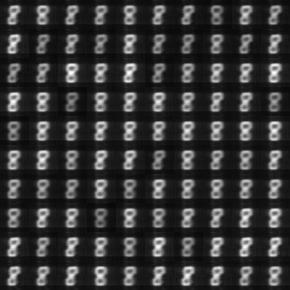}
		\end{minipage}}%\hspace{-2mm}
	\subfigure[NCALTS]{
		\begin{minipage}[t]{0.13\textwidth}
			\centering
			{\scriptsize{CPU: 19.64}}\vspace{0.5mm}
			\includegraphics[width=2.4cm,height=2.4cm]{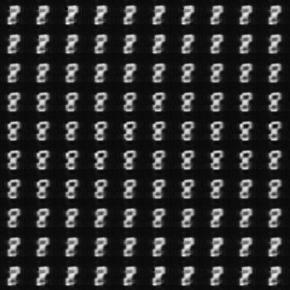}
		\end{minipage}}%\hspace{-2mm}
	\subfigure[TFM-TTP]{	
		\begin{minipage}[t]{0.13\textwidth}
			\centering
			{\scriptsize{CPU: 16.66}}\vspace{0.5mm}
			\includegraphics[width=2.4cm,height=2.4cm]{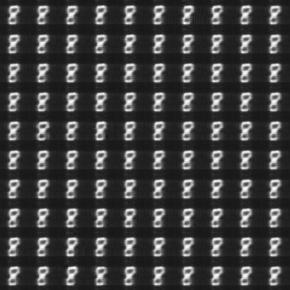}
		\end{minipage}}%\hspace{-2mm}
	\caption{ Recovered images and CPU time (in seconds) of different methods
for the  handwritten digits ``8'' corrupted by salt-and-pepper impulse noise with noise level  10\%
 and zero-mean Gaussian noise with variance $\sigma^2=0.05$.}\label{Fig: digits-mixnoise}		
\vspace{-8pt}
\end{figure*}

\begin{figure*}[!t]
\subfigcapskip=3pt%子标题与图片的距离
\setlength{\abovecaptionskip}{-0.5pt}%图片与注释之间的距离
\centering
	\subfigure[Observation]{
		\begin{minipage}[t]{0.13\textwidth}
			\centering
			\vspace{0.5mm}
			\includegraphics[width=2.4cm,height=3.3cm]{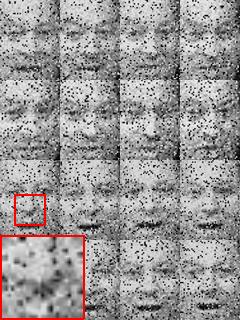}
		\end{minipage}}%\hspace{-2mm}
	\subfigure[RASL]{
		\begin{minipage}[t]{0.13\textwidth}
			\centering
			{\scriptsize{CPU: 95.35}}\vspace{0.5mm}
			\includegraphics[width=2.4cm,height=3.3cm]{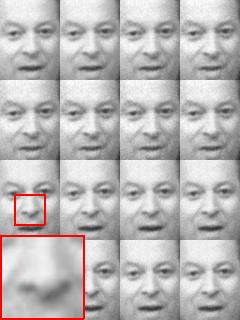}
		\end{minipage}}%\hspace{-2mm}
	\subfigure[t-GRASTA]{	
		\begin{minipage}[t]{0.13\textwidth}
			\centering
			{\scriptsize{CPU: 132.14}}\vspace{0.5mm}
			\includegraphics[width=2.4cm,height=3.3cm]{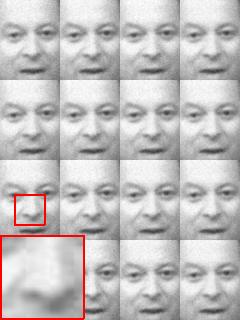}
		\end{minipage}}%\hspace{-2mm}
        \subfigure[NQLSD]{
		\begin{minipage}[t]{0.13\textwidth}
			\centering
			{\scriptsize{CPU: 225.16}}\vspace{0.5mm}
			\includegraphics[width=2.4cm,height=3.3cm]{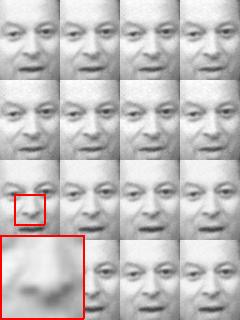}
		\end{minipage}}%\hspace{-2mm}
	\subfigure[$\ell_p$-ADMM]{
		\begin{minipage}[t]{0.13\textwidth}
			\centering
			{\scriptsize{CPU: 207.55}}\vspace{0.5mm}
			\includegraphics[width=2.4cm,height=3.3cm]{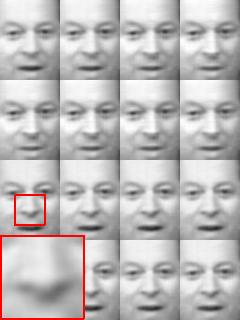}
		\end{minipage}}%\hspace{-2mm}
	\subfigure[NCALTS]{
		\begin{minipage}[t]{0.13\textwidth}
			\centering
			{\scriptsize{CPU: 65.36}}\vspace{0.5mm}
			\includegraphics[width=2.4cm,height=3.3cm]{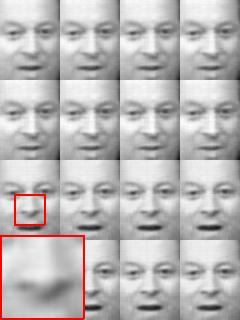}
		\end{minipage}}%\hspace{-2mm}
	\subfigure[TFM-TTP]{	
		\begin{minipage}[t]{0.13\textwidth}
			\centering
			{\scriptsize{CPU: 54.73}}\vspace{0.5mm}
			\includegraphics[width=2.4cm,height=3.3cm]{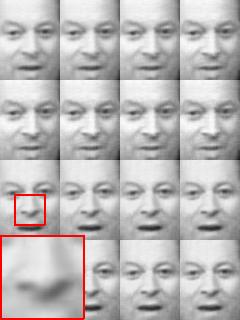}
		\end{minipage}}%\hspace{-2mm}
	\caption{ Recovered images, zoomed regions, and CPU time (in seconds)
of different methods for the AI Gore talking dataset corrupted by salt-and-pepper impulse noise
with noise level 15\% and zero-mean Gaussian noise with variance $\sigma^2=0.05$.}\label{Fig: Gore-mixnoise}
\vspace{-8pt}
\end{figure*}

\subsubsection{Natural Images}

In this subsection, we show the recovery and alignment performance of different methods for the three real-world natural images in Figures \ref{Fig: window}-\ref{Fig: Gore}.
In Figure \ref{Fig: window}, the visual comparisons and zoomed regions of different methods
are shown for the windows dataset.
It can be seen that the alignment performance of  NCALTS and TFM-TTP are better than
that of RASL, t-GRASTA, NQLSD, and $\ell_p$-ADMM, especially for the zoomed regions of these images.
Moreover, the images recovered by TFM-TTP are more clear than those recovered by other methods.
The details of texture of the zoomed regions of TFM-TTP are
kept better than those of RASL, t-GRASTA, NQLSD, $\ell_p$-ADMM, and NCALTS,
where the TFM-TTP preserves sharp edges of the images better.
For the computational time, the RASL requires less CPU time  than other methods.
While its recovery performance is not good in terms of alignment and visual quality.
However, the computational time required by  TFM-TTP is much less than that required by t-GRASTA, NQLSD, $\ell_p$-ADMM, and NCALTS.
In particular, the recovery performance of TFM-TTP is slightly better than that of NCALTS,
while the TFM-TTP is much faster than NCALTS.

In Figure \ref{Fig: digits}, we show the recovered images
and CPU time (in seconds) of different methods for the handwritten number ``8''.
It can be seen that the alignment performance and
recovered images of  NCALTS and TFM-TTP are much better than those of
RASL, t-GRASTA, NQLSD, and $\ell_p$-ADMM. Compared with NCALTS, the TFM-TTP can algin these images slightly better,
see the third row and last column of the figure.
Moreover, the CPU time required by TFM-TTP is less than that required by other methods.

Figure \ref{Fig: Gore} shows the visual comparisons, zoomed regions,
and CPU time (in seconds) of the AI Gore talking dataset recovered by different methods.
It can be seen from the zoomed regions of these images that
the alignment performance of $\ell_p$-ADMM, NCALTS, and TFM-TTP is much better than that of RASL, t-GRASTA, and NQLSD,
which is shown by eyes of the zoomed regions.
Moreover, the TFM-TTP outperforms $\ell_p$-ADMM and NCALTS in terms of the visual quality.
Although the CPU time of RASL is much less than that of other methods,
the recovery and alignment performance of images of RASL are much worse than that of NCALTS and TFM-TTP.
Besides, the TFM-TTP takes less CPU time than NCALTS since computing the SVDs of the frontal slices of a tensor
is much expensive than the transformed tensor-tensor product.
\subsubsection{Natural Images with Salt-and-Pepper Impluse Noise}

In this subsection, we show the performance of different methods for the three real-world image datasets
corrupted by  salt-and-pepper impulse noise in Figures \ref{Fig: window salt}-\ref{Fig: Gore-salt}.
%%we add 20\% `salt and pepper' noise to the three datasets as the test images.
%Then we apply our method and RASL, t-GRASTA, NQLSD, $\ell_p$-ADMM, NCALTS to
%align the natural images with sparse noise,
%and the recovered images of different methods are shown.
Figure \ref{Fig: window salt} shows the recovered images, zoomed regions,
and CPU time (in seconds) of different methods for the natural
window images corrupted by salt-and-pepper impulse noise with noise level $20\%$.
It can be seen that the images recovered by $\ell_p$-ADMM, NCALTS,
and TFM-TTP are aligned better than those recovered by RASL, t-GRASTA, and NQLSD.
Moreover, the recovered images of TFM-TTP are more clear than those of RASL, t-GRASTA,
NQLSD, $\ell_p$-ADMM, and NCALTS,
which can be seen from the left upper corners of the zoomed regions.
Furthermore, the CPU time required by TFM-TTP is much less than that required by other compared methods.

In Figure \ref{Fig: digits salt}, we  show
the recovered images and CPU time (in seconds) of different methods
for the handwritten digits ``8'' corrupted by  salt-and-pepper impulse noise with noise level 10\%.
It can be observed that the images recovered by TFM-TTP are more clear than those recovered by
 RASL, t-GRASTA, NQLSD, $\ell_p$-ADMM, and NCALTS.
 Some handwritten digits ``8'' are missing for the RASL, t-GRASTA, NQLSD, and $\ell_p$-ADMM.
 Moreover, NCALTS and TFM-TTP perform much better that RASL, t-GRASTA, NQLSD, and $\ell_p$-ADMM in terms of image alignment performance.
 Although the CPU time required by RASL is less than that required by TFM-TTP,
 its recovery quality is much worse than that of TFM-TTP.
In addition, the  TFM-TTP requires less CPU time than t-GRASTA, NQLSD, $\ell_p$-ADMM, and NCALTS.
 %\textcolor{red}{Compared to t-GRASTA, NQLSD, $\ell_p$-ADMM, and NCALTS, TFM-TTP requires a lot less CPU time.}

%The CPU time required by TFM-TTP is much less than that required by t-GRASTA, NQLSD, $\ell_p$-ADMM, and NCALTS.
%There are some handwritten digits ``8'' images with missing information
%in the recovered images of RASL, t-GRASTA, NQLSD, and $\ell_p$-ADMM.
%%In the recovered images of RASL, t-GRASTA, NQLSD, and $\ell_p$-ADMM, there exists some handwritten digits ``8'' images with missing information.
%Apparently, our method outperforms RASL, $\ell_p$-ADMM and NCALTS in terms of alignment.
%Futhermore,
%the images recovered by our method are sharper than those recovered by $\ell_p$-ADMM and NCALTS.
%In terms of running time, t-GRASTA takes three times longer than NTF-TTP, and $\ell_p$-ADMM takes more than twice as long as NTF-TTP.
%NQLSD and NCALTS use similar running time, and both are slower than our proposed NTF-TTP.\par
Figure \ref{Fig: Gore-salt} shows the recovered images, zoomed regions, and CPU time (in seconds)
of different methods for  the AI Gore talking dataset corrupted by 20\% salt-and-pepper impulse noise.
We can see from this figure that the images obtained by  TFM-TTP are more clear than those obtained by RASL, t-GRASTA, NQLSD,
$\ell_p$-ADMM, and NCALTS, especially for the zoomed regions of these images,
where the nose of TFM-TTP keeps more details than those of other methods.
Besides, the alignment performance of $\ell_p$-ADMM, NCALTS, TFM-TTP is better than that of RASL, t-GRASTA, and NQLSD.
And the CPU time required by  TFM-TTP is much less than that required by other compared  methods.}
\subsubsection{Natural Images with Mixed Noise}

In this subsection, we show the recovery and alignment performance
of different methods for the three real-world image datasets
in Figures \ref{Fig: window-mixnoise}-\ref{Fig: Gore-mixnoise},
which are corrupted by  salt-and-pepper impulse noise and additive Gaussian noise simultaneously.
Figure \ref{Fig: window-mixnoise} shows the recovered images, zoomed regions, and CPU time (in seconds)
of different methods for the window images corrupted  by  salt-and-pepper impulse noise with noise level 30\%
and  zero-mean Gaussian noise with variance $\sigma^2=0.05$.
It can be seen that the $\ell_p$-ADMM, NCALTS, and  TFM-TTP
can align the images better than RASL, t-GRASTA, and NQLSD.
And the images recovered by TFM-TTP are more clear than those recovered by RASL, t-GRASTA,
 NQLSD,  $\ell_p$-ADMM, and NCALTS, which can be seen from the zoomed regions of these images.
 Furthermore, the CPU time required by TFM-TTP is much less than that required by other compared methods.

In Figure \ref{Fig: digits-mixnoise}, we show the recovered images and CPU time (in seconds)
of different methods for the handwritten digits images corrupted by salt-and-pepper impulse noise
 with noise level 10\% and
and zero-mean Gaussian noise with variance $\sigma^2=0.05$.
It can be observed that the images recovered by TFM-TTP are more clear than those
recovered by RASL, t-GRASTA,
 NQLSD,  $\ell_p$-ADMM, and NCALTS.
 Some handwritten digits images obtained by the RASL, t-GRASTA,
 NQLSD,  and $\ell_p$-ADMM are missing.
 Moreover, the alignment of TFM-TTP is better than that of other methods.
And the computational time of TFM-TTP is much less than that of RASL, t-GRASTA,
 NQLSD,  $\ell_p$-ADMM, and NCALTS.

% We add
%to the as the experimental images.
%Figure (a) shows
%the natural handwritten digits images with mixed noise.
%%and then apply our method with RASL, t-GRASTA, NQLSD, $\ell_p$-ADMM and NCALTS.
%The recovery results by different methods
%are shown in Figure \ref{Fig: digits-mixnoise} (b)-(f).
%It can be observed from Fig \ref{Fig: digits-mixnoise} (b)-(f) that
%the images recovered by NTF-TTP obtain better visual results
%and NTF-TTP perform better alignment than those of NCALTS.
%Some of the images recovered by t-GRASTA and NQLSD have missing information.
%The images recovered by our method are clearer than
%that recovered by RASL and $\ell_p$-ADMM.
%From time consumption, our method is the most efficient.
%Moreover, we can find that
%RASL is as efficient as $\ell_p$-ADMM,
%while t-GRASTA requires the highest cost in this experiment.\par

Figure \ref{Fig: Gore-mixnoise}  shows the recovered images, zoomed regions, and CPU time (in seconds)
of different methods for the AI Gore talking dataset corrupted by salt-and-pepper impulse noise
with noise level 15\% and zero-mean Gaussian noise with variance $\sigma^2=0.05$.
We can see from this figure that the zoomed regions of images recovered by TFM-TTP are much more clear than
those recovered by RASL, t-GRASTA, NQLSD, $\ell_p$-ADMM, and NCALTS,
where the nose of the zoomed region of TFM-TTP is more clear than those of other compared methods.
Furthermore, the alignment of images obtained by NCALTS and TFM-TTP is
better than that obtained by RASL, t-GRASTA, NQLSD, and $\ell_p$-ADMM,
e.g., see the first images of these images.
Besides, the CPU time required by TFM-TTP is much less than that required by other compared methods.

\section{Concluding Remarks}\label{Concludsion}

In this paper, we have proposed a low-rank tensor factorization method
based on transformed tensor-tensor product for aligning a batch of linearly correlated  images,
where the observed images are deformed by some unknown
domain transformations and corrupted by sparse noise and Gaussian noise simultaneously.
By stacking these images as the frontal slices of a third-order tensor, the observed tensor after domain transformations
 can be decomposed into the sum of a low-rank tensor, a sparse tensor, and a noisy tensor with Gaussian distribution.
Then we propose to  factorize the underlying tensor into the
$\mathbf U$-product of two smaller tensors under any unitary transformation to explore its low-rankness,
whose computation is much cheaper than that of the transformed tensor nuclear norm since
one needs to perform the SVDs of a series of matrices in the transformed tensor nuclear norm.
Furthermore, the tensor $\ell_p$ norm $(0<p<1)$ and the tensor Frobenius norm are employed
to characterize the sparsity of the sparse noise
and the distribution of Gaussian noise, respectively.
A generalized Gauss-Newton algorithm is adopted to solve the proposed model via linearizing the domain transformations,
 where  a proximal Gauss-Seidel algorithm is adapted
to solve the resulting subproblem.
The convergence and  convergence rate of the proximal Gauss-Seidel algorithm
are established under very mild conditions.
Numerical examples on many image datasets have shown the effectiveness and efficiency of our TFM-TTP method
when compared to other existing methods.
%our proposed method not only obtains better recovery quality, but also is faster compared to other.

A future research topic is to establish the convergence
of the generalized Gauss-Newton algorithm for solving our nonconvex model.
Moreover, it is also interesting to analyze the error bound
of the proposed model in future work (cf. \cite{Qiu_2021, Zhang_2021, zhang2022sparse}).

%Compared with the traditional method of
%constraining the low-rank tensor with the nuclear norm, our method greatly
%improves the efficiency of the algorithm and avoids performing t-SVD computation.
%We propose to apply non-convex penalty instead of convex penalty to constrain the
%sparsity of the noise, which better removes the noise and also ensures the quality of the recovered image.
%Furthermore, we illustrate the convergence of our proposed method and the
%convergence rate of the algorithm. Numerical experimental results show that
%our method not only obtains good image alignment, but also runs at a
% much higher speed compared with most widely used image alignment methods.
% In the future work, the complexity of the NTF-TTP algorithm needs to be studied.

\section*{Acknowledgements}
The authors are grateful to Dr. Xiai Chen and Prof. Xiaoqin Zhang
for sharing the codes of  NQLSD \cite{Chen_2016} and $\ell_q$-ADMM \cite{Zhang_2021}, respectively.
The authors are also grateful to Prof. Di Wang for helpful discussions about the geometric transformation in image alignment.

\section*{Appendix A}
%arre
In this section, we give the definitions of the subdifferential
and  the  Kurdyka-$\L$ojasiewicz (KL) property of a function, respectively,
which play a vital role for establishing the convergence and rate of convergence  of the proximal Gauss-Seidel algorithm.

\begin{definition}\cite[Definition 8.3]{rockafellar2009variational}
Consider a function $f:\mathbb{R}^n\rightarrow(\infty,+\infty]$
and a point $\overline{\mathbf{x}}$ with $f(\overline{\mathbf{x}})$ finite.
For any $\mathbf{x}\in\mathbb{R}^n$, one says that

(i) $\mathbf{v}$ is a regular subgradient of $f$ at $\overline{\mathbf{x}}$,
written as $\mathbf{v}\in\widehat{\partial} f(\overline{\mathbf{x}})$, if
$$
{\lim\inf}_{\mathbf{x}\rightarrow\overline{\mathbf{x}},\mathbf{x}\neq \overline{\mathbf{x}}}\frac{f(\mathbf{x})-f(\mathbf{x})-\langle \mathbf{v}, \mathbf{x}-\overline{\mathbf{x}} \rangle}{\|\mathbf{x}-\overline{\mathbf{x}}\|}\geq0.
$$

(ii) $\mathbf{v}$ is a subgradient of  $f$ at $\overline{\mathbf{x}}$,
written as $\mathbf{v}\in\partial f(\overline{\mathbf{x}})$,
if there are sequences $\mathbf{x}^k\rightarrow \overline{\mathbf{x}}$,
$f(\mathbf{x}^k)\rightarrow f(\overline{\mathbf{x}})$,
and $\mathbf{v}^k\in \widehat{\partial} f(\mathbf{x}^k)$ with $\mathbf{v}^k\rightarrow \mathbf{v}$.
\end{definition}

\begin{definition}\label{DeKLF}
\cite[Definition 2.1]{Yu2021} %[Kurdyka-$\L$ojasiewicz property]
Let $f:\mathbb R^n\rightarrow \mathbb R\cup\{+\infty\}$ be a proper lower semicontinuous function.
We say that $f$ has the Kurdyka-$\L$ojasiewicz (KL)
property at point $x^{*}\in\text{dom}(\partial f)$,
%with $\varphi(s)=\frac{\mu}{1-\theta}s^{1-\theta}$, $\theta\in[0,1), \mu>0$,
if there exist a neighborhood $U$ of $x^{*}$, $\eta\in(0,+\infty]$ and
a continuous concave function $\varphi : [0,\eta)\rightarrow \mathbb R_{+}$ such that:
%for all $\mathcal Z\in U^{'}\cap [f(Z^{*})<f<f(Z^{*})+\eta^{'}]$ it holds that
(i)   $\varphi(0)=0$;
(ii)   $\varphi$ is $C^1$ on $(0,\eta)$;
(iii)   for all $s\in(0,\eta)$, $\varphi{'}(s)>0$;
(iv)   and for all $x \text{ in } U \cap [f(x^{*})<f<f(x^{*})+\eta]$ the KL inequality holds:
\begin{equation}\label{test316}
\varphi{'}(f(x)-f(x^{*}))\mathop{\mathrm{dist}}(0,\partial f(x)) \geq 1.
\end{equation}
If $f$ satisfies the KL property at $x^*\in\text{dom}(\partial f)$,
and the $\varphi(s)$ in (\ref{test316}) can be
chosen as $\varphi(s)=\hat{c}s^{1-\theta}$ for
some $\hat{c}>0$ and $\theta\in[0,1)$,
then we say that $f$ satisfies the KL property at point $x^{*}$ with exponent $\theta$.
\end{definition}

A proper closed function $f$ satisfying the KL property at every point in $\text{dom}(\partial f)$ is
said to be a KL function, and a proper closed function $f$ satisfying the KL property
with exponent  $[0,1)$ at every point
in $\text{dom}(\partial f)$ is said to be a KL function with exponent $\theta$ \cite{Yu2021}.
\section*{Appendix B. Proof of Theorem 1}

%\section{Proof of Theorem 1}
%\begin{IEEEproof}[Proof of Theorem 1]
%\textbf{Proof of Theorem 1}
It follows from (\ref{eq12}), (\ref{eq13}), (\ref{eq14}) and (\ref{eq15}) that
%\setlengh\jot{9pt}
\begin{equation}\label{test101}
\left\{
             \begin{array}{lr}
             F(\mathcal X^{k+1} ,\mathcal Y^{k},\mathcal S^{k},\Delta\tau^{k})+\frac{\alpha^{k}}{2}\|\mathcal X^{k+1}-\mathcal {X}^{k} \|_{F}^{2}
             \vspace{1ex} \\ \quad \quad \quad \quad \quad \quad \quad \quad \quad \quad \quad
             \leq  F(\mathcal X^{k} ,\mathcal Y^{k},\mathcal S^{k},\Delta\tau^{k}), &   \vspace{1ex}  \\%% \vspace{1ex} 调整行间距，大括号内的数字表示行间距大小
             F(\mathcal X^{k+1} ,\mathcal Y^{k+1},\mathcal S^{k},\Delta\tau^{k})+\frac{\beta^{k}}{2}\|\mathcal Y^{k+1}-\mathcal {Y}^{k} \|_{F}^{2}
             \vspace{1ex}\\ \quad \quad \quad \quad \quad \quad \quad \quad \quad \quad \quad
             \leq  F(\mathcal X^{k+1} ,\mathcal Y^{k},\mathcal S^{k},\Delta\tau^{k}), &  \vspace{1ex}   \\
             F(\mathcal X^{k+1} ,\mathcal Y^{k+1},\mathcal S^{k+1},\Delta\tau^{k})+\frac{\gamma^{k}}{2}\|\mathcal S^{k+1}-\mathcal {S}^{k} \|_{F}^{2}
             \vspace{1ex}\\ \quad \quad \quad \quad \quad \quad \quad \quad \quad \quad \quad
              \leq  F(\mathcal X^{k+1} ,\mathcal Y^{k+1},\mathcal S^{k},\Delta\tau^{k}), &  \vspace{1ex}   \\
             F(\mathcal X^{k+1} ,\mathcal Y^{k+1},\mathcal S^{k+1},\Delta\tau^{k+1})+\frac{\xi^{k}}{2}\| \Delta\tau^{k+1}-{\Delta\tau}^{k} \|_{F}^{2}
             \vspace{1ex}\\ \quad \quad \quad \quad \quad \quad \quad \quad \quad \quad \quad
             \leq F(\mathcal X^{k+1} ,\mathcal Y^{k+1},\mathcal S^{k+1},\Delta\tau^{k}). &  \vspace{1ex}   \\
             \end{array}
\right.
\end{equation}
Summing the inequalities in (\ref{test101}) to obtain that
\begin{equation}\label{test24}
\begin{aligned}
F(\mathcal X^{k+1} ,\mathcal Y^{k+1},\mathcal S^{k+1},\Delta\tau^{k+1})+\frac{\alpha^{k}}{2}\|\mathcal X^{k+1}-\mathcal {X}^{k} \|_{F}^{2}\\
+\frac{\beta^{k}}{2}\|\mathcal Y^{k+1}-\mathcal {Y}^{k} \|_{F}^{2}
+\frac{\gamma^{k}}{2}\|\mathcal S^{k+1}-\mathcal {S}^{k} \|_{F}^{2}\\
+\frac{\xi^{k}}{2}\| \Delta\tau^{k+1}-{\Delta\tau}^{k} \|_{F}^{2}\leq F(\mathcal X^{k} ,\mathcal Y^{k},\mathcal S^{k},\Delta\tau^{k}).
\end{aligned}
\end{equation}
Moreover, the optimality conditions of (\ref{eq12}), (\ref{eq13}), (\ref{eq14}) and (\ref{eq15})  yield
\begin{equation}\label{test102}
\left\{
             \begin{array}{lr}
             0= \nabla_{\mathcal X} Q(\mathcal X^{k+1} ,\mathcal Y^{k},\mathcal S^{k},\Delta\tau^{k})   \\ \quad \quad \quad \quad \quad \quad \quad \quad
             + \rho_1\mathcal X^{k+1}+\alpha^{k}(\mathcal X^{k+1}-\mathcal {X}^{k}),&   \vspace{1ex}             \\
             0= \nabla_{\mathcal Y} Q(\mathcal X^{k+1} ,\mathcal Y^{k+1},\mathcal S^{k},\Delta\tau^{k}) \\ \quad \quad \quad \quad \quad \quad \quad \quad
             +\rho_2\mathcal Y^{k+1}+\beta^{k}(\mathcal Y^{k+1}-\mathcal {Y}^{k}), &   \vspace{1ex}             \\
             0\in\nabla_{\mathcal S} Q(\mathcal X^{k+1} ,\mathcal Y^{k+1},\mathcal S^{k+1},\Delta\tau^{k})
             \vspace{1ex} \\ \quad \quad \quad \quad \quad \quad \quad \quad
             +\partial(\lambda{\|\mathcal S^{k+1} \|}_{p}^{p})+\gamma^{k}(\mathcal S^{k+1}-\mathcal {S}^{k}), &   \vspace{1ex}             \\
             0=\nabla_{\Delta\tau} Q(\mathcal X^{k+1} ,\mathcal Y^{k+1},\mathcal S^{k+1},\Delta\tau^{k+1})\\ \quad \quad \quad \quad \quad \quad \quad \quad
             +\rho_{3}\Delta\tau^{k+1}+\xi^{k}(\Delta\tau^{k+1}-{\Delta\tau}^{k}). &   \vspace{1ex}
             \end{array}
\right.
\end{equation}
Let
\begin{equation}\label{test106}
\left\{
\begin{array}{lr}
\mathcal V_{\mathcal X}^{k+1}:=-\nabla_{\mathcal X} Q(\mathcal X^{k+1} ,\mathcal Y^{k},\mathcal S^{k},\Delta\tau^{k})-\alpha^{k}(\mathcal X^{k+1}-\mathcal{X}^{k}), \vspace{1ex}\\
\mathcal V_{\mathcal Y}^{k+1}:=-\nabla_{\mathcal Y} Q(\mathcal X^{k+1} ,\mathcal Y^{k+1},\mathcal S^{k},\Delta\tau^{k})-\beta^{k}(\mathcal Y^{k+1}-\mathcal {Y}^{k}), \vspace{1ex}\\
\mathcal V_\mathcal S^{k+1}  :=-\nabla_{\mathcal S } Q(\mathcal X^{k+1} ,\mathcal Y^{k+1},\mathcal S^{k+1},\Delta\tau^{k})-\gamma^{k}(\mathcal S^{k+1}-\mathcal {S}^{k}), \vspace{1ex}\\
\mathcal V_{\Delta\tau}^{k+1}:=-\nabla_{\Delta\tau} Q(\mathcal X^{k+1} ,\mathcal Y^{k+1},\mathcal S^{k+1},\Delta\tau^{k+1})   \\ \quad\quad\quad\quad\quad\quad\quad\quad\quad\quad\quad
 -\xi^{k}(\Delta\tau^{k+1}-{\Delta\tau}^{k}). \vspace{1ex}
%\\ \quad \quad \quad \quad \quad \quad \quad \quad \quad \quad \quad
\end{array}
\right.
\end{equation}

Combining (\ref{test102}) and (\ref{test106}), we obtain that
\begin{equation}\label{test107}
\begin{aligned}
 \mathcal V_{\mathcal X }^{k+1}&= \rho_{1}\mathcal X^{k+1}, &
 \mathcal V_{\mathcal Y }^{k+1}&=\rho_{2}\mathcal Y^{k+1},    \\
 \mathcal V_\mathcal S^{k+1}   &\in \partial(\lambda{\|\mathcal S^{k+1}\|}_{p}^{p}),  &
 \mathcal V_{\Delta\tau}^{k+1} &=\rho_{3}\Delta\tau^{k+1}.
\end{aligned}
\end{equation}

Notice that $0<\underline d<\overline d< +\infty $ and
$\underline d \leq \alpha^k,\beta^k,\gamma^k,\xi^k  \leq \overline d $.
Therefore, the sequence $\{(\mathcal X^{k} ,\mathcal Y^{k},\mathcal S^{k},\Delta\tau^{k})\}$
satisfies
\begin{equation}\label{test103}
\left\{
             \begin{array}{lr}
             \|\nabla_{\mathcal X} Q(\mathcal X^{k+1} ,\mathcal Y^{k},\mathcal S^{k},\Delta\tau^{k})+  \mathcal V_{\mathcal X }^{k+1}\|_F
             \\ \quad \quad \quad \quad \quad \quad \quad \quad \quad \quad \quad
             \leq \overline d\| \mathcal X^{k+1}-\mathcal {X}^{k}\|_F,  &   \vspace{1ex}             \\
             \|\nabla_{\mathcal Y} Q(\mathcal X^{k+1} ,\mathcal Y^{k+1},\mathcal S^{k},\Delta\tau^{k})+\mathcal V_{\mathcal Y }^{k+1}\|_F
             \\ \quad \quad \quad \quad \quad \quad \quad \quad \quad \quad \quad
             \leq \overline d\| \mathcal Y^{k+1}-\mathcal {Y}^{k}\|_F,  &   \vspace{1ex}             \\
             \|\nabla_{\mathcal S} Q(\mathcal X^{k+1} ,\mathcal Y^{k+1},\mathcal S^{k+1},\Delta\tau^{k})+\mathcal V_\mathcal S^{k+1}\|_F
             \\ \quad \quad \quad \quad \quad \quad \quad \quad \quad \quad \quad
             \leq \overline d\|\mathcal S^{k+1}-\mathcal {S}^{k}\|_F,  &   \vspace{1ex}             \\
             \|\nabla_{\Delta\tau} Q(\mathcal X^{k+1} ,\mathcal Y^{k+1},\mathcal S^{k+1},\Delta\tau^{k+1})+\mathcal V_{\Delta\tau}^{k+1} \|_F
             \\ \quad \quad \quad \quad \quad \quad \quad \quad \quad \quad \quad
             \leq \overline d\|\Delta\tau^{k+1}-{\Delta\tau}^{k}\|_F. &   \vspace{1ex}
             \end{array}
\right.
\end{equation}

Furthermore, it follows from  \cite[Example 4]{Bolte2013}
that the $\ell_p$ norm ($p>0$) is a semi-algebraic function when $p$ is rational.
In particular, when $p$ is taken to be 2, the $\ell_p$ norm is just the tensor Frobenius norm.
As a consequence,
$F(\mathcal X ,\mathcal Y,\mathcal S,\Delta\tau)$ in (\ref{test23}) is a semi-algebraic function \cite[Example 2]{Bolte2013}.
Note that $F(\mathcal X ,\mathcal Y,\mathcal S,\Delta\tau)$  is proper and continuous.
Then we can get that $F(\mathcal X ,\mathcal Y,\mathcal S,\Delta\tau)$ is a $\emph{KL}$ function \cite[Theorem 3]{Bolte2013}.
%Note that $F$ is lower bounded and continuous
%\begin{equation}\label{test100}
%F(\mathcal X ,\mathcal Y,\mathcal S,\Delta\tau) \geq 0,  \quad \text{for any }(\mathcal X ,\mathcal Y,\mathcal S,\Delta\tau),
%\end{equation}
%which implies that $F$ is lower bounded.

%Next, we prove that $F$ is coercive.
By the definition of $F(\mathcal X ,\mathcal Y,\mathcal S,\Delta\tau)$ in (\ref{test23}),
one can easily obtain that $F(\mathcal X ,\mathcal Y,\mathcal S,\Delta\tau)$ is bounded from below and tends to infinity
as $\|( \mathcal X,\mathcal Y,\mathcal S,\Delta\tau) \|_{F}$ tends to infinity.
Hence, we can deduce that $F(\mathcal X ,\mathcal Y,\mathcal S,\Delta\tau)$ is coercive.
%, i.e.,
%\begin{equation}\label{test104}
%\lim\limits_{\|( \mathcal X,\mathcal Y,\mathcal S,\Delta\tau) \|_{F}\rightarrow \infty} F(\mathcal X ,\mathcal Y,\mathcal S,\Delta\tau) \rightarrow \infty.
%\end{equation}
Now we show that the sequence $\{(\mathcal X^{k} ,\mathcal Y^{k},\mathcal S^{k},\Delta\tau^{k})\} $ generated by Algorithm 2 is bounded.
 Assume that the sequence $\{(\mathcal X^{k} ,\mathcal Y^{k},\mathcal S^{k},\Delta\tau^{k})\} $ is unbounded,
 there exists a subsequence $\{(\mathcal X^{k_i} ,\mathcal Y^{k_i},\mathcal S^{k_i},\Delta\tau^{k_i})\}$,
 which tends to infinity
 as $i$ tends to infinity.
By the coercivity of $F$, we obtain that $F(\mathcal X^{k_i} ,\mathcal Y^{k_i},\mathcal S^{k_i},\Delta\tau^{k_i})$
tends to infinity as $i$ tends to infinity, which is contradictory to (\ref{test24}).
Therefore, we get that
the sequence $\{(\mathcal X^{k} ,\mathcal Y^{k},\mathcal S^{k},\Delta\tau^{k})\} $ is bounded.
It follows from
\cite[Theorem 6.2]{Attouch_2011} that the desired conclusion is attained.
%\end{IEEEproof}
%\qed
\section*{Appendix C. Proof of Theorem 2}
%\section{Proof of Theorem 2}
\begin{lemma}\label{thm3}%Choose a starting point $\mathcal Z^{0}$.
Let $\{\mathcal Z^{k}\}$ be a sequence generated by Algorithm 2.  %\ref{alg:algorithm2},
whose limit point is denoted by  $\mathcal{Z}^*$.
Suppose that $p$ is rational.
Then the following statements hold: \par
(i) There exists a constant $C>0$ such that
$$
%{\big\|w^{k+1}\big\|}_{F}\leq M{\big\|\mathcal Z^{k}-\mathcal Z^{k+1}\big\|}_{F}.
\textup{dist}(0,\partial F(\mathcal Z^{k+1}))\leq C{\|\mathcal Z^{k+1}-\mathcal Z^{k}\|}_{F}.
$$

(ii) There exist $\rho>0$, $\mu>0$ and a KL exponent $\theta\in[0,1)$ of $F$ such that
% $\theta$ and $\mu$ are defined as in Definition 4.1. Then
%$\rho=\frac{\min \{\alpha_{k},\beta_{k},\gamma_{k},\xi_{k}\}}{2M}$.
%There exist an exponent $\theta\in (0,1)$ and a positive constant $\mu$ such that%\quad If $\theta\in(\frac{1}{2},1)$, then there exist $\eta >0$ such that\\
\begin{equation}\label{test116}
\begin{aligned}
&\sum\limits_{j=k}^{\infty}\|\mathcal Z^{j+1}-\mathcal Z^{j}\|_{F}\\
&\leq \| \mathcal Z^{k}-\mathcal {Z}^{k-1}\|_F+ \frac{\mu}{\rho(1-\theta)}(F(\mathcal Z^{k})-F(\mathcal Z^{*}))^{1-\theta}.
\end{aligned}
\end{equation}

(iii) We have
$
\sum_{k=0}^{\infty}\|\mathcal Z^{k+1}-\mathcal Z^{k}\|_{F}<+\infty,
$
hence $\lim\limits_{k\rightarrow\infty}\|\mathcal Z^{k+1}-\mathcal Z^{k}\|_{F}=0$.

(iv) For any positive  integer $k$, it holds that
\begin{equation}\label{test401}
\sum_{j=k}^{\infty}\|\mathcal Z^{j+1}-\mathcal Z^{j}\|_{F} \geq \|\mathcal Z^{k}-\mathcal Z^{*}\|_{F}.
\end{equation}
\end{lemma}

\begin{IEEEproof}[Proof](i)
%\textbf{Proof. }(i)
%Note that
%\begin{equation}\label{test127}
%\text{dist}(0,\partial F(\mathcal Z^{k+1}))=\inf\left\{\|  \mathcal{A} \|_F,  \forall \mathcal{A}\in \partial F(\mathcal Z^{k+1})\right\}.
%\end{equation}
% It is difficult to solve the distance function $\text{dist}(0,\partial F(\mathcal Z^{k+1}))$ directly. Instead, we find an upper bound of the function. Next, we present the upper bound.
%By (\ref{test107}) and the optimality conditions (\ref{test102}), we can easily get
The subdifferential of (\ref{test23}) at the $(k+1)$th iteration  yields
\begin{eqnarray*}
\left\{
             \begin{array}{lr}
             \rho_1\mathcal X^{k+1}+\nabla_{\mathcal X} Q(\mathcal X^{k+1} ,\mathcal Y^{k+1},\mathcal S^{k+1},\Delta\tau^{k+1})
             \\~~~~~~~~~~~~~~~~~~=
             \nabla_{\mathcal X} F(\mathcal X^{k+1} ,\mathcal Y^{k+1},\mathcal S^{k+1},\Delta\tau^{k+1}),  &\\
             \rho_2\mathcal Y^{k+1}+\nabla_{\mathcal Y} Q(\mathcal X^{k+1} ,\mathcal Y^{k+1},\mathcal S^{k+1},\Delta\tau^{k+1})\\~~~~~~~~~~~~~~~~~~=
             \nabla_{\mathcal Y} F(\mathcal X^{k+1} ,\mathcal Y^{k+1},\mathcal S^{k+1},\Delta\tau^{k+1}), &\\
             \mathcal S_1^{k+1}\hspace{1mm}+\nabla_{\mathcal S} Q(\mathcal X^{k+1} ,\mathcal Y^{k+1},\mathcal S^{k+1},\Delta\tau^{k+1}) \\~~~~~~~~~~~~~~~~~~
             \in \partial_{\mathcal S} F(\mathcal X^{k+1} ,\mathcal Y^{k+1},\mathcal S^{k+1},\Delta\tau^{k+1}),   &\\
             \rho_3\Delta\tau^{k+1}+\nabla_{\Delta\tau} Q(\mathcal X^{k+1} ,\mathcal Y^{k+1},\mathcal S^{k+1},\Delta\tau^{k+1})\\~~~~~~~~~~~~~~~~~~ =
             \nabla_{\Delta\tau} F(\mathcal X^{k+1} ,\mathcal Y^{k+1},\mathcal S^{k+1},\Delta\tau^{k+1}),& \\
             \end{array}
\right.
\end{eqnarray*}
where $\mathcal S_1^{k+1} \in \partial(\lambda{\|\mathcal S^{k+1}\|}_{p}^{p})$ and $Q$ is defined as (\ref{eq11}).
%where $\mathcal V_{\mathcal X }^{k+1}, \mathcal V_{\mathcal Y }^{k+1}, \mathcal V_\mathcal S^{k+1}$ are defined as (\ref{test107}).
Let $Q^{k+1}=Q(\mathcal X^{k+1} ,\mathcal Y^{k+1},\mathcal S^{k+1},\Delta\tau^{k+1})$.
Then, by \cite[Proposition 2.1]{Attouch_2010}, we get that
\begin{equation}\label{test109}
\begin{split}
 w^{k+1}:=(\rho_1\mathcal X^{k+1}+\nabla_{\mathcal X} Q^{k+1}, \rho_2\mathcal Y^{k+1}
+\nabla_{\mathcal Y} Q^{k+1},\\
\mathcal S_1^{k+1}+\nabla_{\mathcal S} Q^{k+1}, \rho_3\Delta\tau^{k+1}+\nabla_{\Delta\tau} Q^{k+1}) \\
\in \partial F(\mathcal X^{k+1} ,\mathcal Y^{k+1},\mathcal S^{k+1},\Delta\tau^{k+1}).
\end{split}
\end{equation}
%where $Q^{k+1}=Q(\mathcal X^{k+1} ,\mathcal Y^{k+1},\mathcal S^{k+1},\Delta\tau^{k+1})$ and
%Here (\ref{test109}) implies that $w^{k+1}$ is an upper bound of the function $\text{dist}(0,\partial F(\mathcal Z^{k+1}))$. %Therefore, we obtain from (\ref{test109}) and (\ref{test127}) that
Hence, we deduce
\begin{equation}\label{test110}
\text{dist}(0,\partial F(\mathcal Z^{k+1}))\leq \| w^{k+1} \|_F.
\end{equation}
%By (\ref{test110}), we know that $w^{k+1}$ is an upper bound of $\text{dist}(0,\partial F(\mathcal Z^{k+1}))$.appearing in (\ref{eq11})
%Next, we prove that (\ref{test115}) is true by proving ${\big\|w^{k+1}\big\|}_{F}\leq M{\big\|\mathcal Z^{k}-\mathcal Z^{k+1}\big\|}_{F}$.
It follows from Theorem 1
that the sequence $\{\mathcal Z^{k}\}$ is bounded.
Therefore, there exists a compact set $\mathfrak{D}$
such that ${\mathcal Z^{k}\subset \mathfrak{D}}$ for any $k$.
 Note that $Q$ in (\ref{eq11}) is a quadratic function,
 then $\nabla Q$ is Lipschitz continuous on any bounded subset of $\mathfrak{D}$ with a Lipschitz constant $L$.
 Hence, we obtain that
\begin{equation}\label{test120}
\begin{split}
&~~~\|\nabla_{\mathcal X} Q(\mathcal X^{k+1} ,\mathcal Y^{k+1},\mathcal S^{k+1},\Delta\tau^{k+1})+\rho_1\mathcal X^{k+1}\|_F \\
&\leq \|\nabla_{\mathcal X} Q(\mathcal X^{k+1} ,\mathcal Y^{k},\mathcal S^{k},\Delta\tau^{k})+\rho_1\mathcal X^{k+1}\|_F \\
&~~~~~~~~~+\|\nabla_{\mathcal X} Q(\mathcal X^{k+1} ,\mathcal Y^{k+1},\mathcal S^{k+1},\Delta\tau^{k+1})\\
&~~~~~~~~~~~~-\nabla_{\mathcal X} Q(\mathcal X^{k+1} ,\mathcal Y^{k},\mathcal S^{k},\Delta\tau^{k})\|_F \\
&\leq \overline d\| \mathcal X^{k+1}-\mathcal {X}^{k}\|_F+\|\nabla_{\mathcal X} Q(\mathcal X^{k+1} ,\mathcal Y^{k+1},\mathcal S^{k+1},\Delta\tau^{k+1})\\
&~~~~~~~~~~~-\nabla_{\mathcal X} Q(\mathcal X^{k+1} ,\mathcal Y^{k},\mathcal S^{k},\Delta\tau^{k})\|_F\\
&\leq \overline d\| \mathcal X^{k+1}-\mathcal {X}^{k}\|_F+\|\nabla Q(\mathcal X^{k+1} ,\mathcal Y^{k+1},\mathcal S^{k+1},\Delta\tau^{k+1})\\
&~~~~~~~~~~~-\nabla Q(\mathcal X^{k+1} ,\mathcal Y^{k},\mathcal S^{k},\Delta\tau^{k})\|_F\\
&\leq \overline d\| \mathcal X^{k+1}-\mathcal {X}^{k}\|_F+L\|(\mathcal Y^{k+1},\mathcal S^{k+1},\Delta\tau^{k+1})\\
&~~~~~~~~~~~-(\mathcal Y^{k},\mathcal S^{k},\Delta\tau^{k})\|_F\\
&\leq\overline d\| \mathcal X^{k+1}-\mathcal {X}^{k}\|_F+L\|\mathcal Z^{k+1}-\mathcal {Z}^{k}\|_F \\
&\leq(\overline d+L)\| \mathcal Z^{k+1}-\mathcal {Z}^{k}\|_F,
\end{split}
\end{equation}
where the second inequality holds by (\ref{test103}).
In a similar vein, one can get
\begin{equation}\label{Subdfhhy}
\begin{split}
&\|\nabla_{\mathcal Y} Q(\mathcal X^{k+1} ,\mathcal Y^{k+1},\mathcal S^{k+1},\Delta\tau^{k+1})+\rho_2\mathcal Y^{k+1}\|_F\\
&\leq(\overline d+L)\| \mathcal Z^{k+1}-\mathcal {Z}^{k}\|_F, \\
&\|\nabla_{\mathcal S} Q(\mathcal X^{k+1} ,\mathcal Y^{k+1},\mathcal S^{k+1},\Delta\tau^{k+1})+\mathcal S_1^{k+1}\|_F\\
&\leq(\overline d+L)\| \mathcal Z^{k+1}-\mathcal {Z}^{k}\|_F, \\
&\|\nabla_{\Delta\tau} Q(\mathcal X^{k+1} ,\mathcal Y^{k+1},\mathcal S^{k+1},\Delta\tau^{k+1})+\rho_3 \Delta\tau^{k+1}\|_F\\
&\leq\overline d\| \mathcal Z^{k+1}-\mathcal {Z}^{k}\|_F.
\end{split}
\end{equation}
Combining (\ref{test109}), (\ref{test110}), (\ref{test120}) and (\ref{Subdfhhy}), we get that
%$$
%{\|w^{k+1}\|}_{F}\leq C\| \mathcal Z^{k+1}-\mathcal {Z}^{k}\|_F,
%$$
%
%As a consequence, we obtain
\begin{equation}\label{test128}
\text{dist}(0,\partial F(\mathcal Z^{k+1}))\leq\|w^{k+1}\|_{F}\leq C\| \mathcal Z^{k+1}-\mathcal {Z}^{k}\|_F,
\end{equation}
where $C=4\overline d+3L$.

%where $w^{k+1}$ is defined as (\ref{test109}). Thus, (\ref{test115}) is proven to be true.\\
(ii) %Assume that $\theta$ belongs to $(0,1)$.
Define
$
h(s):=\frac{\mu}{1-\theta}(s-F(\mathcal Z^{*}))^{1-\theta},\ \forall  s\geq F(\mathcal Z^{*}),
$
where $\mu>0$ is a constant and $\theta\in(0,1)$.
Observe that $h(s)$ is a concave function for $s\geq F(\mathcal Z^{*})$,
and its derivative function is $h'(s)=\frac{\mu}{(s-F(\mathcal Z^{*}))^{\theta}}$ for $s>F(\mathcal Z^{*}).$
It then follows from the concavity of the function $h(s)$ that
\begin{equation}\label{test122}
\begin{aligned}
h(F(\mathcal Z^{k}))-h(F(\mathcal Z^{k+1}))
\geq h{'}(F(\mathcal Z^{k}))(F(\mathcal Z^{k})-F(\mathcal Z^{k+1}))\\
=\frac{\mu}{(F(\mathcal Z^{k})-F(\mathcal Z^{*}))^{\theta}}(F(\mathcal Z^{k})-F(\mathcal Z^{k+1})).
\end{aligned}
\end{equation}
Note that $F$ in (\ref{test23}) is a proper closed semi-algebraic function when $p$ is rational,
whose proof is given in Theorem 1.
Therefore, we know that $F$ satisfies the KL property at
$\mathcal Z^{*}$ with $\varphi(s)={\mu}s^{1-\theta}$ \cite{Attouch_2010},
where $\varphi$ comes from Definition \ref{DeKLF},
 and $\theta\in[0,1)$, $\mu>0$. Notice that the KL inequality in (\ref{test316}) yields
%$\theta$ and $\mu$ are defined as in this lemma. Then the KL inequality yields
\begin{align}\label{test126}
(F(\mathcal Z^k)-F(\mathcal Z^{*}))^{\theta}\leq\mu(1-\theta) \mathop{\mathrm{dist}}(0,\partial F(\mathcal Z^k))\\ \nonumber
\leq \mu C\| \mathcal Z^{k}-\mathcal {Z}^{k-1}\|_F,
\end{align}
where the last inequality holds by (\ref{test128}) and $\theta\in[0,1)$.

On the other hand, it follows from (\ref{test24}) that
\begin{equation}\label{test121}
\begin{aligned}
&F(\mathcal Z^{k})-F(\mathcal Z^{k+1})\\
\geq   \ & \frac{\alpha^{k}}{2}\|\mathcal X^{k+1}-\mathcal {X}^{k} \|_{F}^{2}+\frac{\beta^{k}}{2}\|\mathcal Y^{k+1}-\mathcal {Y}^{k} \|_{F}^{2}\\
 & +\frac{\gamma^{k}}{2}\|\mathcal S^{k+1}-\mathcal {S}^{k} \|_{F}^{2}+\frac{\xi^{k}}{2} \|\Delta\tau^{k+1}-{\Delta\tau}^{k}\|_{F}^{2} \\
\geq & \ \frac{\delta^k}{2}\| \mathcal Z^{k+1}-\mathcal {Z}^{k}\|_F^2,
\end{aligned}
\end{equation}
where $\delta^k:=\text{min}\{\alpha^{k},\beta^{k},\gamma^{k},\xi^{k}\}\geq \underline{d}$.
%\begin{align}\label{test125}
%F(\mathcal Z^{k})-F(\mathcal Z^{k+1}).
%\end{align}
%Combining (\ref{test126}), (\ref{test128}), (\ref{test122}) and (\ref{test125}) gives
%Therefore, we have from (\ref{test130}) and (\ref{test125})
In light of (\ref{test122}), (\ref{test126}), and (\ref{test121}), we have
\begin{equation*}\label{test123}
\begin{aligned}
h(F(\mathcal Z^{k}))-h(F(\mathcal Z^{k+1}))
\geq\frac{\delta^k \| \mathcal Z^{k+1}-\mathcal {Z}^{k}\|_F^2}{2C\| \mathcal Z^{k}-\mathcal {Z}^{k-1}\|_F},
\end{aligned}
\end{equation*}
which implies that
\[
\begin{split}
&2\| \mathcal Z^{k+1}-\mathcal {Z}^{k}\|_F\\
&\leq 2\Big(\frac{2C}{\delta^k}\Big)^{\frac{1}{2}}\Big(h(F(\mathcal Z^{k}))-h(F(\mathcal Z^{k+1}))\Big)^{\frac{1}{2}}\| \mathcal Z^{k}-\mathcal {Z}^{k-1}\|_F^{\frac{1}{2}} \\
&\leq \frac{2C}{\delta^k}\Big(h(F(\mathcal Z^{k}))-h(F(\mathcal Z^{k+1}))\Big)+\| \mathcal Z^{k}-\mathcal {Z}^{k-1}\|_F,
\end{split}
\]
where the second inequality holds by the fact that $a^2+b^2\geq 2ab$.
Summing the above inequality from $k$ to infinity to obtain
\begin{equation*}
\begin{aligned}
\sum\limits_{j=k}^{\infty}\frac{2C}{\delta^k}\Big(h(F(\mathcal Z^{j}))-h(F(\mathcal Z^{j+1}))\Big)+\sum\limits_{j=k}^{\infty}\| \mathcal Z^{j}-\mathcal {Z}^{j-1}\|_F\\
\geq2\sum\limits_{j=k}^{\infty}\| \mathcal Z^{j+1}-\mathcal {Z}^{j}\|_F.
\end{aligned}
\end{equation*}
Therefore, we deduce that%{\color{red}{using \cite[Inference 6.4]{Zhou1995a}}}
\begin{equation*}
\begin{aligned}
&\sum\limits_{j=k}^{\infty}{\big\|\mathcal Z^{j+1}-\mathcal Z^{j}\big\|}_{F}\\
&\leq \| \mathcal Z^{k}-\mathcal {Z}^{k-1}\|_F+\frac{2C}{\delta^k}\sum\limits_{j=k}^{\infty}(h(F(\mathcal Z^{j}))-h(F(\mathcal Z^{j+1})))\\
&=\| \mathcal Z^{k}-\mathcal {Z}^{k-1}\|_F+\frac{2C}{\delta^k}\Big(h(F(\mathcal Z^{k}))-\lim_{j\rightarrow \infty}h(F(\mathcal Z^{j}))\Big)\\
&\leq\| \mathcal Z^{k}-\mathcal {Z}^{k-1}\|_F+\frac{2C}{\delta^k}h(F(\mathcal Z^{k}))\\
&=\| \mathcal Z^{k}-\mathcal {Z}^{k-1}\|_F+\frac{2C\mu}{\delta^k(1-\theta)}(F(\mathcal Z^{k})-F(\mathcal Z^{*}))^{1-\theta} \\
&\leq \| \mathcal Z^{k}-\mathcal {Z}^{k-1}\|_F+\frac{\mu}{\rho(1-\theta)}(F(\mathcal Z^{k})-F(\mathcal Z^{*}))^{1-\theta},
\end{aligned}
\end{equation*}
where  $\rho:=\frac{\underline{d}}{2C}$.

(iii) Setting $k=1$ in (\ref{test116}) yields
\begin{equation}\label{test800}
\begin{split}
\sum\limits_{j=0}^{\infty}{\|\mathcal Z^{j+1}-\mathcal Z^{j}\|}_{F}
=\sum\limits_{j=1}^{\infty}{\|\mathcal Z^{j+1}-\mathcal Z^{j}\|}_{F}+\| \mathcal Z^{1}-\mathcal {Z}^{0}\|_F \\
\leq 2\| \mathcal Z^{1}-\mathcal {Z}^{0}\|_F+ \frac{\mu}{\rho(1-\theta)}(F(\mathcal Z^{1})-F(\mathcal Z^{*}))^{1-\theta}<+\infty,
\end{split}
\end{equation}
where $\mathcal Z^{*}$ is a critical point of $F$.

(iv) Define $g_{k}(\mathcal{Z}):=\mathcal{Z}^{k+1}-\mathcal{Z}^{k}$.
From Theorem 1, we obtain that $\{\mathcal{Z}^{k+1}\}$ is bounded.
Denote $E$ to be a bounded measurable set
and $\{\mathcal{Z}^{k+1}\}\subset E$.
Thus $g_{k}(\mathcal{Z})$ is a measurable function on $E$.
In addition, $\|g_{k}(\mathcal{Z})\|_{F}=\|\mathcal{Z}^{k+1}-\mathcal{Z}^{k}\|_{F}<\infty$
for any $k=1,2,\ldots$
%Using Lemma \ref{thm3}(3) we have
%$\sum\limits_{k=1}^{\infty}\left\|\mathcal{Z}^{k+1}-\mathcal{Z}^{k}\right\|_{F}=\sum\limits_{k=1}^{\infty}\left\|g_{k}(\mathcal{Z})\right\|_{F}<\infty$.
Note that, for any positive integer $n$, we get that
$
\sum_{k=1}^{n} g_{k}(\mathcal{Z})=\sum_{k=1}^{n}\left(\mathcal{Z}^{k+1}-\mathcal{Z}^{k}\right)= \mathcal{Z}^{n+1}-\mathcal{Z}^{1}.
$
Therefore, $\sum_{k=1}^{n} g_{k}(\mathcal{Z})$ converges
to $\mathcal{Z}^{*}-\mathcal{Z}^{1}$ as $n$ tends to infinity,
 which follows from the fact that the sequence $\{\mathcal{Z}^{k}\}$ converges to $\mathcal Z^{*}$.
%It is known that the sequence $\{\mathcal{Z}^{k}\}$ converges to $\mathcal Z^{*}$.
%Thus $\sum_{k=1}^{n} g_{k}(\mathcal{Z})$ converges to $\mathcal{Z}^{*}-\mathcal{Z}^{1}$.
%Thus we obtain
%that $\sum_{k=1}^{\infty} g_{k}(\mathcal{Z})$ converges almost everywhere on $E$.
Combining \cite[Corollary 6.4]{Zhou1995a}
with $\mathcal Z^{k}\rightarrow \mathcal Z^{*}, k\rightarrow\infty$, we get
\[
\begin{split}
\sum_{j=k}^{\infty}\left\|\mathcal{Z}^{j+1}-\mathcal{Z}^{j}\right\|_{F}
&\geq \Big\|\sum_{j=k}^{\infty}(\mathcal{Z}^{j+1}-\mathcal{Z}^{j})\Big\|_{F}\\
&=\Big\|\lim _{n \rightarrow \infty}\sum_{j=k}^{n}(\mathcal{Z}^{j+1}-\mathcal{Z}^{j})\Big\|_{F}\\
&=\left\|\lim _{n \rightarrow \infty}\mathcal{Z}^{n+1}-\mathcal Z^{k}\right\|_{F}=\|\mathcal Z^{k}-\mathcal Z^{*}\|_{F}.
\end{split}
\]
The proof is completed. %\qed
\end{IEEEproof}
%\begin{theorem}\label{thm4}
%Let $\{\mathcal Z^{k}\}$ be a sequence generated by Algorithm 2,
%whose limit point is denoted by $\mathcal Z^{*}$,
%and $\theta\in[0,1)$ be  the KL exponent of $F$.
%Suppose that $p$ is rational.
%Then the following assertions hold:
%
%(i)  If $\theta=0$, then the sequence $\{\mathcal Z^{k}\}$  converges in a finite number of steps.\par
%(ii)  If $\theta\in(0,\frac{1}{2}]$, then the sequence $\{\mathcal Z^{k}\}$  converges R-linearly, i.e.,
%there exist $c >0$ and $\vartheta\in (0,1)$ such that
%$
%\|\mathcal Z^{k}-\mathcal Z^{*}\|_{F} \leq c \vartheta^k.
%$
%\par
%(iii)  If $\theta\in(\frac{1}{2},1)$, then the sequence $\{\mathcal Z^{k}\}$  converges R-sublinearly,
%i.e., there exists $c >0$ such that
%$
%\|\mathcal Z^{k}-\mathcal Z^{*}\|_{F} \leq c k^{-\frac{1-\theta}{2\theta-1}}.
%$
% \end{theorem}
\begin{IEEEproof}[Proof of Theorem 2]
%\textbf{Proof of Theorem 2}
(i)
When $\theta=0$, %we adopt the convention $0^0=0$ (see \cite[Remark 4]{Attouch_2007}),
we must have $F(\mathcal Z^{k_0})=F(\mathcal Z^{*})$ for some $k_0$.
Otherwise, for sufficiently large $k$, we have $F(\mathcal Z^{k})>F(\mathcal Z^{*})$.
It follows from (\ref{test126}) that $\mu \mathop{\mathrm{dist}}(0,\partial F(\mathcal Z^{k}))\geq 1$ for all $k$,
which is impossible since $\mathcal Z^{k}\rightarrow \mathcal Z^{*}$ and $0\in\partial F(\mathcal Z^{*})$.
Therefore, there exists some $k_0$ such that $F(\mathcal Z^{k_0})=F(\mathcal Z^{*})$.
Since $F$ is monotonically non-increasing, we have that
 $\mathcal Z^{k}=\mathcal Z^{k_0}=\mathcal Z^{*}$ for all $k>k_0$.

(ii)
Let $\Delta_k:=\sum_{j=k}^{\infty}\|\mathcal Z^{j+1}-\mathcal Z^{j}\|_{F}$.
It follows from Lemma \ref{thm3}(iv) that $\Delta_k \geq \|\mathcal Z^{k}-\mathcal Z^{*}\|_{F}$.
%From the triangle inequality, it is easy to get $\Delta_k \geq {\|\mathcal Z^{k}-\mathcal Z^{\infty}\|}_{F}$.
%Combining Theorem \ref{thm1} and Lemma \ref{thm3}(3), one gets $\mathcal Z^{k}\rightarrow \mathcal Z^{*}, k\rightarrow\infty$. Then we obtain that $\Delta_k \geq {\|\mathcal Z^{k}-\mathcal Z^{*}\|}_{F}$.
Moreover, Lemma \ref{thm3}(ii) leads to
%With (\ref{test116}) we have
\begin{equation}\label{test111}
\begin{split}
\Delta_k
&=\sum\limits_{j=k}^{\infty}\|\mathcal Z^{j+1}-\mathcal Z^{j}\|_{F}\\
%\leq \sum\limits_{k=0}^{\infty}{\big\|\mathcal Z^{k}-\mathcal Z^{k+1}\big\|}_{F}
&\leq \| \mathcal Z^{k}-\mathcal {Z}^{k-1}\|_F+\frac{\mu}{\rho(1-\theta)}(F(\mathcal Z^{k})-F(\mathcal Z^{*}))^{1-\theta}\\
&= \| \mathcal Z^{k}-\mathcal {Z}^{k-1}\|_F+\frac{\mu}{\rho(1-\theta)}((F(\mathcal Z^{k})-F(\mathcal Z^{*}))^{\theta})^{\frac{1-\theta}{\theta}} \\
%&\leq \| \mathcal Z^{k}-\mathcal {Z}^{k-1}\|_F+\frac{\mu}{\rho(1-\theta)}(\mu\text{dist}(0,\partial F(\mathcal Z^{k})))^{\frac{1-\theta}{\theta}} \\
&\leq \| \mathcal Z^{k}-\mathcal {Z}^{k-1}\|_F+\frac{\mu}{\rho(1-\theta)}\Big(\mu C \|\mathcal Z^{k}-\mathcal Z^{k-1}\|_{F}\Big)^{\frac{1-\theta}{\theta}}\\
&=(\Delta_{k-1}-\Delta_{k})+c_1(\Delta_{k-1}-\Delta_{k})^{\frac{1-\theta}{\theta}},
\end{split}
\end{equation}
where the second inequality holds by (\ref{test126})
%the last inequality holds by (\ref{test128}),
and the last equality holds by letting $c_1:=\frac{\mu}{\rho(1-\theta)}(\mu C)^{\frac{1-\theta}{\theta}}$.
%By using, we have that
%\begin{equation}\label{test112}
%\begin{split}
%\Delta_k
%\end{split}
%\end{equation}
%Suppose that Lemma \ref{thm3} (3) implies that $\sum\limits_{k=0}^{\infty}{\|\mathcal Z^{k+1}-\mathcal Z^{k}\|}_{F}<\infty$.we suppose that $\sum\limits_{j=0}^{\infty}{\|\mathcal Z^{j+1}-\mathcal Z^{j}\|}_{F}$ converges to $S$, where $S<\infty$ is a positive constant. As a consequence,

Let $S_k:=\sum_{j=0}^{k}\|\mathcal Z^{j+1}-\mathcal Z^{j}\|_{F}$.
By Lemma \ref{thm3}(iii),
 we know that $\sum_{j=0}^{\infty}{\|\mathcal Z^{j+1}-\mathcal Z^{j}\|}_{F}<+\infty$,
which implies that ${\lim\limits_{k\to\infty}S_k=\sum_{j=0}^{\infty}{\|\mathcal Z^{j+1}-\mathcal Z^{j}\|}_{F}}$ exists.
Note that
$
\Delta_k=\sum_{j=k}^{\infty}\|\mathcal Z^{j+1}-\mathcal Z^{j}\|_{F}
=\sum_{j=0}^{\infty}\|\mathcal Z^{j+1}-\mathcal Z^{j}\|_{F}-S_{k-1}.
$
Hence
\begin{equation}\label{DelKlim}
\lim\limits_{k\to\infty}\Delta_k=\sum_{j=0}^{\infty}\|\mathcal Z^{j+1}-\mathcal Z^{j}\|_{F}-\lim\limits_{k\to\infty}S_{k-1}=0.
\end{equation}

If $\theta\in(0,\frac{1}{2}]$, then ${\frac{1-\theta}{\theta}}\geq 1$.
Notice that $\Delta_k$ is  monotonically non-increasing.
We observe that (\ref{test111}) reads (for $k$ sufficiently large)
$
\Delta_k \leq c_2(\Delta_{k-1}-\Delta_{k}),
$
where $c_2=1+c_1>0$. Consequently, we have
$
\Delta_{k} \leq \frac{c_2}{c_2+1}\Delta_{k-1}.
$
Let $\vartheta:=\frac{c_2}{c_2+1}\in(0,1)$ and $c=\Delta_0$.
By Lemma \ref{thm3}(iv),
we get that
$$
{\|\mathcal Z^{k}-\mathcal Z^{*}\|}_{F}\leq\Delta_{k}
\leq \vartheta\Delta_{k-1}\leq \cdots \leq \vartheta^k\Delta_0=c\vartheta^k.
$$

(iii) If $\theta\in(\frac{1}{2},1)$, then $0<{\frac{1-\theta}{\theta}}< 1$.
Therefore, by (\ref{test111}), we get that there exists a positive integer $K_0$ such that for any $k>K_0$,
\begin{equation}\label{CKD}
\Delta_k^\frac{\theta}{1-\theta} \leq c_3(\Delta_{k-1}-\Delta_{k}),
\end{equation}
where $c_3=(1+c_1)^\frac{\theta}{1-\theta}$.

Define $f(s):=s^{-\frac{\theta}{1-\theta}}, s>0$.
Let $R\in(1,\infty)$.
First, assume that $f(\Delta_k)\leq Rf(\Delta_{k-1})$,  we can obtain that
\begin{equation}\label{test113}
\begin{split}
c_3^{-1}
&\leq \Delta_k^{-\frac{\theta}{1-\theta}}(\Delta_{k-1}-\Delta_{k})=f(\Delta_k)(\Delta_{k-1}-\Delta_{k})\\
&\leq\int_{\Delta_{k}}^{\Delta_{k-1}} Rf(\Delta_{k-1}) ds\leq \int_{\Delta_{k}}^{\Delta_{k-1}} Rf(s) ds\\
&=\frac{R(1-\theta)}{1-2\theta}\left(\Delta_{k-1}^{\frac{1-2\theta}{1-\theta}}-\Delta_{k}^{\frac{1-2\theta}{1-\theta}}\right),
\end{split}
\end{equation}
where the first inequality holds by (\ref{CKD}).
Let $v:=\frac{1-2\theta}{1-\theta}<0 \text{ and } e_1:=c_3^{-1}\frac{2\theta-1}{R(1-\theta)}$.
The (\ref{test113}) implies that
\begin{equation}\label{test152}
\Delta_{k}^{v}-\Delta_{k-1}^{v}\geq e_1 > 0.
%\Delta_{k}^{v}-\Delta_{k-1}^{v}\geq -v {c_1}^{-\frac{\theta}{1-\theta}}
\end{equation}

Second, assume that $f(\Delta_k)> Rf(\Delta_{k-1})$, i.e.,
$
\Delta_k^{-\frac{\theta}{1-\theta}}>R\Delta_{k-1}^{-\frac{\theta}{1-\theta}},
$
which implies that
$
{\Delta_k}< R^{-\frac{1-\theta}{\theta}}\Delta_{k-1}.
$
It follows from  $v=\frac{1-2\theta}{1-\theta}<0$ that
$$
{\Delta_k^v}> (R^{-\frac{1-\theta}{\theta}})^v\Delta_{k-1}^v,
$$
which implies
$
{\Delta_k^v}-\Delta_{k-1}^v\geq ((R^{-\frac{1-\theta}{\theta}})^v-1)\Delta_{k-1}^v.
$
Since $R^{-\frac{1-\theta}{\theta}}\in(0,1)$, we have $(R^{-\frac{1-\theta}{\theta}})^{v}-1>0$.
Moreover, it can be derived from (\ref{DelKlim})
that there exist a constant $c_4>0$ and a  positive integer $K_1$ such that $\Delta_{k-1}\leq c_4$
for any $k>K_1$.
Therefore, we have $((R^{-\frac{1-\theta}{\theta}})^v-1)\Delta_{k-1}^v>((R^{-\frac{1-\theta}{\theta}})^v-1)c_4^{v}$.
Let $e_2=((R^{-\frac{1-\theta}{\theta}})^v-1)c_4^{v}$, we can deduce that
\begin{equation}\label{test151}
{\Delta_k^v}-\Delta_{k-1}^v > e_2>0.
\end{equation}

Let $\hat{e}:=\min\{e_1,e_2\}$.
Combining (\ref{test152}) with (\ref{test151}), we obtain
\begin{equation}\label{test153}
{\Delta_k^v}-\Delta_{k-1}^v \geq \hat{e}.
\end{equation}
Let $\hat k:=\max\{K_0,K_1\}$.
As a consequence, for any $k\geq 2\hat k$,
we have
$
\Delta_{k}^{v}-\Delta_{\hat k}^{v}=\sum_{l=\hat k+1}^{k}({\Delta_l^v}-\Delta_{l-1}^v)\geq\hat e(k-\hat k),
$
which yields
\begin{equation}\label{test114}
\begin{split}
\Delta_{k}^{v}
&\geq \Delta_{\hat k}^{v}+\hat e(k-\hat k)\geq \hat e(k-\hat k)\geq \frac{\hat e k}{2},
\end{split}
\end{equation}
where the last inequality holds by $\hat k\leq \frac{k}{2}$.
Let $c:=(\frac{\hat e }{2})^{\frac{1}{v}}$.
Therefore, we deduce that
$
{\|\mathcal Z^{k}-\mathcal Z^{*}\|}_{F}\leq \Delta_{k}\leq c k^{\frac{1}{v}}=c k^{-\frac{1-\theta}{2\theta-1}},
$
where the first inequality holds by Lemma \ref{thm3}(iv).
 This completes the proof.
\end{IEEEproof}

%\begin{thebibliography}{1}
\bibliographystyle{IEEEtran}
\bibliography{Refalg1}
%\end{thebibliography}

%\newpage
% your own custom macro containing the $\backslash${\tt{includegraphics}} command to make things
%\bf{If you include a photo:}\vspace{-33pt}
%\begin{IEEEbiography}[{\includegraphics[width=1in,height=1.25in,clip,keepaspectratio]{fig1}}]{Michael Shell}
%Use $\backslash${\tt{begin\{IEEEbiography\}}} and then for the 1st argument use $\backslash${\tt{includegraphics}} to declare and link the author photo.
%Use the author name as the 3rd argument followed by the biography text.
%\end{IEEEbiography}

%
%\vfill

\end{document}